\documentclass[10pt,twocolumn,letterpaper]{article}

\usepackage{cvpr}              

\usepackage{graphicx}
\usepackage{amsmath}
\usepackage{amssymb}
\usepackage{booktabs}

\usepackage[pagebackref,breaklinks,colorlinks]{hyperref}

\usepackage[capitalize]{cleveref}
\crefname{section}{Sec.}{Secs.}
\Crefname{section}{Section}{Sections}
\Crefname{table}{Table}{Tables}
\crefname{table}{Tab.}{Tabs.}

\def\cvprPaperID{348} 
\def\confName{CVPR}
\def\confYear{2022}

\usepackage{tabularx}
\usepackage{multirow}
\usepackage{multicol}
\usepackage{caption}

\usepackage{pifont}

\usepackage[normalem]{ulem}
\useunder{\uline}{\ul}{}

\usepackage{cuted}
\usepackage{capt-of}

\usepackage[export]{adjustbox}
\usepackage[accsupp]{axessibility}

\usepackage{appendix}

\begin{document}

\title{Beyond Cross-view Image Retrieval: \\
Highly Accurate Vehicle Localization Using Satellite Image}

\author{Yujiao Shi and Hongdong Li\\
The Australian National University, Canberra, Australia\\
{\tt\small firstname.lastname@anu.edu.au}
}
\maketitle

\begin{abstract} 
This paper addresses the problem of vehicle-mounted {\em camera localization} by matching a ground-level image with an overhead-view satellite map.  Existing methods often treat this problem as cross-view {\em image retrieval}, and use learned deep features to match the ground-level query image to a partition (\eg, a small patch) of the satellite map. By these methods, the localization accuracy is limited by the partitioning density of the satellite map (often in the order of tens meters).  Departing from the conventional wisdom of image retrieval, this paper presents a novel solution that can achieve highly-accurate localization. The key idea is to formulate the task as pose estimation and solve it by neural-net based optimization. Specifically, we design a two-branch {CNN} to extract robust features from the ground and satellite images, respectively. To bridge the vast cross-view domain gap, we resort to a Geometry Projection module that projects features from the satellite map to the ground-view, based on a relative camera pose. Aiming to minimize the differences between the projected features and the observed features, we employ a differentiable Levenberg-Marquardt ({LM}) module to search for the optimal camera pose iteratively. The entire pipeline is differentiable and runs end-to-end. 
Extensive experiments on standard autonomous vehicle localization datasets have confirmed the superiority of the proposed method. Notably, \eg, starting from a coarse estimate of camera location within a wide region of $40\text{m}\times40\text{m}$, with an 80\% likelihood our method quickly reduces the lateral location error to be within $5\text{m}$ on a new {KITTI} cross-view dataset.

\end{abstract}

\section{Introduction}
\label{sec:intro}

\begin{figure}[t!]
\setlength{\abovecaptionskip}{0pt}
    \setlength{\belowcaptionskip}{0pt}
    \centering
    \begin{subfigure}{\linewidth}
    \centering
    \includegraphics[width=0.95\linewidth]{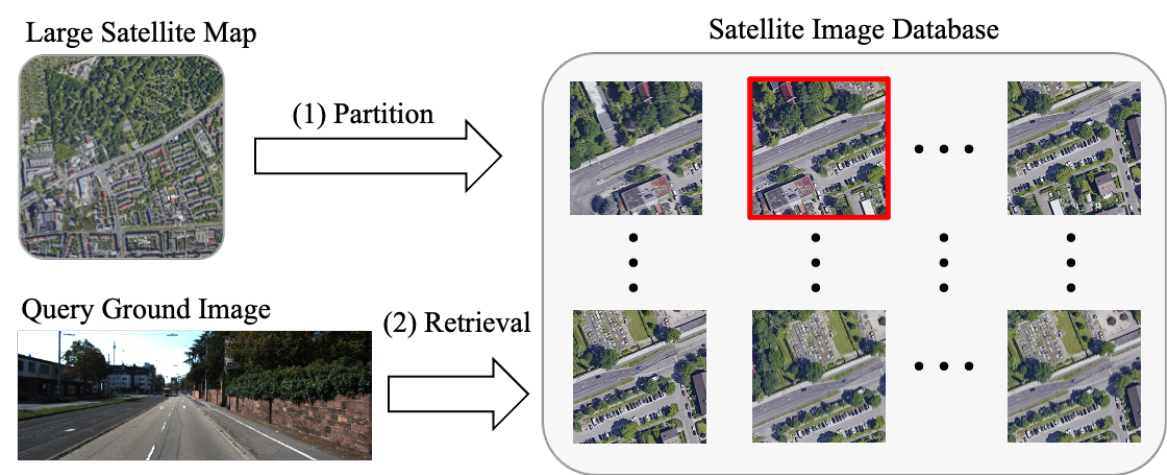}
    \caption{Cross-view image retrieval-based localization}
    \label{openfig:IR}
    \end{subfigure}
    \begin{subfigure}{\linewidth}
    \centering
    \includegraphics[width=0.95\linewidth]{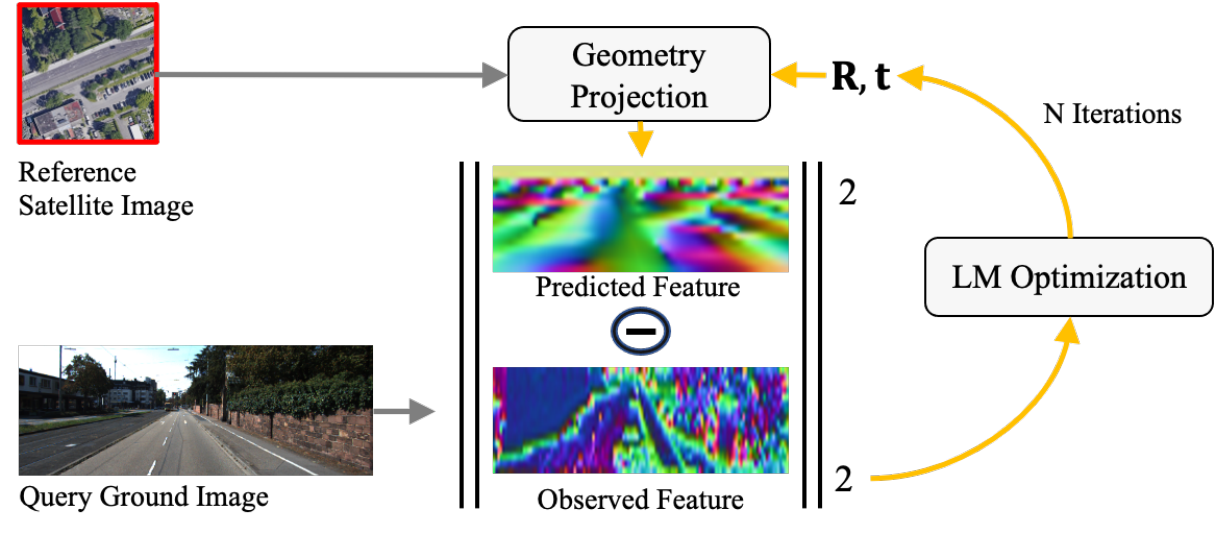}
    \caption{Proposed cross-view camera pose optimization}
    \label{openfig:LM}
    \end{subfigure}
    \caption{\small Comparison of two schemes for cross-view camera localization. 
    (a) Conventional image retrieval-based scheme first splits a large satellite map into small image patches and constructs a reference database. Given a query image, they find the most similar image from a database.  GPS-tag of the retrieved image is regarded as the query camera's location.  By this method, the localization accuracy is limited by the sample density of database images. 
    (b) The paper proposes a novel scheme for cross-view localization, which formulates the task as camera pose optimization.  We tentatively project the satellite deep features to a ground viewpoint from an initial camera pose estimate by a Geometry Projection module. Then, a differentiable Levernberg-Marquardt optimization procedure is applied to refine the camera pose estimation by minimizing the differences between the predicted and observed features. 
    }
    \label{fig:openfigure}
\end{figure}

Image-based camera localization~\cite{arandjelovic2016netvlad, chen2017deep, lowry2015visual, sattler2016large, torii201524, sattler2017large, kim2017learned, noh2017large, ge2020self} has attracted increasing attention from the community due to its practical applications in various fields, including autonomous driving, virtual and augmented reality. Recently, this technique has been extended to the cross-view setting, \ie, localization by matching a ground-level image to an overhead-view satellite map to determine the ground camera's pose.

Existing learning-based image localization methods often treat this task as an instance of image retrieval, and solve it by {\em metric learning} \cite{workman2015location, workman2015wide, vo2016localizing, tian2017cross,  zhai2017predicting, Hu_2018_CVPR, Liu_2019_CVPR, Regmi_2019_ICCV, Cai_2019_ICCV, sun2019geocapsnet, shi2019spatial, shi2020optimal, shi2020looking, zhu2021revisiting, toker2021coming}.
They match a ground image to many candidate satellite patches in a large satellite map covering the geographical region of interest and then retrieve the most similar one.
The query camera's location is then assigned as the GPS tag of the retrieved satellite patch. Although promising results have been achieved, their estimation accuracy is limited by the sampling density of those satellite images.  A recent work~\cite{zhu2021vigor} further discussed location refinement by using a deep network to regress the relative displacement. However, the significant domain differences between the satellite and ground-view images makes it very difficult to obtain accurate regression in the cross-view setting.

Departing from the traditional idea of image retrieval, we propose to solve accurate cross-view localization task by direct pose optimization. Specifically, we first employ two CNNs to extract deep features from the two-view images. The learned features are expected to be robust to view changes and discriminative for feature correspondences. Then, we devise a Geometry Projection module, which approximately projects satellite features to a ground viewpoint, based on the current camera pose estimate, to bridge the domain gap between the views. Finally, a differentiable Levenberg-Marquardt (LM) algorithm is embedded in the pipeline to refine the pose. The LM optimization aims to find the optimal camera pose such that the predicted deep features originated from the satellite image match well to the corresponding deep features extracted from the ground-view image. Please refer to Fig.~\ref{fig:openfigure} for an overview.

We evaluate our method on two standard benchmarks for autonomous driving, \ie  KITTI~\cite{geiger2013vision} and Ford multi-AV datasets~\cite{agarwal2020ford}. 
Both the datasets contain ground-level images by a vehicle-mounted camera with GT poses but without satellite maps. We supplement them with corresponding high-definition satellite maps, downloaded from Google Map~\cite{google}, for the evaluation of the proposed method.

\section{Prior Arts}

\textbf{Image based localization.} 
Image-based localization is often formulated as an image retrieval problem and tackled by metric learning techniques.  It is solved by ground-to-ground (G2G) image matching~\cite{arandjelovic2016netvlad, chen2017deep, lowry2015visual, sattler2016large, torii201524, sattler2017large, kim2017learned, noh2017large, ge2020self, cummins2008fab, galvez2012bags, mur2015orb}. 
Since the G2G image matching cannot localize query images whose reference counterparts are not available, many recent works resort to the widespread satellite images to construct the database~\cite{castaldo2015semantic, workman2015location, vo2016localizing, tian2017cross,  zhai2017predicting, Hu_2018_CVPR, Liu_2019_CVPR, Regmi_2019_ICCV, Cai_2019_ICCV, sun2019geocapsnet, shi2019spatial, shi2020optimal, shi2020looking, zhu2021revisiting, toker2021coming, shi2022geometry, shi2022accurate}.  

These works approximate the query camera pose as the pose of the top-1 retrieved reference image. 
They remain effective at scale, but the pose estimates in this manner are very coarse. 
In this work, we introduce a novel approach to increase localization accuracy.

\smallskip
\textbf{3D structure based localization.}
Works on 3D structure based localization usually employ a 3D scene model as reference for query camera localization~\cite{svarm2016city, donoser2014discriminative, li2012worldwide, li2010location, lynen2015get, sattler2015hyperpoints, sattler2016efficient, taira2018inloc, zeisl2015camera, sattler2018benchmarking, toft2020long, liu2019stochastic, cao2014minimal, larsson2016outlier, zhou2020da4ad, brejcha2020landscapear}. 
Among these algorithms, works~\cite{sarlin2021back, von2020lm} also use the LM optimization for camera pose estimation. However, they are designed for ground-to-ground localization only and require knowing the 3D coordinates of image key points. 
This paper only uses a high-definition satellite image as a reference and solves the ground-to-satellite localization when 3D scene models are unavailable.

\smallskip
\textbf{SLAM/VO.} Simultaneous Localization and Mapping (SLAM) and Visual Odometry (VO) techniques have been traditionally used for vehicle localization~\cite{ventura2014global, schneider2018maplab, mur2017visual, middelberg2014scalable, lynen2020large, lynen2015get, jones2011visual, geppert2019efficient, dutoit2017consistent, stenborg2020using, hu2020image}.  They first estimate relative camera poses between consecutive image frames, and then integrate them for global pose computation. As such, they suffer from error accumulation, leading to an estimation drift. 
Our proposed method only relies on a single frame. Hence, it can complement the SLAM/VO method as a novel way of (satellite image-based) loop-closure. 
\begin{figure*}[ht]
\setlength{\abovecaptionskip}{0pt}
    \setlength{\belowcaptionskip}{0pt}
    \centering
    \includegraphics[width=0.82\linewidth]{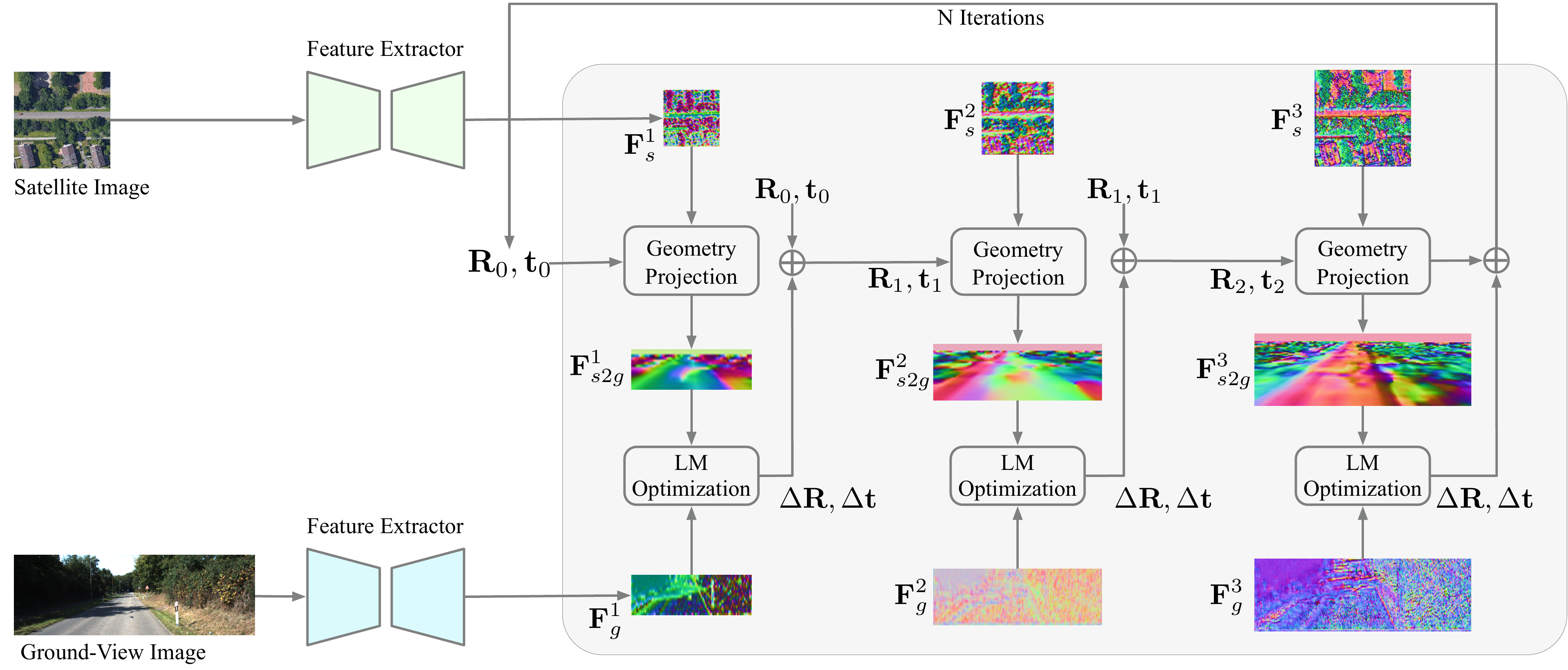}
    \caption{\small An overview of the proposed highly-accurate camera pose optimization procedure. 
    Our method employs a two-branch deep network to extract multi-scale robust features from ground and satellite images, respectively. Next, the Geometry Projection module maps the satellite features to the ground-view domain, based on an initial camera pose $\mathbf{R}_0, \mathbf{t}_0$. 
    By minimizing the differences between the projected satellite features $\mathbf{F}_{s2g}$ and the observed ground-view features $\mathbf{F}_{g}$, an LM optimization is employed to find the optimal camera pose. The LM optimization is executed in a coarse to fine manner. }
    \label{fig:framework_high_accuracy}
\end{figure*}

\section{Method Overview}

Given a coarse initial estimate of a ground camera's pose, we aim to optimize this pose in high accuracy by matching it to a companion satellite map.   
Instead of formulating this task as image retrieval, we propose a pose optimization framework described below. 

Our framework consists of three components: (i) a two-branch deep network for feature learning, (ii) a Geometry Projection module, and (iii) a differentiable Levenberg-Marquardt (LM) optimizer, as shown in Fig.~\ref{fig:framework_high_accuracy}.

\subsection{Deep robust feature learning}
Deep neural networks are shown to be powerful in learning robust features that are resilient to viewpoint changes and suitable for the task of visual localization~\cite{workman2015location, workman2015wide, vo2016localizing, tian2017cross, Liu_2019_CVPR, Regmi_2019_ICCV, Cai_2019_ICCV, sun2019geocapsnet, shi2019spatial, shi2020looking}. 
We design a two-branch neural network to extract deep features from the ground and satellite images separately. The U-net structure is used to learn multi-scale feature representations. 

\subsection{Cross-view feature alignment}
To bridge the evident cross-view domain gap, we devise a Geometry Projection module which aligns the two view features in the ground-view domain based on cross-view geometry.  Our geometry projection module projects satellite-view features to a ground viewpoint by establishing approximate geometric correspondences and using a relative camera pose between the two views.

\subsection{Iterative pose refinement}
We aim to find the optimal ground camera pose such that the projected features from the satellite domain are the most similar to the real observed ground-view features. To this end, we develop a new differentiable LM optimizer for cross-view feature alignment. The LM optimizer iteratively refines the camera pose to match the cross-view feature maps.  

When the initial camera pose is too far from the ground-truth pose, the projected contents from a satellite image will be considerably different from the observed ground image, causing a local minima problem. We apply a coarse-to-fine multi-scale LM update strategy to mitigate this issue.  Features at the coarsest level have a larger receptive field in the original image and thus are suitable for coarse-level global search in the solution space. Conversely, features at finer scales have a larger spatial resolution and encode more detailed scene information. Hence, they are more informative for precise pose refinement.

\section{Highly-accurate Pose Optimization}

In this section, we provide a detailed analysis of the proposed method. 
As mentioned before, we adopt a two branch CNN to extract deep features from the ground and satellite images,
denoted as $\mathbf{F}_g^l \in \mathbb{R}^{H_g^l \times W_g^l \times C^l}$ and $\mathbf{F}_s^l \in \mathbb{R}^{H_s^l \times W_s^l \times C^l}$, respectively, where $l = 1, 2, 3$ indicates the scale level. 
The ground and satellite branches share the same architecture but do not share weights. In this way, they can adapt to their respective domains.
The features at each level are $L_2$ normalized to increase their robustness for cross-view matching.

\subsection{Satellite-to-ground geometry projection}

We introduce a Geometry Projection module to establish the cross-view geometric correspondences. 
Our geometry projection module projects satellite-view features to a ground viewpoint based on the relative camera pose between the two view images. 

We set the world coordinate system to the initial camera pose estimate, with its location corresponds to the reference satellite image center, $x$ axis parallel to the $v^s$ direction of the satellite image, $y$ axis pointing downward, and $z$ axis parallel to the $u^s$ direction.  
A 3D point $(x, y, z)$ in the world coordinate system is mapped to a satellite pixel coordinate by the orthographic projection, 
\begin{equation}
    [u_s, v_s]^T = [\frac{z}{\alpha} + u_s^0, \frac{x}{\alpha} + v_s^0]^T,
    \label{eq:sat}
\end{equation}
where $\alpha$ as the per-pixel real-word distance of a satellite feature map, and $(u_s^0, v_s^0)$ is the satellite feature map center. 


The transformation from the real camera coordinate system to the world coordinate system is formulated as

\begin{equation}
    [x, y, z]^T = \mathbf{R} ([x_c, y_c, z_c]^T + \mathbf{t}), 
    \label{eq:world}
\end{equation}
where $\mathbf{R}$ and $\mathbf{t}$ are the rotation and translation matrices, respectively.
The projection from a 3D point to a pin-hole camera image plane is given by 
\begin{equation}
    w [u^g, v^g, 1]^T = \mathbf{K} [x_c, y_c, z_c]^T, 
    \label{eq:grd}
\end{equation}
where $\mathbf{K}$ is the camera intrinsic and $w$ is a scale factor. 

From Eq.~\eqref{eq:sat}$\sim$\eqref{eq:grd}, we can derive the mapping from a ground-view pixel to a satellite pixel as
\begin{equation}
\begin{bmatrix}
u_s  \\
v_s \\
 z 
\end{bmatrix} = 
\begin{bmatrix}
\frac{1}{\alpha} & 0 & 0 \\
0 &\frac{1}{\alpha}   & 0  \\
 0&0  & 1 
\end{bmatrix} 
\left ( 
w \mathbf{R}\mathbf{K}^{-1}\begin{bmatrix}
u_g \\
v_g \\
1
\end{bmatrix} + \mathbf{R}\mathbf{t}
 \right )
+
\begin{bmatrix}
 u_s^0\\
 v_s^0\\
0
\end{bmatrix}.
\label{eq:h}
\end{equation}

When the depth map of the ground-view image is available, \ie, $w$ is given, the satellite to ground projection can be easily conducted by Eq.~\eqref{eq:h}. 
However, it is challenging to estimate depths from a single ground image. 
Considering the overlap between a ground and a satellite image mainly lies on the ground plane, our geometry projection is conducted by using the homography of the ground plane. 
In other words, we make all the ground-view pixels corresponds to the points on the ground plane by setting $y_c$ in Eq.~\eqref{eq:grd} to the distance between the query camera to the ground plane. 
Then, $w$ can be computed from Eq.~\eqref{eq:grd}. 

This projection defined on ground plane homography is only approximately correct.  To handle objects higher than the ground place and reduce distortions in the projection, we project deep features rather than RGB pixels to measure the gap. These deep features encode high-level semantic information and therefore are less sensitive to object heights than RGB values. 
By Eq.~\eqref{eq:h}, we conduct the satellite to ground geometry projection using bilinear interpolation, obtaining {\small $\mathbf{F}_{s2g}^l \in \mathbb{R}^{H_g^l \times W_g^l \times C^l}$} as the projected ground-view features from satellite features at scale $l$. 


\subsection{Multi-level LM optimization}
\label{sec:LM}

The differences between the satellite and the ground observations are given by, 
\begin{equation}
    \mathbf{e}^l = \mathbf{F}_{s2g}^l - \mathbf{F}_{g}^l. 
    \label{eq:e}
\end{equation}
The objective is to find the optimal pose $\hat{\mathbf{R}}$ and $\hat{\mathbf{t}}$ of the ground camera by minimizing the following loss function,
\begin{equation}
\hat{\xi} = \mathop{\arg \min}_{\xi} \| \mathbf{e}^l \|_2^2, 
    \label{eq:LM_objective}
\end{equation}
where $\| \cdot \|_2^2$ is the $L_2$ norm, $\xi = \{\mathbf{R}, \mathbf{t}\}$ and $\hat{\xi}$ corresponds to its optimal solution. 
We solve this non-linear least squares problem by the Levenberg-Marquardt(LM) optimization algorithm~\cite{levenberg1944method,marquardt1963algorithm}. 

For each level $l$, we compute a Jacobian matrix and a Hessian matrix, 
\begin{equation}
    \mathbf{J} = \frac{\partial \mathbf{F}_{s2g}}{\partial {\xi}} = \frac{\partial \mathbf{F}_{s2g}}{\partial {\mathbf{p}_s}} \frac{\partial {\mathbf{p}_s}}{\partial \xi }, \quad \text{and}\quad \mathbf{H} = \mathbf{J}^T\mathbf{J}, 
    \label{eq:lm_jh}
\end{equation}
where $\mathbf{p}_s$ is the satellite feature map coordinates. 


We choose to use Levenberg's damping formula \ie, 
\begin{math}
    \widetilde{\mathbf{H}} = \mathbf{H} + \lambda \mathbf{I},\end{math} for its convenience in network training.  $\lambda$ is the trade-off parameter which interpolates between the gradient decent ($\lambda = \infty $) and Gaussian-Newton ($\lambda = 0$) algorithm. Alternatively, the Marquardt's damping formula may be used instead \cite{marquardt1963algorithm}.

The pose is updated by, 
\begin{equation}
    \xi_{t+1} = \xi_t + {\widetilde{\mathbf{H}}}^{-1}\mathbf{J}^T\mathbf{e},
    \label{eq:lm_update}
\end{equation}
where $t$ index iterations. 

The LM optimization is first applied at the coarsest feature level and gradually propagates to finer levels. 
This coarse-to-fine (C2F) scheme is executed iteratively until it converges or reaches a maximum iteration of $5$. This multi-scale C2F procedure offers an opportunity to escape from local minima and is more likely to find the global optimum. 

We had attempted to embed a confidence map in Eq.~\eqref{eq:LM_objective}. The intention was to give higher weight to salient visual features (\eg, corner points) and lower weight to textureless regions. However, in our experiments, we did not observe consistent improvement across different test sets. Hence this idea is not employed in our current method.

\begin{figure*}[t!]
\setlength{\abovecaptionskip}{0pt}
    \setlength{\belowcaptionskip}{0pt}
    \centering
    \begin{subfigure}{0.32\linewidth}
    \begin{minipage}{0.64\linewidth}
    \includegraphics[width=\linewidth, height=0.5\linewidth]{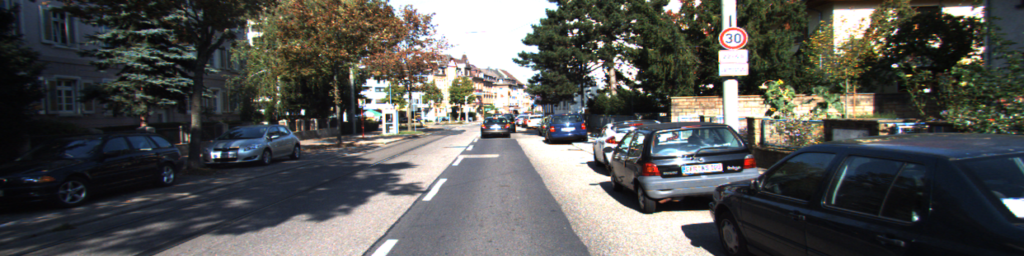}
    \centerline{\footnotesize Query}
    \end{minipage}
    \begin{minipage}{0.32\linewidth}
    \adjincludegraphics[width=\linewidth,trim={{\width/4} {\width/4} {\width/4} {\width/4}},clip]{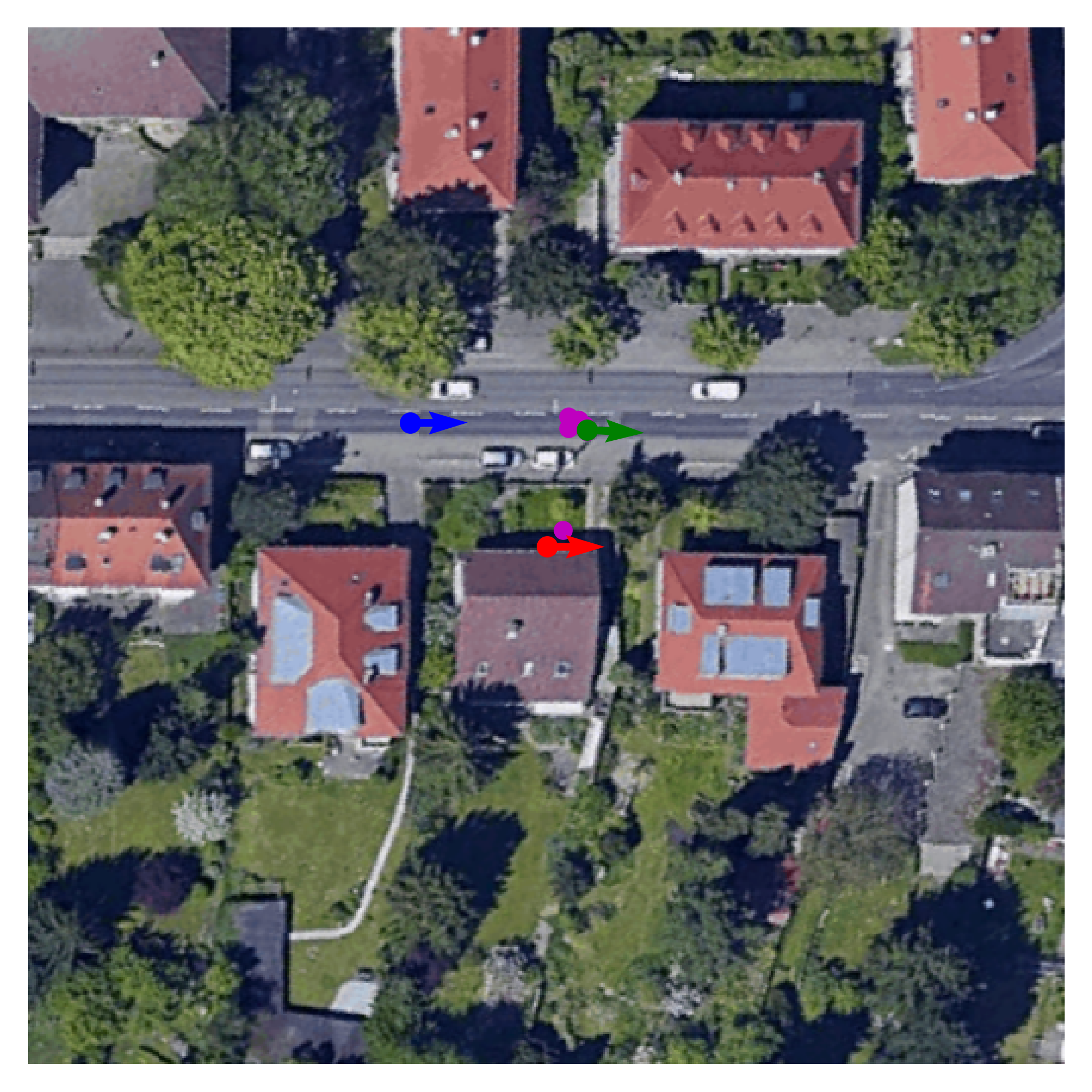}
   \centerline{\footnotesize Reference}
    \end{minipage}
    \end{subfigure}
    \begin{subfigure}{0.32\linewidth}
    \begin{minipage}{0.64\linewidth}
    \includegraphics[width=\linewidth, height=0.5\linewidth]{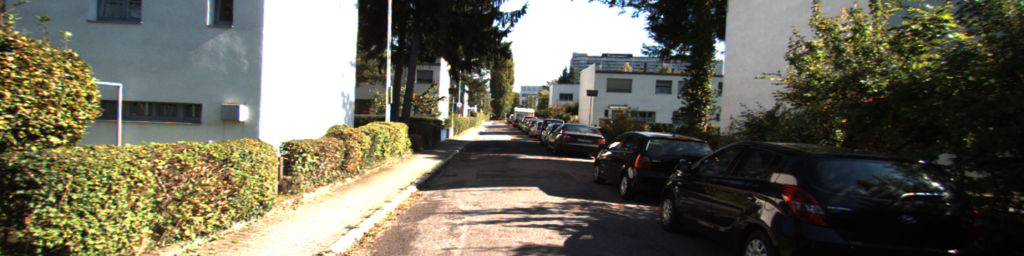}
    \centerline{\footnotesize Query}
    \end{minipage}
    \begin{minipage}{0.32\linewidth}
    \adjincludegraphics[width=\linewidth,trim={{\width/4} {\width/4} {\width/4} {\width/4}},clip]{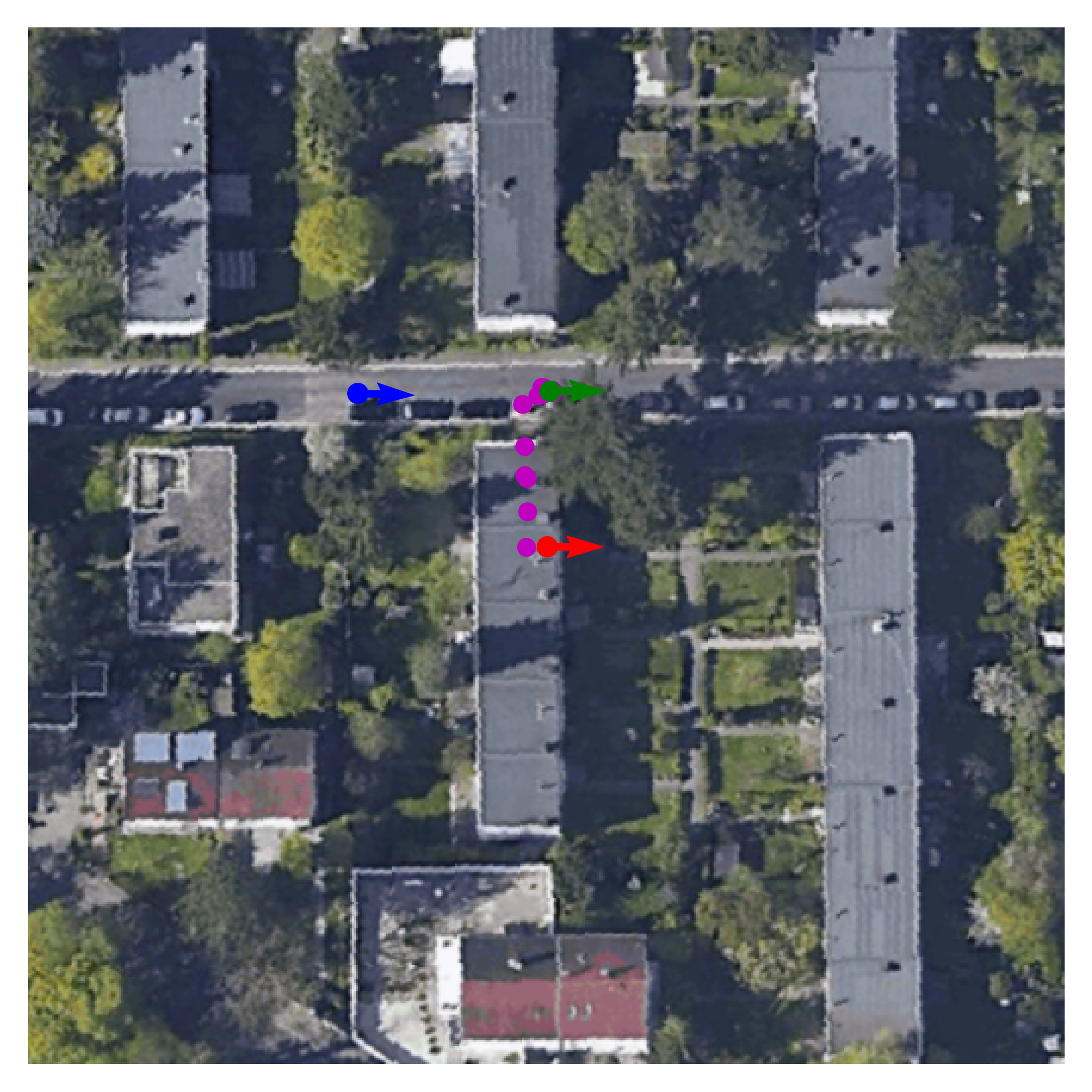}
    \centerline{\footnotesize Reference}
    \end{minipage}
    \end{subfigure}
     \begin{subfigure}{0.32\linewidth}
    \begin{minipage}{0.64\linewidth}
    \includegraphics[width=\linewidth, height=0.5\linewidth]{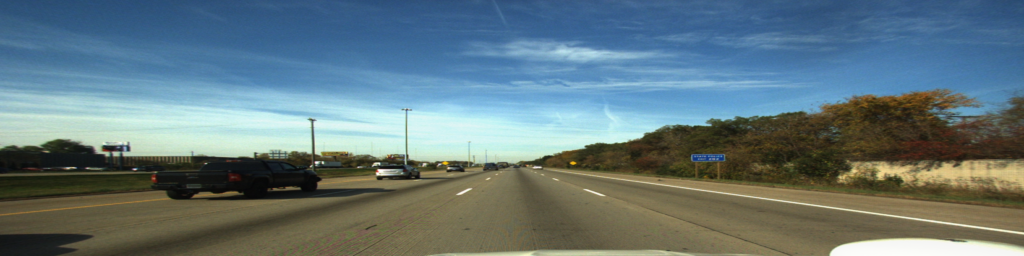}
    \centerline{\footnotesize Query}
    \end{minipage}
    \begin{minipage}{0.32\linewidth}
    \adjincludegraphics[width=\linewidth,trim={{\width/4} {\width/4} {\width/4} {\width/4}},clip]{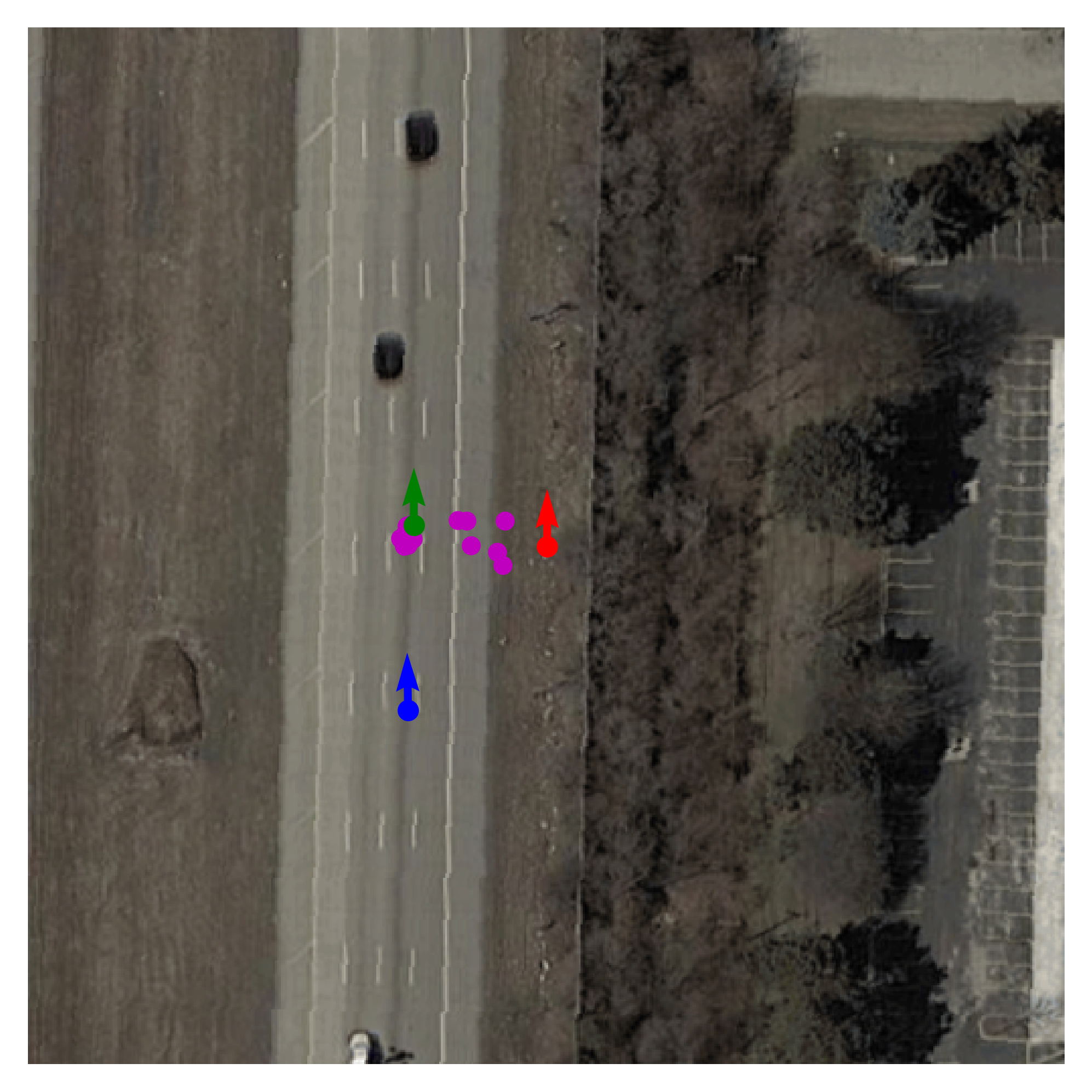}
    \centerline{\footnotesize Reference}
    \end{minipage}
    \end{subfigure}
    \caption{\small Failure cases of our method on longitudinal location optimization. 
    The scene facades shown in the ground-view are not visible in the overhead view. We can only determine the lateral translation of a ground vehicle. Driving along a highway suffers even considerable ambiguity on the longitudinal location optimization. Here, red arrows represent initial poses, green arrows denote final predicted poses, and blue arrows indicate GT poses. The magenta points pinpoint intermediate locations during optimization.}
    \label{fig:failure}
\end{figure*}

\begin{table*}[t!]
\setlength{\abovecaptionskip}{0pt}
\setlength{\belowcaptionskip}{0pt}
\setlength{\tabcolsep}{2pt}
\centering
\footnotesize
\caption{\small Performance comparison between our method and state-of-the-art methods on KITTI dataset.}
\begin{tabular}{c|ccc|ccc|ccc|ccc|ccc|ccc}
\toprule
           & \multicolumn{9}{c|}{Test1}                                                                                                                             & \multicolumn{9}{c}{Test2}                                                                                                                             \\
           & \multicolumn{3}{c|}{Lateral}                          & \multicolumn{3}{c|}{Longitudinal}                         & \multicolumn{3}{c|}{Azimuth}                      & \multicolumn{3}{c|}{Lateral}                          & \multicolumn{3}{c|}{Longitudinal}                         & \multicolumn{3}{c}{Azimuth}                      \\
           & $d=1$          & $d=3$          & $d=5$          & $d=1$         & $d=3$          & $d=5$          & $\theta=1$     & $\theta=3$     & $\theta=5$     & $d=1$          & $d=3$          & $d=5$          & $d=1$         & $d=3$          & $d=5$          & $\theta=1$     & $\theta=3$     & $\theta=5$     \\\midrule
CVM-NET~\cite{Hu_2018_CVPR}    & 5.83           & 17.41          & 28.78          & 3.47          & 11.18          & 18.42          & -              & -              & -              & 6.96           & 21.55          & 35.24          & 3.58          & 10.45          & 17.53          & -              & -              & -              \\
CVFT~\cite{shi2020optimal}       & 7.71           & 22.37          & 36.28          & 3.82          & 11.48          & 18.63          & -              & -              & -              & 7.20           & 22.05          & 36.21          & 3.63          & 11.11          & 18.46          & -              & -              & -              \\
SAFA~\cite{shi2019spatial}       & 9.49           & 29.31          & 46.44          & 4.35          & 12.46          & 21.10          & -              & -              & -              & 9.15           & 27.83          & 44.27          & 4.22          & 11.93          & 19.65          & -              & -              & -              \\
Polar-SAFA~\cite{shi2019spatial} & 9.57           & 30.08          & 45.83          & 4.56          & 13.01          & 21.12          & -              & -              & -              & 10.02          & 29.09          & 46.19          & 3.82          & 11.87          & 19.84          & -              & -              & -              \\
DSM~\cite{shi2020looking}        & 10.12          & 30.67          & 48.24          & 4.08          & 12.01          & 20.14          & 3.58           & 13.81          & 24.44          & 10.77          & 31.37          & 48.24          & 3.87          & 11.73          & 19.50          & 3.53           & 14.09          & 23.95          \\
VIGOR~\cite{zhu2021vigor}      & 18.61          & 49.06          & 69.79          & 4.29          & 13.01          & 21.47          & -              & -              & -              & 17.38          & 48.20          & 70.79          & 4.07          & 12.52          & 20.14          & -              & -              & -              \\
\textbf{Ours}       & \textbf{35.54} & \textbf{70.77} & \textbf{80.36} & \textbf{5.22} & \textbf{15.88} & \textbf{26.13} & \textbf{19.64} & \textbf{51.76} & \textbf{71.72} & \textbf{27.82} & \textbf{59.79} & \textbf{72.89} & \textbf{5.75} & \textbf{16.36} & \textbf{26.48} & \textbf{18.42} & \textbf{49.72} & \textbf{71.00}\\ \bottomrule
\end{tabular}
\label{tab:sota_kitti}
\end{table*}

\begin{table*}[t!]
\setlength{\abovecaptionskip}{0pt}
\setlength{\belowcaptionskip}{0pt}
\setlength{\tabcolsep}{2pt}
\centering
\footnotesize
\caption{\small Performance comparison between our method and state-of-the-art image retrieval approaches on the Ford multi-AV dataset.}
\begin{tabular}{c|ccc|ccc|ccc|ccc|ccc|ccc}
\toprule
           & \multicolumn{9}{c|}{Log1}                                                                                                                             & \multicolumn{9}{c}{Log2}                                                                                                                             \\
           & \multicolumn{3}{c|}{Lateral}                          & \multicolumn{3}{c|}{Longitudinal}                         & \multicolumn{3}{c|}{Azimuth}                      & \multicolumn{3}{c|}{Lateral}                          & \multicolumn{3}{c|}{Longitudinal}                         & \multicolumn{3}{c}{Azimuth}                      \\
           & $d=1$          & $d=3$          & $d=5$          & $d=1$         & $d=3$          & $d=5$          & $\theta=1$     & $\theta=3$     & $\theta=5$     & $d=1$          & $d=3$          & $d=5$          & $d=1$         & $d=3$          & $d=5$          & $\theta=1$     & $\theta=3$     & $\theta=5$     \\\midrule
CVM-NET~\cite{Hu_2018_CVPR}    & 9.14           & 25.67          & 41.33          & 4.81          & 13.19          & 21.90          & -              & -              & -              & 9.82           & 28.60          & 47.06          & 4.24          & 11.83          & 20.34          & -             & -              & -              \\
CVFT~\cite{shi2020optimal}       & 10.57          & 31.10          & 51.19          & 3.52          & 11.43          & 20.38          & -              & -              & -              & 12.21          & 35.07          & 57.61          & 4.40          & 12.18          & 21.41          & -             & -              & -              \\
SAFA~\cite{shi2019spatial}       & 9.33           & 28.71          & 47.95          & 4.33          & 11.76          & 20.14          & -              & -              & -              & 11.22          & 34.10          & 53.39          & 5.02          & 13.36          & 22.89          & -             & -              & -              \\
Polar-SAFA~\cite{shi2019spatial} & 9.05           & 28.62          & 47.10          & 4.43          & 12.14          & 21.10          & -              & -              & -              & 12.02          & 35.63          & 56.21          & 4.29          & 12.13          & 20.28          & -             & -              & -              \\
DSM~\cite{shi2020looking}        & 12.00          & 35.29          & 53.67          & 4.33          & 12.48          & 21.43          & 3.52           & 13.33          & 23.67          & 8.45           & 24.85          & 37.64          & 3.94          & 12.24          & 21.41          & 2.23          & 7.67           & 13.42          \\
VIGOR~\cite{zhu2021vigor}      & 20.33          & 52.48          & 70.43          & \textbf{6.19} & 16.05          & 25.76          & -              & -              & -              & 20.87          & 54.87          & 75.64          & \textbf{5.98} & \textbf{16.88} & \textbf{27.23} & -             & -              & -              \\
\textbf{Ours}       & \textbf{46.10} & \textbf{70.38} & \textbf{72.90} & 5.29          & \textbf{16.38} & \textbf{26.90} & \textbf{44.14} & \textbf{72.67} & \textbf{80.19} & \textbf{31.20} & \textbf{66.46} & \textbf{78.27} & 4.80          & 15.27          & 25.76          & \textbf{9.74} & \textbf{30.83} & \textbf{51.62}
\\ \bottomrule
\end{tabular}
\label{tab:sota_ford}
\end{table*}

\subsection{Training objective}

The LM optimization is implemented in a differentiable manner in our pipeline (within a feed-forward pass). The network is trained end-to-end. We use the GT camera poses as our network supervision, 
\begin{equation}
    \mathcal{L} = \sum_t \sum_l (\|\hat{\mathbf{R}}_t^l - \mathbf{R}^*\|_1 + \|\hat{\mathbf{t}}_t^l - \mathbf{t}^*\|_1), 
\end{equation}
where $\hat{\mathbf{R}}_t^l$ and $\hat{\mathbf{t}}_t^l$ is the predicted pose by our method at the $t_{\text{th}}$ iteration and $l_{\text{th}}$ level, $\mathbf{R}^*$ and $\mathbf{t}^*$ is the GT camera pose.  

During training, when the camera pose provided by the LM optimization deviates from the GT value, the error will be backpropagated to the feature extraction network and update its parameters. In this way, our network is trained to learn useful cross-view features for pose optimization.



\section{Satellite-augmented KITTI and Ford Multi-AV dataset}
We evaluate the feasibility of the proposed method in two standard autonomous driving datasets,\ie,  KITTI~\cite{geiger2013vision} and Ford multi-AV dataset~\cite{agarwal2020ford}. Cameras used in both dataset are intrinsically calibrated. 
Based on the GPS provided by the datasets, we collect satellite images from Google Map~\cite{google}. The satellite images are downloaded with zoom 18. 
The per-pixel resolution for satellite images in KITTI is $0.20$ m, and for Ford multi-AV, it is $0.22$ m.

\smallskip \noindent
\textbf{KITTI.} The KITTI dataset contains stereo images captured by a moving vehicle from different trajectories at different times. 
There is barely any revisited trajectory. 
We split the entire dataset (raw data) into three subsets, one for training and two for testing, denoted as Training, Test1, and Test2, respectively. 
The Training and Test1 sets are from the same region, while Test2 is in a different area. Test2 is used to evaluate the generalization ability of an algorithm. 
We use the left image in a stereo pair as our query image. 

\smallskip \noindent
\textbf{Ford multi-AV dataset. } 
The Ford multi-AV dataset consists of data captured by three vehicles, V1, V2, and V3. Each vehicle is equipped with $7$ cameras. 
Among the three vehicles, only V2 captured images from six trajectories (Log1$\sim$Log6) at two different dates/drives, \ie, 2017-08-04 and 2017-10-26. 
Hence, we use the front left camera of V2 as our query images. For each trajectory, we split testing and training sets based on different drives. 
\smallskip \noindent
{\bf{Evaluation Metrics.}} 
Satellite images can only provide a vehicle's location and orientation (\ie, azimuth angle) reference. Thus this paper estimates a 3-DoF vehicle pose by ground-to-satellite matching. We report vehicle's location errors along the longitudinal direction (\ie, driving direction) and along the lateral direction, separately. This is because, using satellite map to localization, the uncertainty of vehicle location along the driving direction is often more considerable than that along the lateral direction. For instance, when a vehicle is driving on the road with tall buildings on both sides, the vehicle's location along the driving direction is mainly determined by the building facades appearance. However, facades are not visible from the satellite image. Driving on a highway is another scenario where the ambiguity is significant because the scenes along the driving direction are often monotonous and repetitive--uninformative for localization.  In contrast, lateral vehicle location can be obtained more reliably using road boundaries. Moreover, multi-lane freeways in rural areas are almost always visible on a satellite map.  
 
When the estimated translation of a camera is within $d$ m to its GT translation along a direction, it is deemed a correct estimation. When the estimated value of a rotation angle is within $\theta$ to its GT value, the estimation is deemed correct. We set $d$ to $1$, $3$, and $5$ respectively, and $\theta$ to $1^\circ$, $3^\circ$ and $5^\circ$ respectively.  Since this work focuses on autonomous driving, we did not test on other cross-view datasets (\eg, \cite{zhai2017predicting}, ~\cite{Liu_2019_CVPR} and ~\cite{zhu2021vigor}), also because they do not provide ground-view heading directions relative to the satellite map, making a meaningful comparison harder. 

\section{Experiments} 
We first compare our method with those fine-grained image retrieval methods, and then conduct experiments to analyze each component in our framework. 

\smallskip \noindent
\textbf{Implementation details.}
The satellite image resolution used in our experiments is $512\times512$, corresponding to a coverage of around $100\times 100 \ \text{m}^2$. 
We assume the city-scale image retrieval has restricted the camera location to be in a region of $40\text{m}\times40\text{m}$ around the satellite image center. 
Within this region, we conduct a high-accuracy pose search. 
The choice of this search region guarantees that the satellite image can provide a reference of at least $30 \ \text{m}$ visual distance for a query camera,  \eg, when a query camera is on the boundary of the region and looking outside. 
The resolution of ground-view images is $256\times1024$. 
The feature-level $l$ corresponds to a scale of $\frac{1}{2^{4-l}}$ with respect to the original image resolution. 
We use Adam optimizer to train our network with a learning rate of $10^{-4}$, $\beta_1=0.9$, and $\beta_2=0.999$. 
The network is trained by 2 epochs on a RTX3090 GPU with a batch size of 3.  
Our method is fully implemented in Pytorch. The pose optimization runtime for a query image is around $500\ \text{ms}$. 
Unless specifically stated, the rotation noise is set to $20^\circ$ throughout the experiments.
The source code and datasets can be accessed at \url{https://github.com/shiyujiao/HighlyAccurate.git}.

\subsection{Comparison with fine-grained image retrieval} 
Given a query image, its retrieved satellite counterpart from a city-scale database provides a coarse location estimate of this query image. 
To refine this pose estimate, a reasonable approach might be splitting the satellite image into small patches further and conducting a fine-grained image retrieval.  
Hence, we first show the performance of the state-of-the-art cross-view localization algorithms by using this fine-grained image retrieval for camera pose refinement.

\textbf{Settings.}
During inference stage for image retrieval based method, we sample a grid within the $40\text{m}\times40\text{m}$ search region uniformly and crop corresponding satellite patches centered at the grid points to construct the fine-grained retrieval database. 
Since our method searches $15$ possible solutions, \ie, $3$ feature levels$\times$5 iterations, the grid size is set to $4 \times 4$ in the fine-grained retrieval for a fair comparison. 
Note that the regular discretized grids only apply to the inference stage. During training stage, the grids are continuously, randomly, and exhaustively sampled. 

\textbf{Competing models.} We compare our method with state of the art {\small CVM-NET~\cite{Hu_2018_CVPR}, CVFT~\cite{shi2020optimal}, SAFA~\cite{shi2019spatial}, Polar-SAFA~\cite{shi2019spatial}, DSM~\cite{shi2020looking}}, and {\small VIGOR~}\cite{zhu2021vigor}. Among these methods, {\small DSM} is the only one that can estimate the orientation of a query camera, while others are restricted to location estimation only.  Only {\small VIGOR} considers the spatial shifts between a query camera location and its matching satellite image center, and they employ two FC layers to regress the spatial shifts.  Toker~\etal~\cite{toker2021coming} needs a matching ground-level image for each database satellite image to train their generator, which is not available in our fine-grained retrieval setting. Hence, we cannot compare to it. The above competing models were retrained (fine-tuned) on our datasets using their original metric learning procedure.

\begin{figure}[t!]
\setlength{\abovecaptionskip}{0pt}
    \setlength{\belowcaptionskip}{0pt}
    \centering
    \begin{subfigure}{\linewidth}
    \centering
    \includegraphics[width=0.45\linewidth]{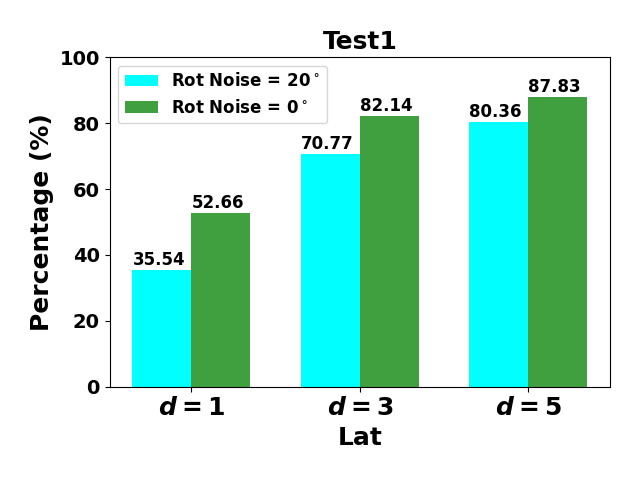}
    \includegraphics[width=0.45\linewidth]{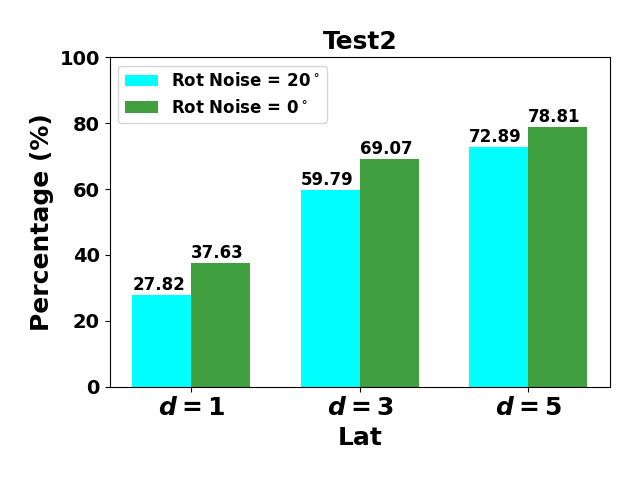}
    \caption{KITTI dataset}
    \label{subfig:KITTI}
    \end{subfigure}
    \begin{subfigure}{\linewidth}
    \centering
    \includegraphics[width=0.45\linewidth]{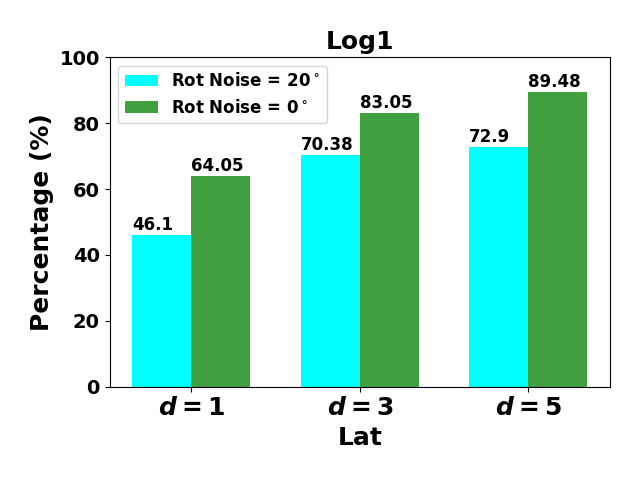}
    \includegraphics[width=0.45\linewidth]{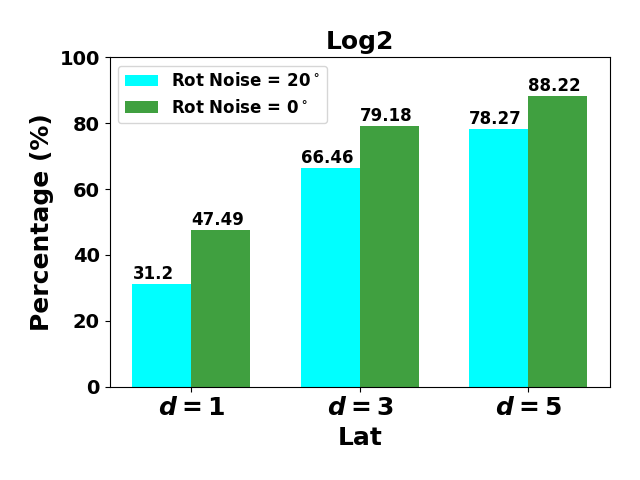}
    \caption{Ford multi-AV dataset}
    \label{subfig:Ford}
    \end{subfigure}
    \caption{Performance comparison of our method when rotation is given or unknown (rotation noise $0^\circ$ Vs. $20^\circ$).  }
    \label{fig:stastistics}
\end{figure}

\textbf{Results.}
The comparison results on {\small KITTI} and Ford multi-AV dataset are presented in Tab.~\ref{tab:sota_kitti} and Tab.~\ref{tab:sota_ford}, respectively. 
For space limits, only the results on the first two logs of the Ford multi-AV dataset are presented in the main paper. 
We provide the performance on the remaining logs of our method in the supplementary material.


From Tab.~\ref{tab:sota_kitti} and Tab.~\ref{tab:sota_ford}, it can be seen that the pure image retrieval-based methods, \ie, {\small CVM-NET~\cite{Hu_2018_CVPR}, CVFT~\cite{shi2020optimal}, SAFA~\cite{shi2019spatial}, Polar-SAFA~\cite{shi2019spatial}, DSM~\cite{shi2020looking}}, show very poor performance on the high-accuracy distance based localization. 
This is not only because the database images are discretized but also because the fine-grained partitions of a satellite image are very similar, inducing large uncertainty in cross-view image matching. 
Since {\small VIGOR} explicitly considers the relative displacement between a query camera center and its matching satellite image center, it achieves better performance compared to the pure image retrieval techniques. 
Moreover, the performance is significantly boosted by using the proposed camera pose optimization mechanism rather than a fine-grained image retrieval.

\begin{figure*}[t!]
\setlength{\abovecaptionskip}{0pt}
    \setlength{\belowcaptionskip}{0pt}
    \centering
    \adjincludegraphics[width=0.145\linewidth,trim={{\width/3.8} {\width/3.8} {\width/3.8} {\width/3.8}},clip]{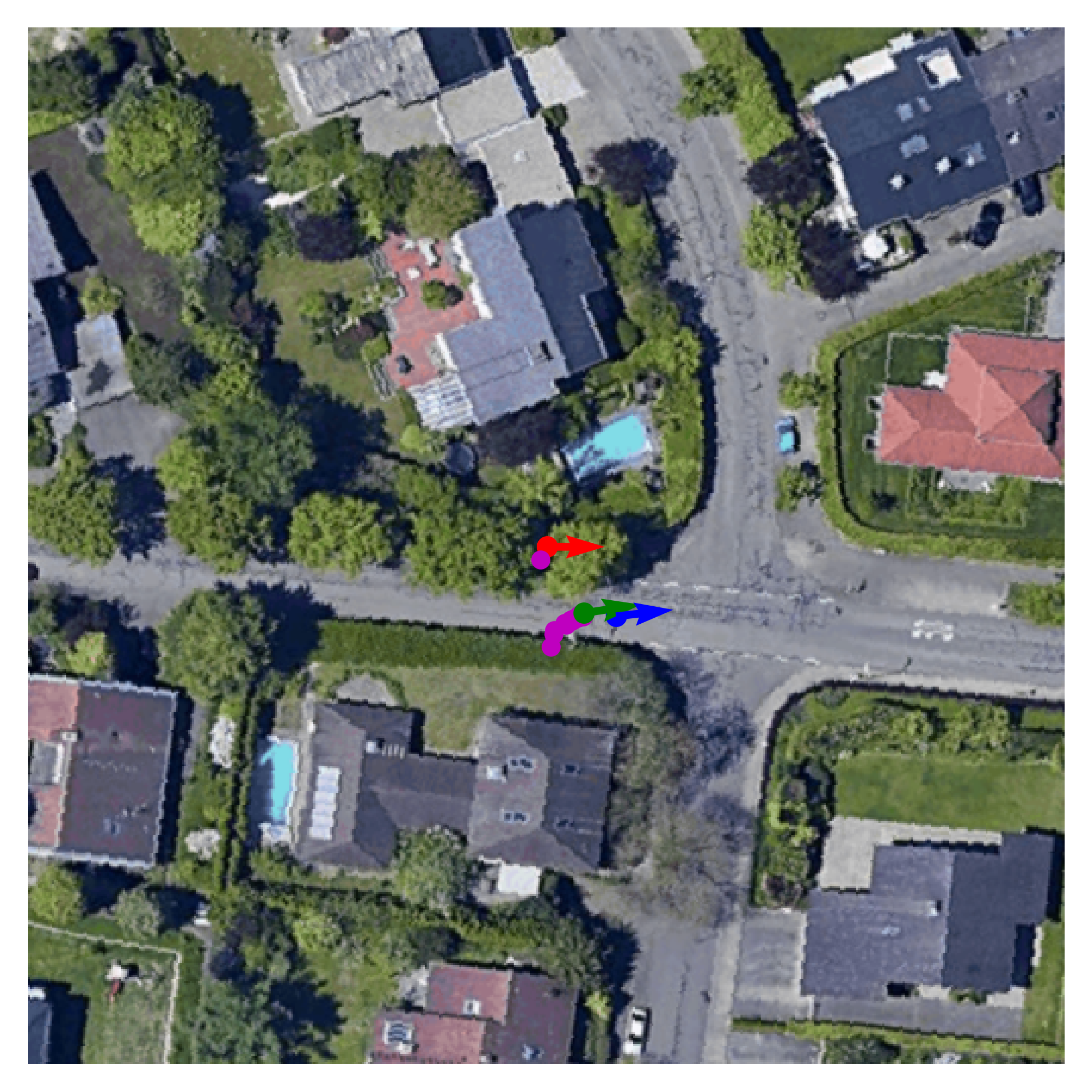} 
    \adjincludegraphics[width=0.145\linewidth,trim={{\width/3.8} {\width/3.8} {\width/3.8} {\width/3.8}},clip]{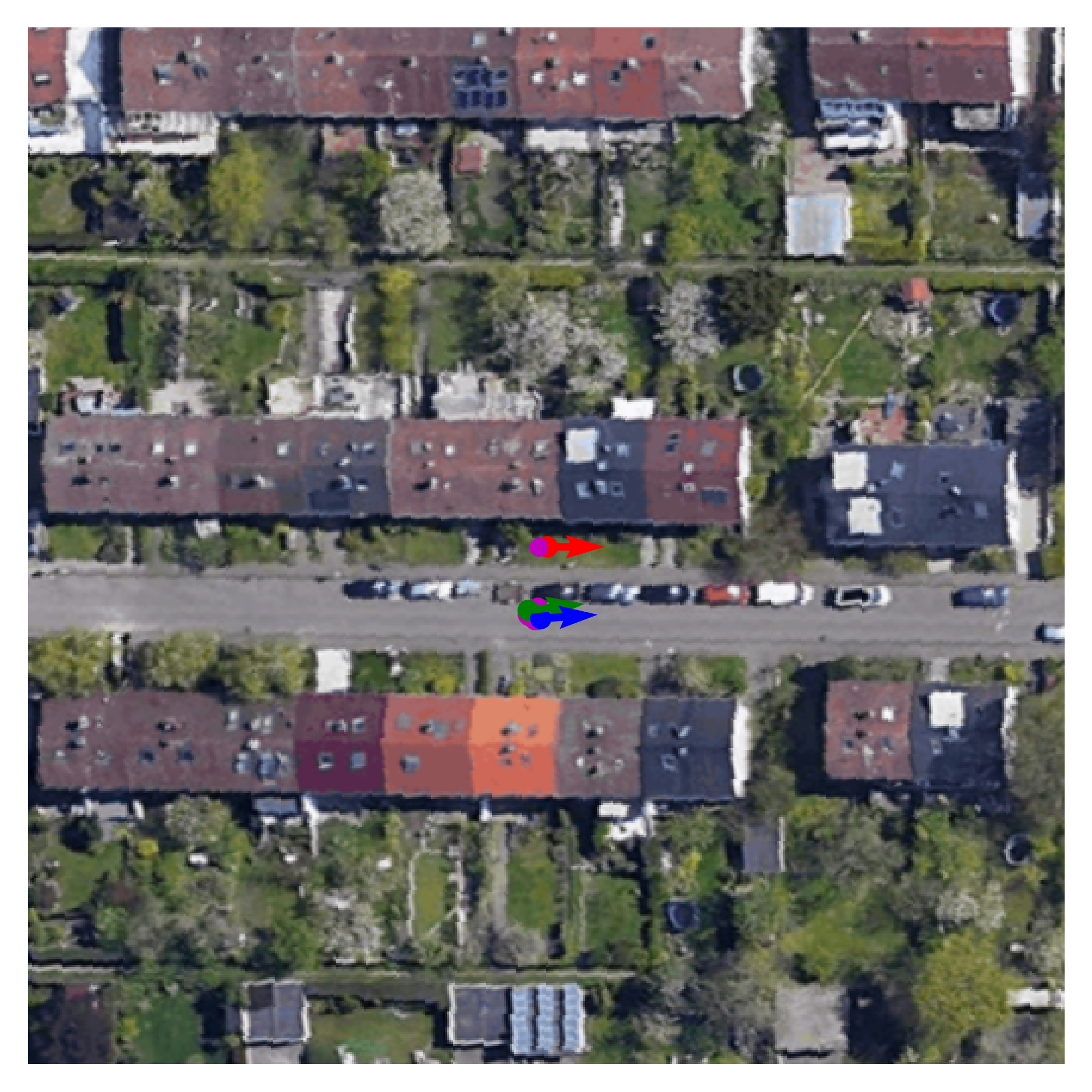} 
    \adjincludegraphics[width=0.145\linewidth,trim={{\width/3.8} {\width/3.8} {\width/3.8} {\width/3.8}},clip]{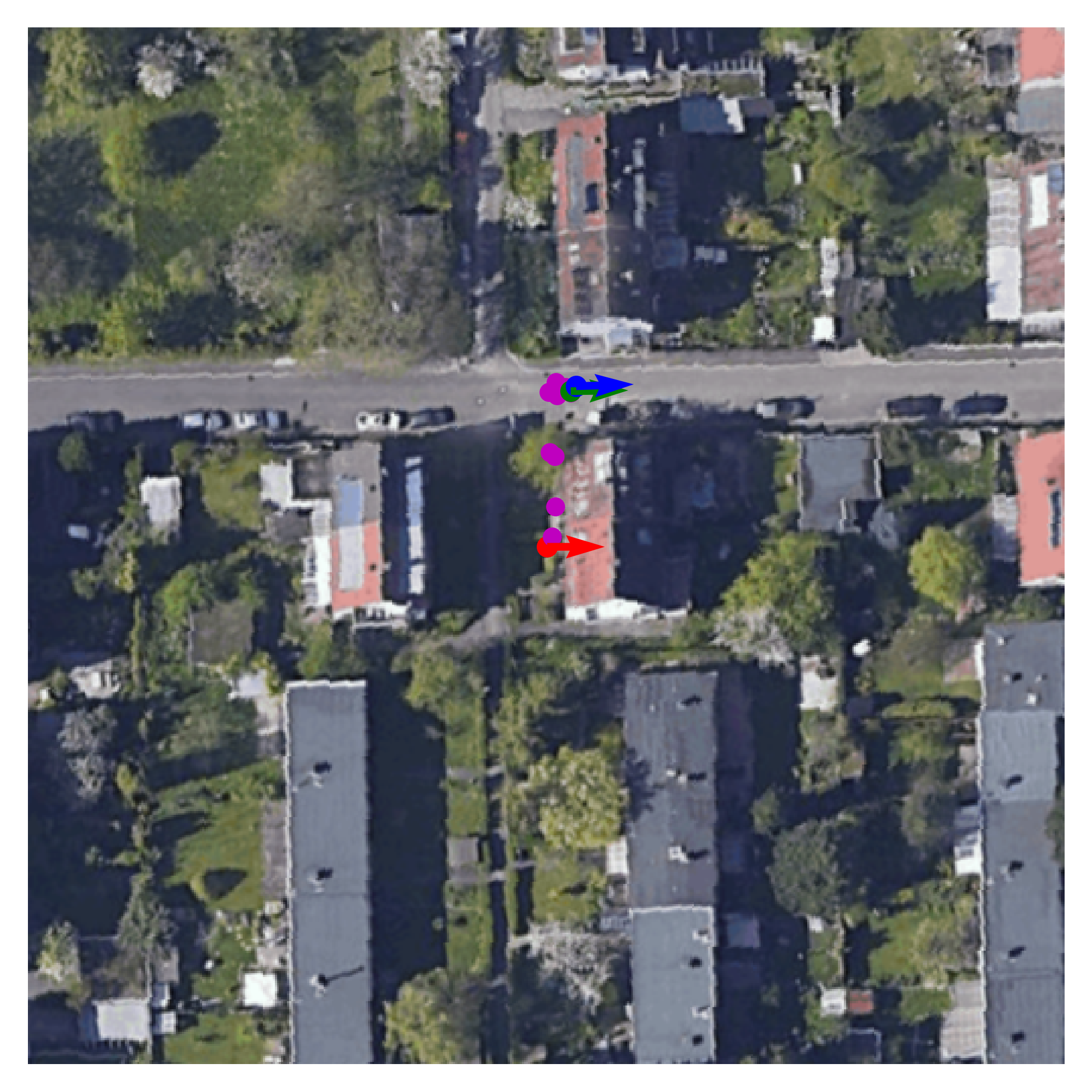} 
    \adjincludegraphics[width=0.145\linewidth,trim={{\width/3.8} {\width/3.8} {\width/3.8} {\width/3.8}},clip]{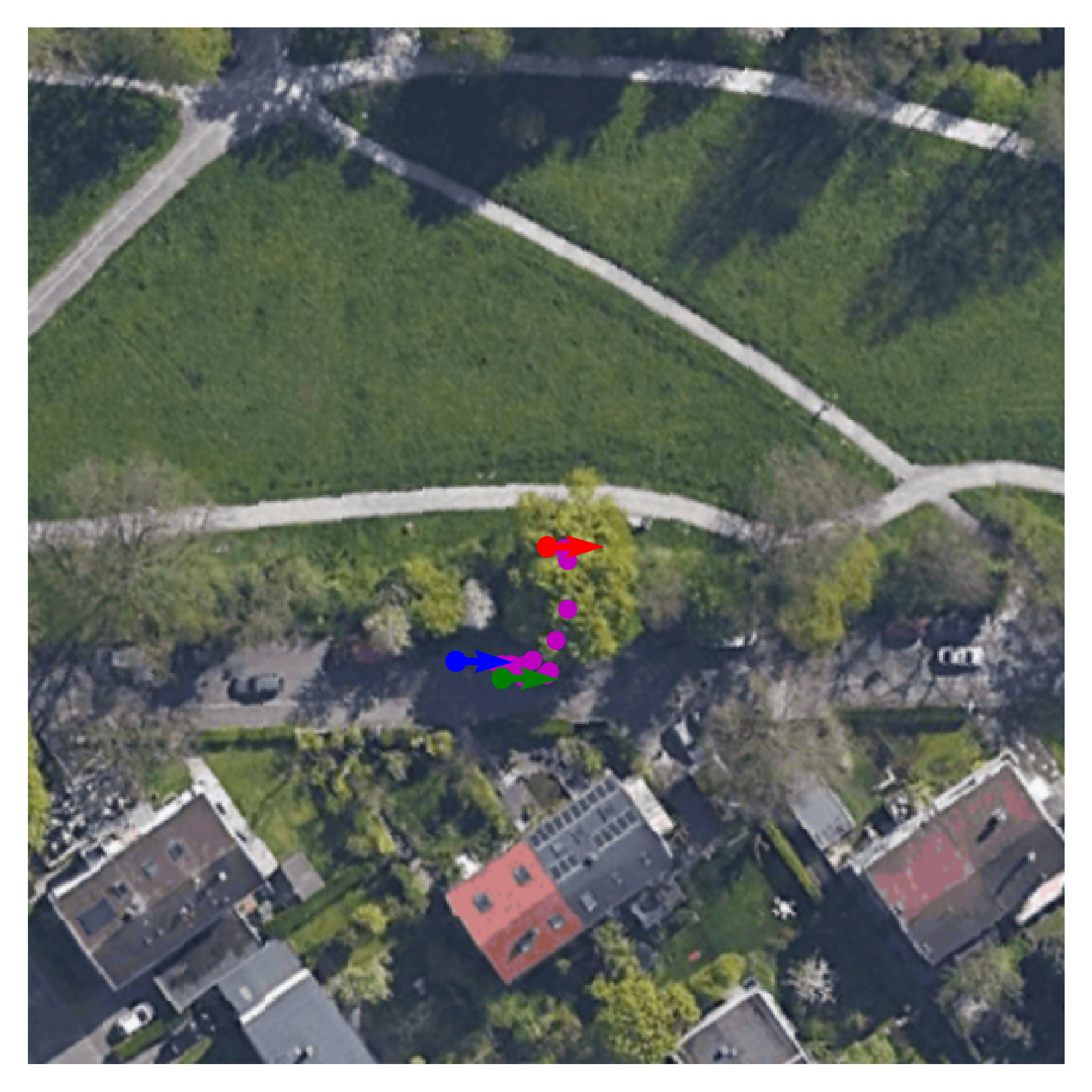} 
    \adjincludegraphics[width=0.145\linewidth,trim={{\width/3.8} {\width/3.8} {\width/3.8} {\width/3.8}},clip]{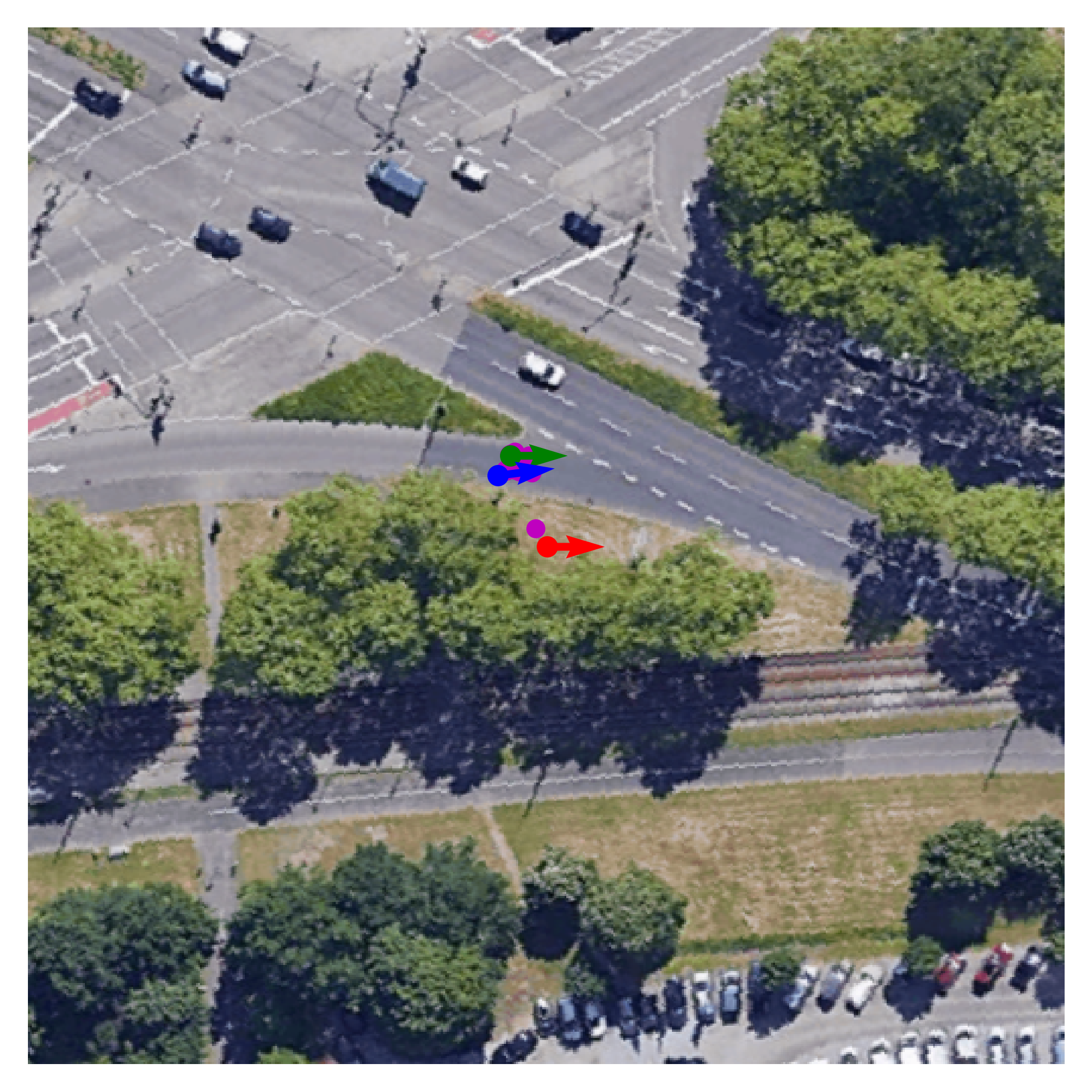} 
    \adjincludegraphics[width=0.145\linewidth,trim={{\width/3.8} {\width/3.8} {\width/3.8} {\width/3.8}},clip]{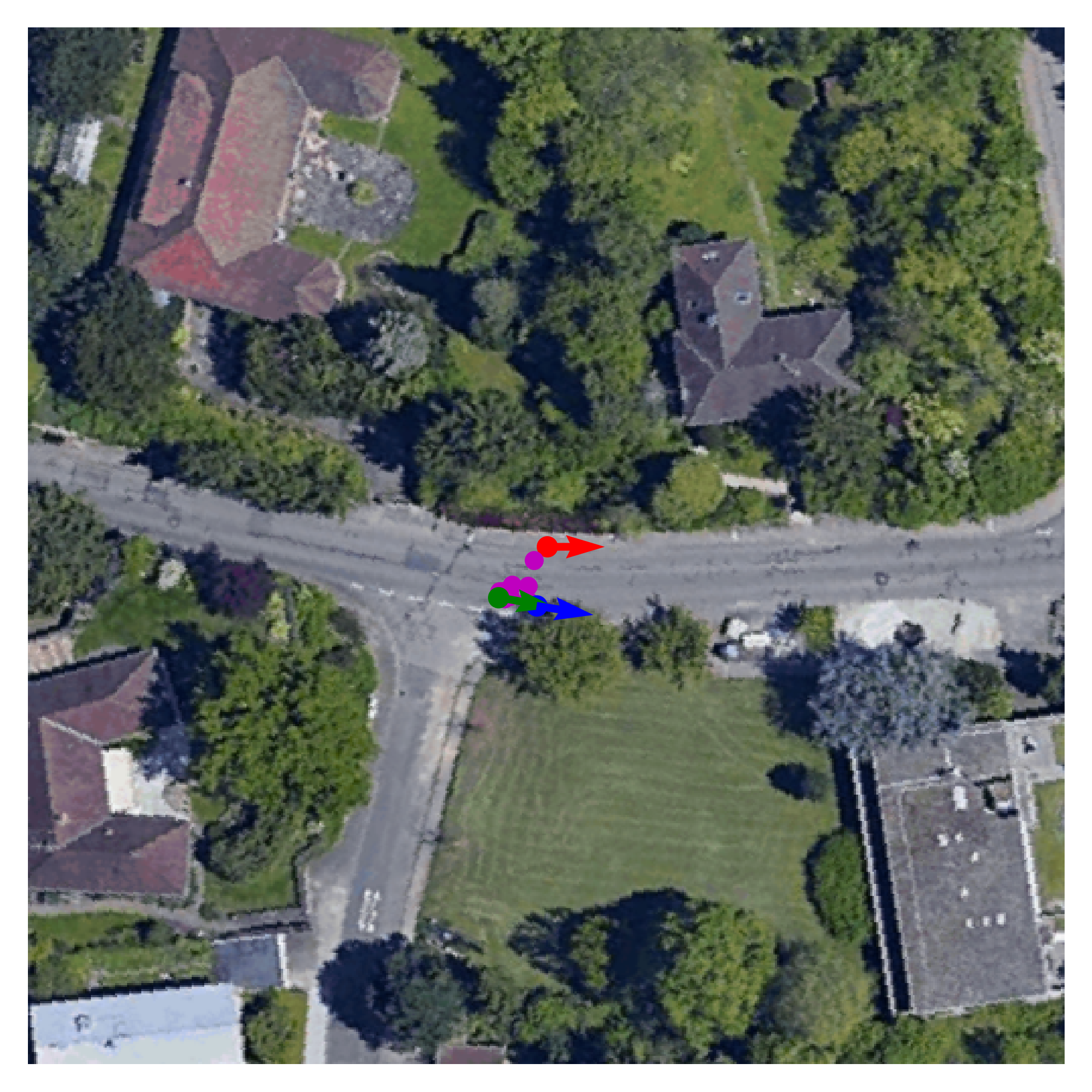}\\
    \adjincludegraphics[width=0.145\linewidth,trim={{\width/3.8} {\width/3.8} {\width/3.8} {\width/3.8}},clip]{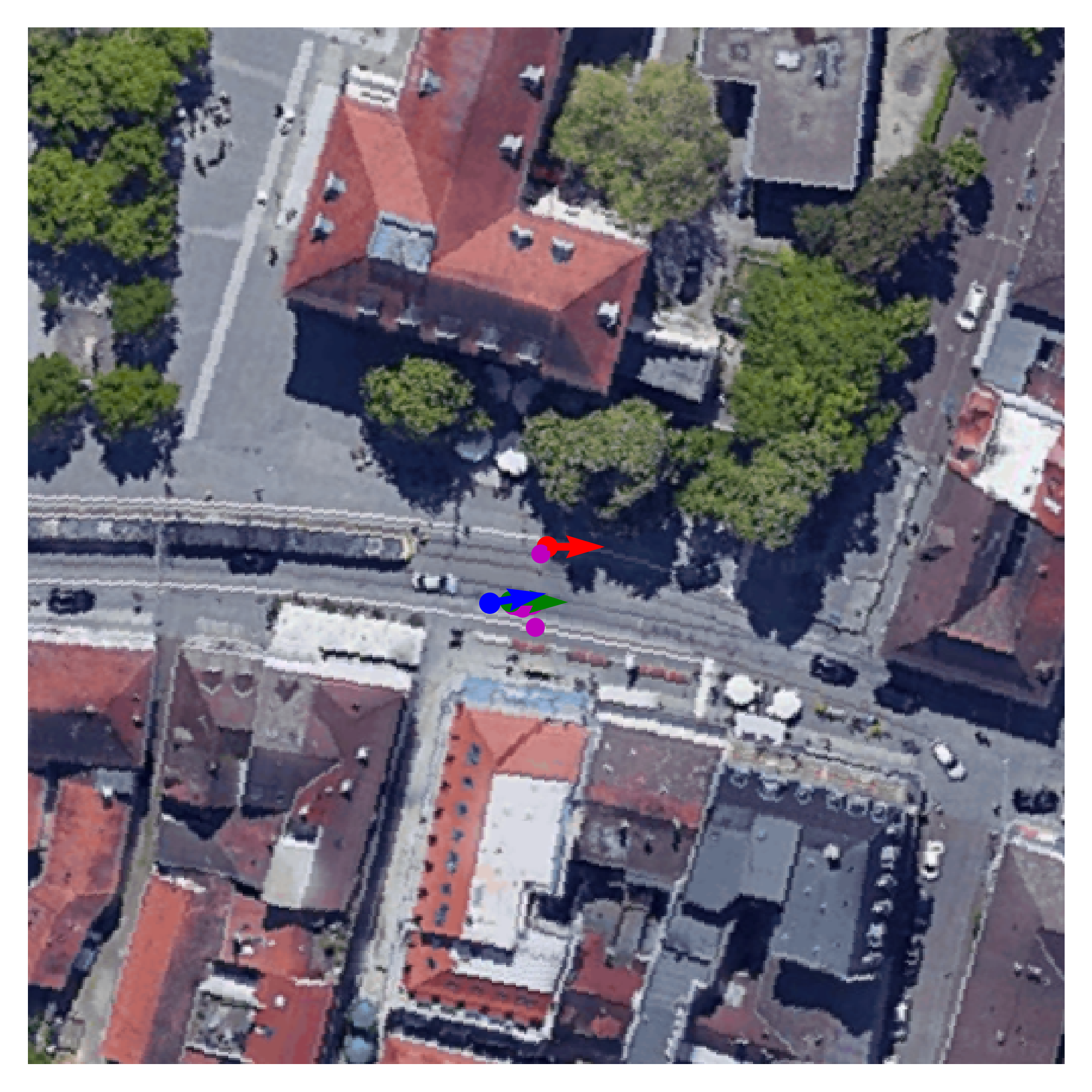} 
    \adjincludegraphics[width=0.145\linewidth,trim={{\width/3.8} {\width/3.8} {\width/3.8} {\width/3.8}},clip]{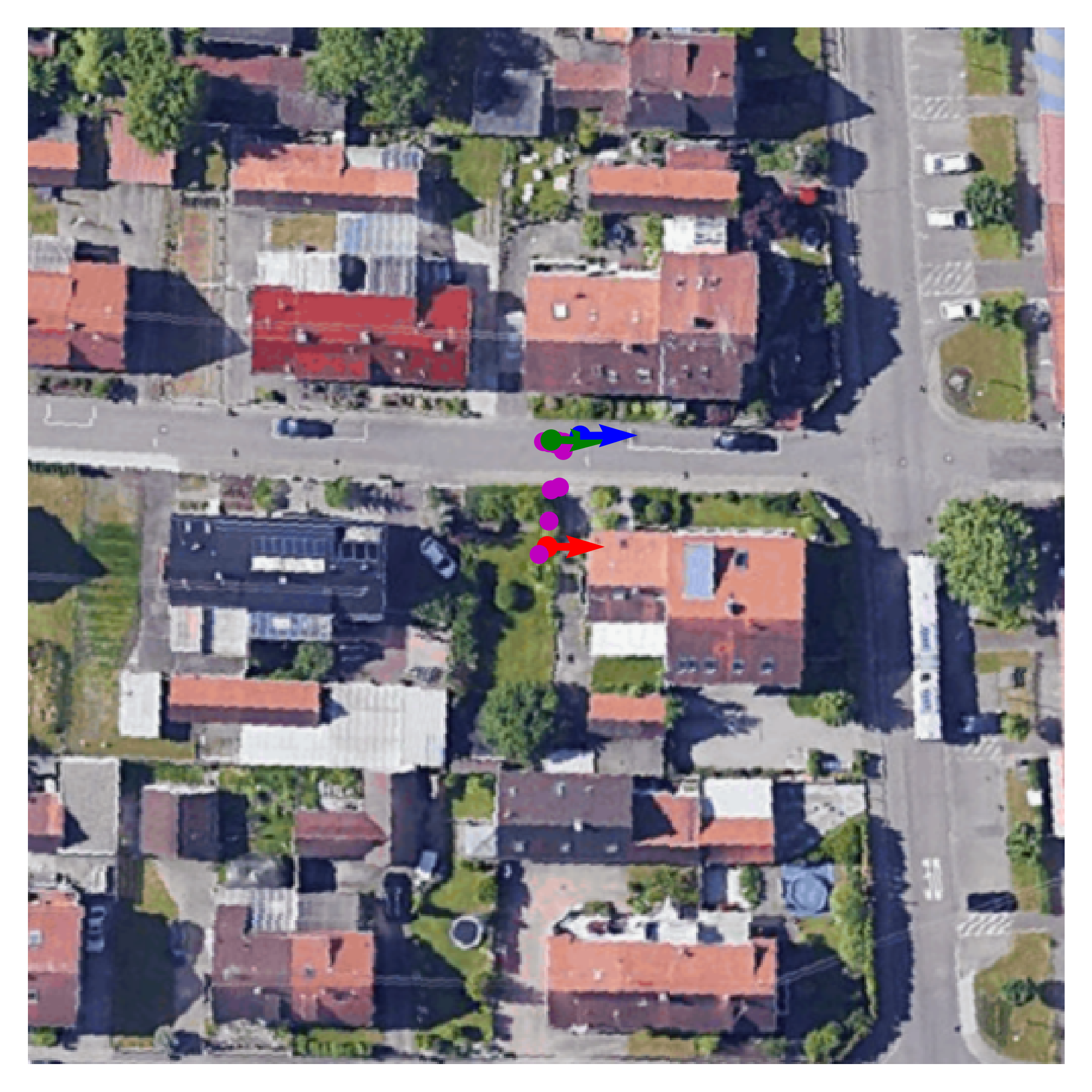} 
    \adjincludegraphics[width=0.145\linewidth,trim={{\width/3.8} {\width/3.8} {\width/3.8} {\width/3.8}},clip]{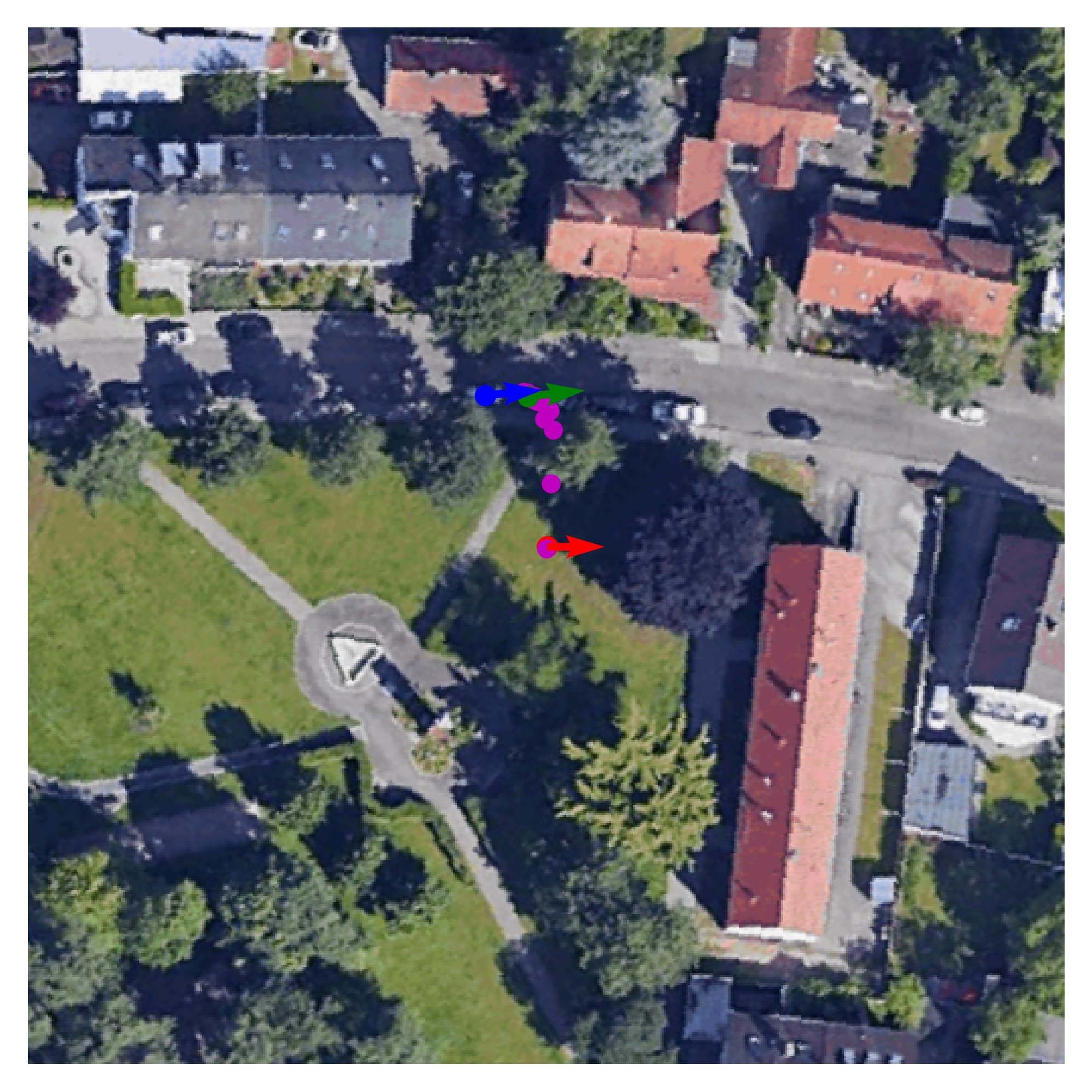} 
    \adjincludegraphics[width=0.145\linewidth,trim={{\width/3.8} {\width/3.8} {\width/3.8} {\width/3.8}},clip]{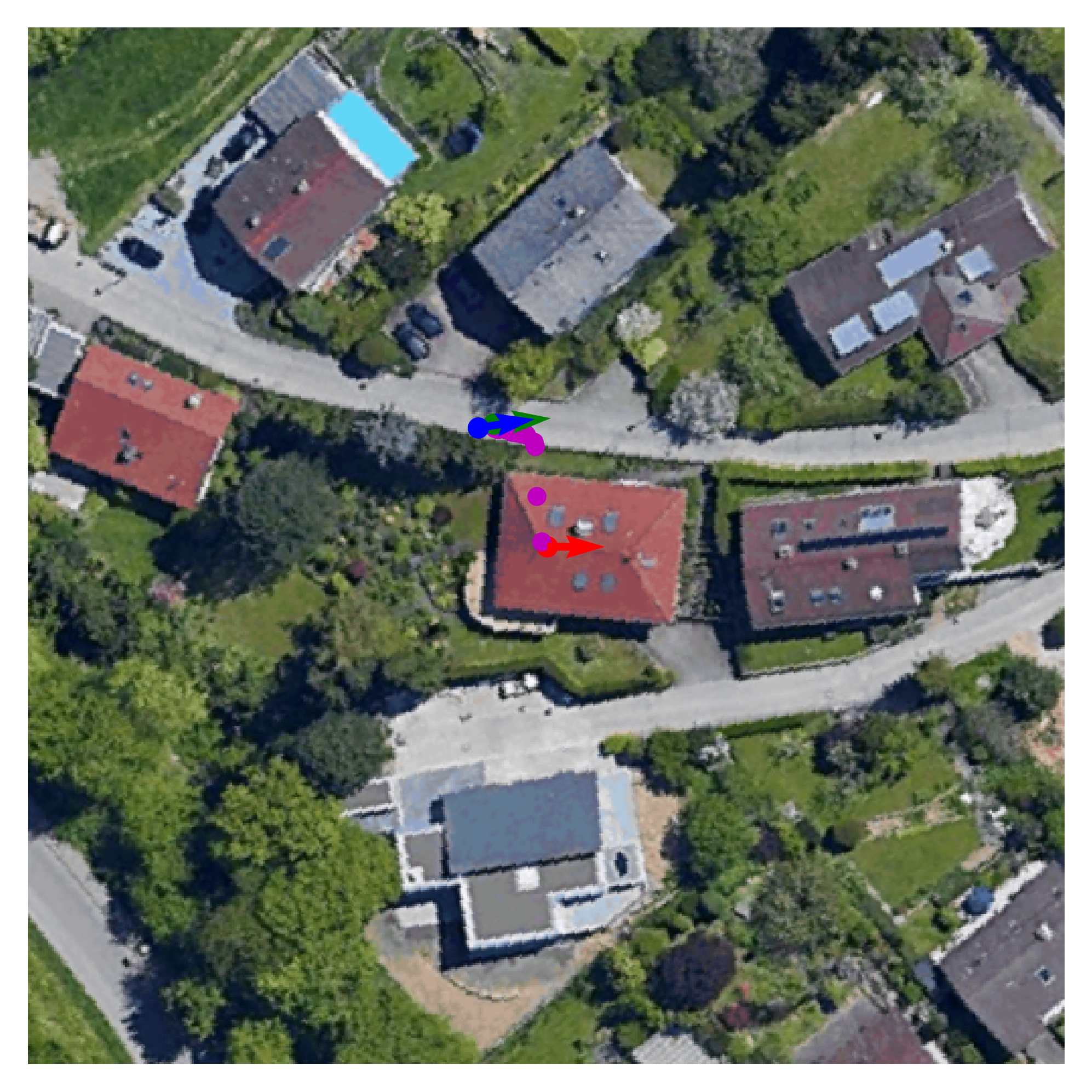} 
    \adjincludegraphics[width=0.145\linewidth,trim={{\width/3.8} {\width/3.8} {\width/3.8} {\width/3.8}},clip]{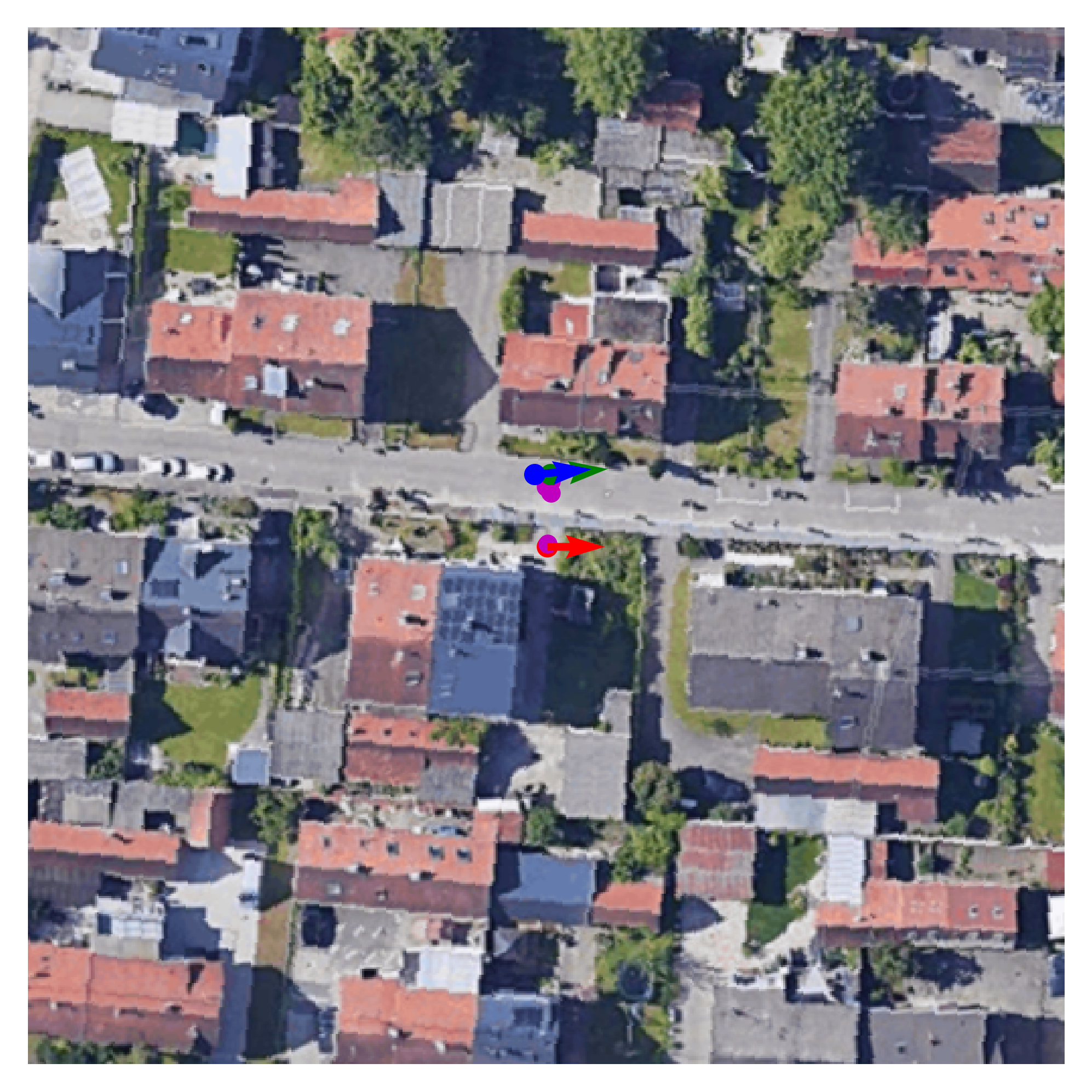} 
    \adjincludegraphics[width=0.145\linewidth,trim={{\width/3.8} {\width/3.8} {\width/3.8} {\width/3.8}},clip]{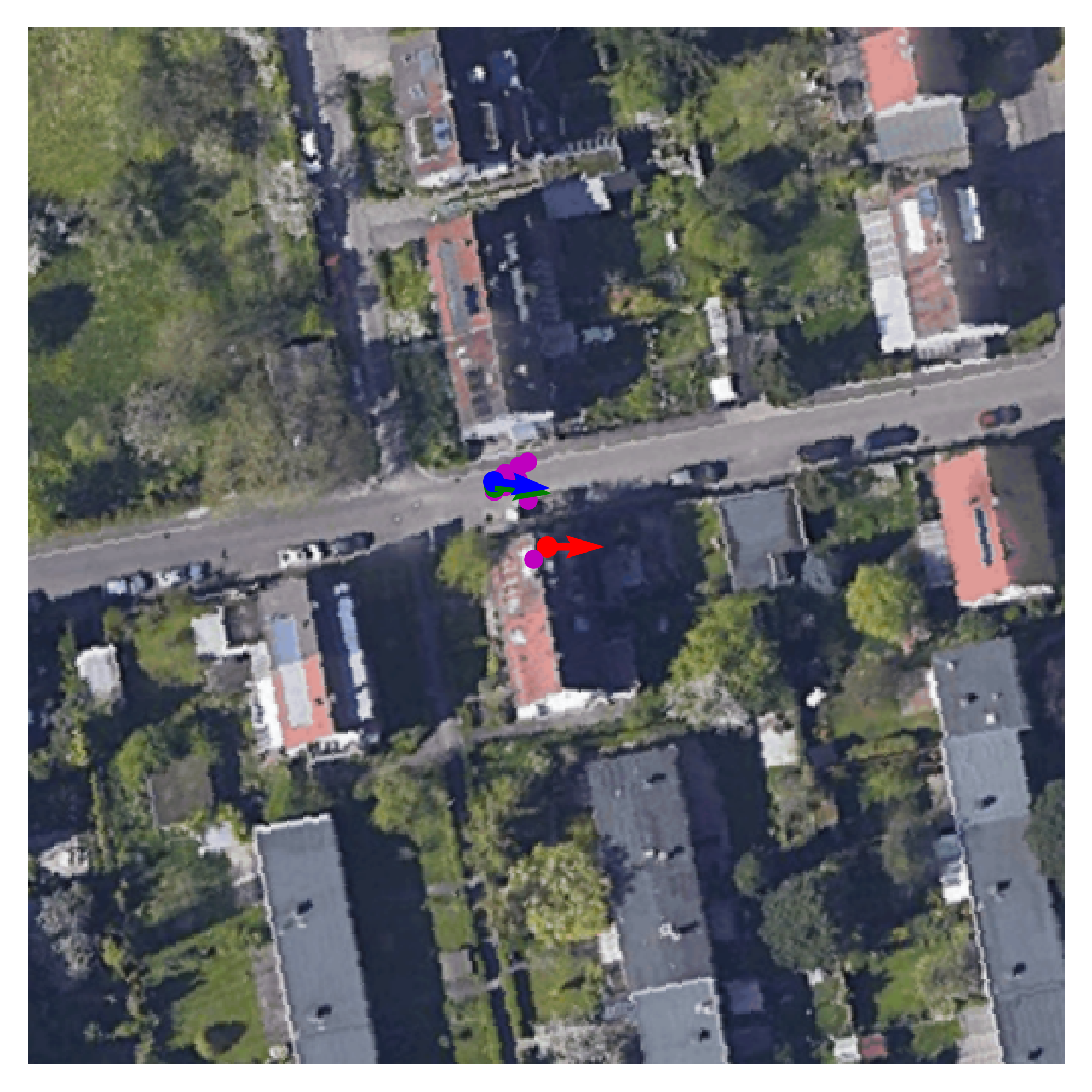}\\
     \caption{Visualization of updated poses during optimization. The arrow and point legends are the same as those in Fig.~\ref{fig:failure}. }
    \label{fig:visual_trajectory}
\end{figure*}

\begin{table*}[t!]
\setlength{\abovecaptionskip}{0pt}
\setlength{\belowcaptionskip}{0pt}
\setlength{\tabcolsep}{2pt}
\centering
\footnotesize
\caption{\small Performance comparison on different ground-and-satellite domain alignment methods on the KITTI dataset.}
\begin{tabular}{c|c|ccc|ccc|ccc|ccc|ccc|ccc}
\toprule
                      &          & \multicolumn{9}{c|}{Test1}                                                                                                                             & \multicolumn{9}{c}{Test2}                                                                                                                             \\ 
                      &          & \multicolumn{3}{c|}{Lateral}                          & \multicolumn{3}{c|}{Longitudinal}                         & \multicolumn{3}{c|}{Azimuth}                      & \multicolumn{3}{c|}{Lateral}                          & \multicolumn{3}{c|}{Longitudinal}                         & \multicolumn{3}{c}{Azimuth}                      \\
                      &          & $d=1$          & $d=3$          & $d=5$          & $d=1$         & $d=3$          & $d=5$          & $\theta=1$     & $\theta=3$     & $\theta=5$     & $d=1$          & $d=3$          & $d=5$          & $d=1$         & $d=3$          & $d=5$          & $\theta=1$     & $\theta=3$     & $\theta=5$     \\ \midrule
\multirow{2}{*}{G2SP} & NN       & 20.30          & 53.25          & 72.12          & 4.93          & 15.08          & 25.31          & \textbf{21.65} & \textbf{54.44} & \textbf{71.88} & 17.01          & 46.12          & 64.41          & 5.16          & 15.18          & 25.31          & \textbf{20.66} & \textbf{51.45} & { 70.03}    \\
                      & H        & { 27.72}    & { 59.98}    & { 71.91}    & \textbf{5.75} & \textbf{16.80} & \textbf{26.13} & 18.13          & 48.77          & 69.26          & { 25.32}    & { 54.63}    & { 64.74}    & { 4.99}    & { 15.61}    & { 26.31}    & 17.37          & 46.57          & 67.70          \\\midrule
\multirow{2}{*}{S2GP} & Polar    & 18.98          & 45.93          & 55.79          & 5.14          & 14.95          & 24.99          & 13.31          & 39.25          & 61.20          & 11.27          & 40.51          & 53.62          & 4.87          & 14.73          & 25.19          & 13.78          & 39.68          & 62.03          \\
                      & H (\textbf{Ours}) & \textbf{35.54} & \textbf{70.77} & \textbf{80.36} & { 5.22}    & { 15.88}    & \textbf{26.13} & { 19.64}    & { 51.76}    & { 71.72}    & \textbf{27.82} & \textbf{59.79} & \textbf{72.89} & \textbf{5.75} & \textbf{16.36} & \textbf{26.48} & { 18.42}    & { 49.72}    & \textbf{71.00} \\ \bottomrule
\end{tabular}
\label{tab:projection}
\end{table*}

\textbf{Visualization.}
As expected, the performance of all the methods on longitudinal direction is worse than that on the lateral direction. 
We give some visual examples of such ambiguities along the longitudinal direction in Fig.~\ref{fig:failure}. Despite this ambiguity, our method is still able to refine lateral poses effectively.  Fig.~\ref{fig:visual_trajectory} gives additional visualizations of the intermediate poses by our method when the scenes are diverse at different regions. 

\textbf{Known orientation.}
In general, the azimuth rotation of a query camera can be easily obtained from a compass, and the rotation estimation accuracy by SLAM and VO methods is usually accurate. Hence, we test our method when orientation information is given.  Fig.~\ref{fig:stastistics} reports the test results. It can be seen that the performances of our method are consistently improved.

\subsection{Method Analysis}
\subsubsection{Effectiveness of GP geometry projection}
\textbf{S2GP Vs. G2SP.}
Compared to satellite images, ground-view images have a larger resolution of scene objects. 
A small change in the camera pose will be magnified on the appearance changes of ground-view images. 
In contrast, the corresponding appearance changes in the overhead view are smaller. 
This sensitiveness of ground-view observations to camera pose changes is a desired property.
It contributes to a higher accuracy of estimated poses. 
Hence, we conduct satellite-to-ground projection (S2GP), instead of ground-to-satellite projection (G2SP), in our geometry projection module. 
Below, we compare the performance of the two projection methods. 
Since we use the homography of the ground plane in the projection, we label them by ``H'' in Tab.~\ref{tab:projection}. 
As expected, the performance of S2GP is superior to that of G2SP.

\textbf{Homography Vs. Polar transform.}
Apart from homography, polar transform was adopted in the literature for bridging the cross-view domain gap \cite{shi2019spatial, shi2020looking, toker2021coming}.  We intend to compare our method with the polar transform in S2GP, labeled as ``Polar''. From the results in Tab.~\ref{tab:projection}, it can be seen the polar transform performs worse than the homography. This is possiblly because the polar transform only accounts for ground-level panoramas rather than images captured by a ground-level pin-hole camera.

\textbf{Explicit geometry projection Vs. Implicit network.}
In contrast to using an explicit geometry transform, we also tested whether a simple neural network can learn an implicit geometric mapping for the same purpose, denoted as ``NN''. 

Here, the NN ablation cannot be conducted in the S2GP direction, because the S2GP is a whole-to-part mapping and will loss much information in the projection. 
When the initial pose is significantly different with the real one, the synthesized ground-view feature map at the initial pose can be totally different with the real observed one. 
In this case, it is impossible to make the synthetic and real feature maps aligned by simple image/feature level rotation and translation. 
Suppose we let NN regenerate a ground feature map for the original satellite image at each pose update. The Jacobian of network parameters will be required at each LM optimization step, which takes a significant amount of GPU memory and is far beyond the capacity of existing 12/24G GPUs.
In contrast, our geometry-based S2GP does not involve any network parameters in the LM optimization and thus is feasible.
Therefore, our NN ablation is conducted in the G2SP direction. 
We employ a network that takes a ground image as input and outputs a synthetic satellite feature map. 
After that, the LM directly rotates and translates the synthetic feature map to register it with its real counterpart without any NN regeneration. 

As shown in the first row of Tab.~\ref{tab:projection}, the results are not favorable. 
Although ``NN'' achieves slightly better performance than our geometry projection (both S2GP and G2SP) on rotation estimation, its capability in handling translation optimization is rather limited. Our reflection on this is that whenever explicit and principle geometric knowledge about the problem at hand is known and can be used, one should use it instead of black-box neural network implementations.

\begin{table*}[t!]
\setlength{\abovecaptionskip}{0pt}
\setlength{\belowcaptionskip}{0pt}
\setlength{\tabcolsep}{2.5pt}
\centering
\footnotesize
\caption{\small Performance comparison by using different optimizers on the KITTI dataset.}
\begin{tabular}{c|ccc|ccc|ccc|ccc|ccc|ccc}
\toprule
         & \multicolumn{9}{c|}{Test1}                                                                & \multicolumn{9}{c}{Test2}                                                                \\
         & \multicolumn{3}{c|}{Lateral} & \multicolumn{3}{c|}{Longitudinal} & \multicolumn{3}{c|}{Azimuth}          & \multicolumn{3}{c|}{Lateral} & \multicolumn{3}{c|}{Longitudinal} & \multicolumn{3}{c}{Azimuth}          \\
         & $d=1$  & $d=3$  & $d=5$ & $d=1$  & $d=3$  & $d=5$ & $\theta=1$ & $\theta=3$ & $\theta=5$ & $d=1$  & $d=3$  & $d=5$ & $d=1$  & $d=3$  & $d=5$ & $\theta=1$ & $\theta=3$ & $\theta=5$ \\ \midrule
SGD      & 16.86       & 39.60          & 51.15          & 4.72          & { 15.29}    & 25.39          & 10.05          & 30.37          & 49.80          & 16.06          & 38.41          & 50.29          & 5.00          & 15.34          & 25.70          & 9.98           & 30.03          & 50.13          \\
ADAM     & 7.13        & 21.15          & 32.97          & { 4.96}    & 15.13          & { 25.63}    & 10.36          & 30.32          & 50.49          & 7.33           & 21.36          & 33.52          & { 5.64}    & 15.38          & { 26.00}    & 10.28          & 30.81          & 50.91          \\
Net      & { 27.14} & { 58.28}    & { 71.91}    & 4.53          & 15.19          & 25.36          & \textbf{45.56} & \textbf{93.19} & \textbf{99.76} & { 20.26}    & { 53.94}    & { 67.42}    & 5.40          & { 15.82}    & 25.58          & \textbf{42.03} & \textbf{92.32} & \textbf{99.81} \\
LM (\textbf{Ours}) & \textbf{35.54}       & \textbf{70.77} & \textbf{80.36} & \textbf{5.22} & \textbf{15.88} & \textbf{26.13} & { 19.64}    & { 51.76}    & { 71.72}    & \textbf{27.82} & \textbf{59.79} & \textbf{72.89} & \textbf{5.75} & \textbf{16.36} & \textbf{26.48} & { 18.42}    & { 49.72}    & { 71.00}    \\  \bottomrule
\end{tabular}
\label{tab:optimizer}
\end{table*}

\subsubsection{Superiority of LM optimization}

\textbf{LM Vs. SGD and ADAM.}
Stochastic Gradient Decent (SGD) and ADAM are widely-used optimization methods in neural network training. 
They have also been demonstrated as effective in many recent Nerf-based methods for scene-specific camera pose estimation~\cite{lin2021barf, yen2020inerf, wang2021nerf}. 
Hence, we compare the LM algorithm employed in this paper with the first-order SGD and Adam on the ground-to-satellite camera pose optimization. 

As shown in Tab.~\ref{tab:optimizer}, it can be seen that the LM optimization performs significantly better than SGD and ADAM.
This is because the adaptive second-order LM optimization, as a variant of Gaussian Newton, is essentially guaranteed to find at least one local minimum of a cost function. 
In contrast, SGD suffers the usual zigzagging behavior and is thus very slow to converge. 
Although ADAM is often better than SGD on neural network training, we found that it performs the worst among the comparison optimizers in this ground-to-satellite pose optimization. 

\textbf{LM Vs. Network-based optimizer.}
Using a network to mimic an optimizer has also been investigated in various tasks, for example, optical flow~\cite{teed2020raft}, view synthesis~\cite{flynn2019deepview}, and object pose estimation~\cite{li2018deepim}. 
Hence, we also compare with a network-based optimizer, denoted as ``Net'' in Tab.~\ref{tab:optimizer}. 
The network-based optimizer is composed of a set of convolutional layers and fully connected layers. 
We also used convolutional GRU and LSTM to construct the network-based optimizer, but we found no significant difference.

Interestingly, we found that the network-based optimizer performs significantly better on rotation optimization while achieving inferior translation optimization performance than LM. 
This is probably because regular CNNs are not inherently rotation-invariant. A slight rotation change in the input signal will lead to a big difference in the CNN feature maps. Such amplified changes give the CNN-based optimizer more power to search for better rotations.
On the other hand, CNNs are translational invariant/equivariant. A small change in translation can be absorbed by higher-level CNN features, adversely affecting the accuracy of translation estimation.  These observations have been confirmed in our experiments as seen in Tab.~\ref{tab:projection}. Using a network for the ground-and-satellite domain mapping performs better than the geometry-guided method on rotation optimization, while worse on translation optimization. It deserves further exploration to better combine the advantages of principled theories (\eg, geometry and LM optimization) with data-driven approaches.

\section{Conclusion}
In this paper, we have proposed a novel method for accurate camera localization using ground-to-satellite cross-view images.  This new method represents a  departure from the conventional wisdom of image retrieval-based localization.
The key challenge lies in properly handling the vast domain gap between the cross-view setting (satellite Vs. ground-view). To this end, we have devised a Geometry Projection module that aligns the two-view features in the ground domain. A principled LM optimization algorithm is employed to optimize the relative camera poses progressively in an end-to-end manner.  

Although this work was motivated by the poor accuracy of conventional image retrieval-based localization, we do not intend to replace the image retrieval-based localization technique. Instead, city-scale place retrieval can provide an initial estimate for a query camera. Our method then refines this pose estimate to higher accuracy.

Our ground-to-satellite pose optimization method can also help the conventional SLAM and visual odometry methods for camera tracking as a novel mechanism for ``loop closure'' in SLAM.  In particular, we remark that combining our method with a VO pipeline may resolve the longitudinal ambiguity issue, achieving all-around highly accurate vehicle localization. Furthermore, we expect the overall performance of our method will be further improved when depth information is available, for example, provided by stereo images or Lidar points. This is left as a future work.

\section{ Acknowledgments}
This research is funded in part by ARC-Discovery grants (DP 190102261 and DP220100800) and a gift from Baidu RAL to HL.
The first author is a China Scholarship Council (CSC)-funded PhD student to ANU. 
We thank all anonymous reviewers and ACs for their constructive suggestions.

{\small
\bibliographystyle{ieee_fullname}
\bibliography{egbib}
}

\newpage
\onecolumn
\appendix
\appendixpage

\section{Training and Testing Splits of the KITTI and the Ford Multi-AV dataset} 
\begin{figure}[ht]
\setlength{\abovecaptionskip}{5pt}
\setlength{\belowcaptionskip}{0pt}
    \centering
    \begin{subfigure}{0.48\linewidth}
    \centering
    \includegraphics[width=0.66\linewidth, height = 0.3\linewidth]{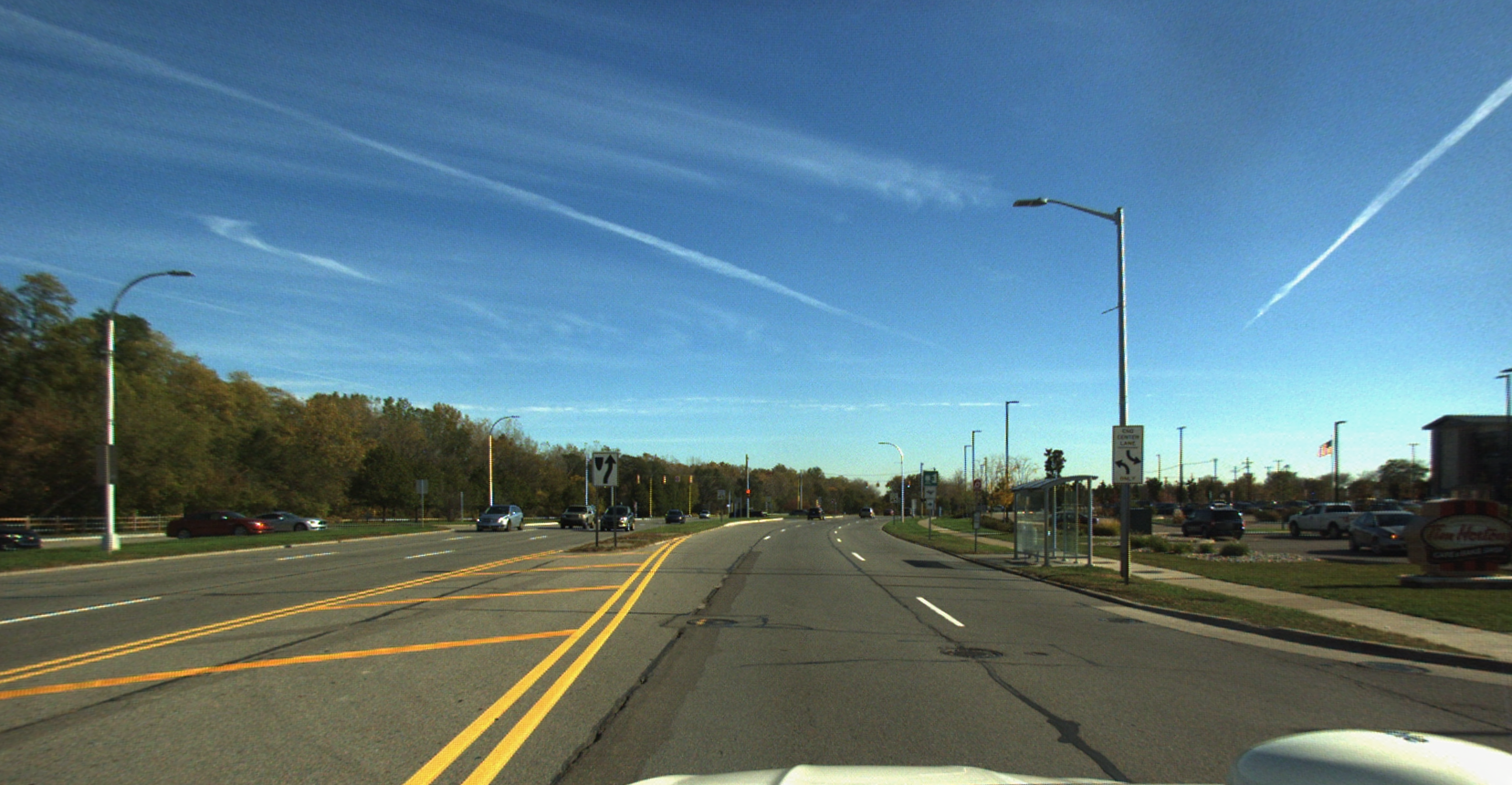}
    \includegraphics[width=0.3\linewidth]{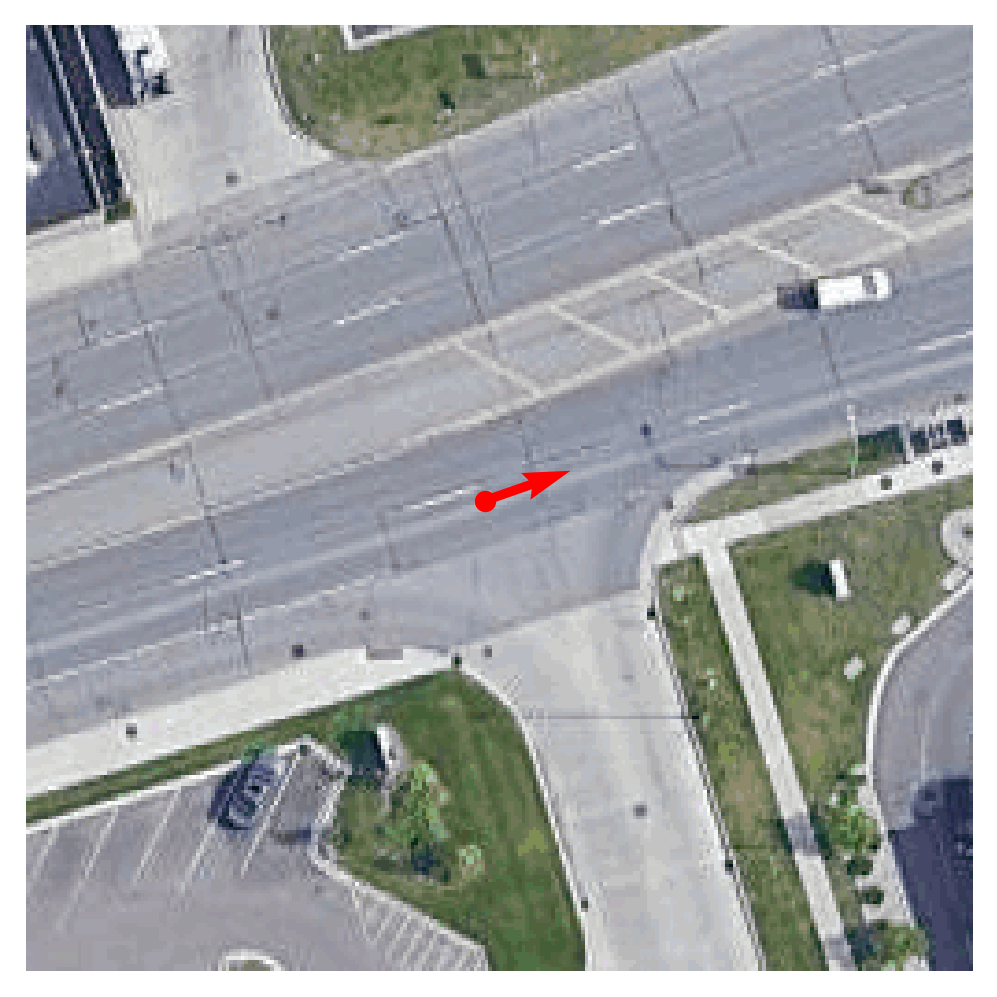}
    \caption{The camera is on the left road, as indicated by the ground image, while its position computed from the GPS tag pinpoints it is near the lane line between the left and right roads, as shown in the satellite image. }
    \end{subfigure}
    \hspace{0.5em}
    \begin{subfigure}{0.48\linewidth}
    \centering
    \includegraphics[width=0.66\linewidth, height = 0.3\linewidth]{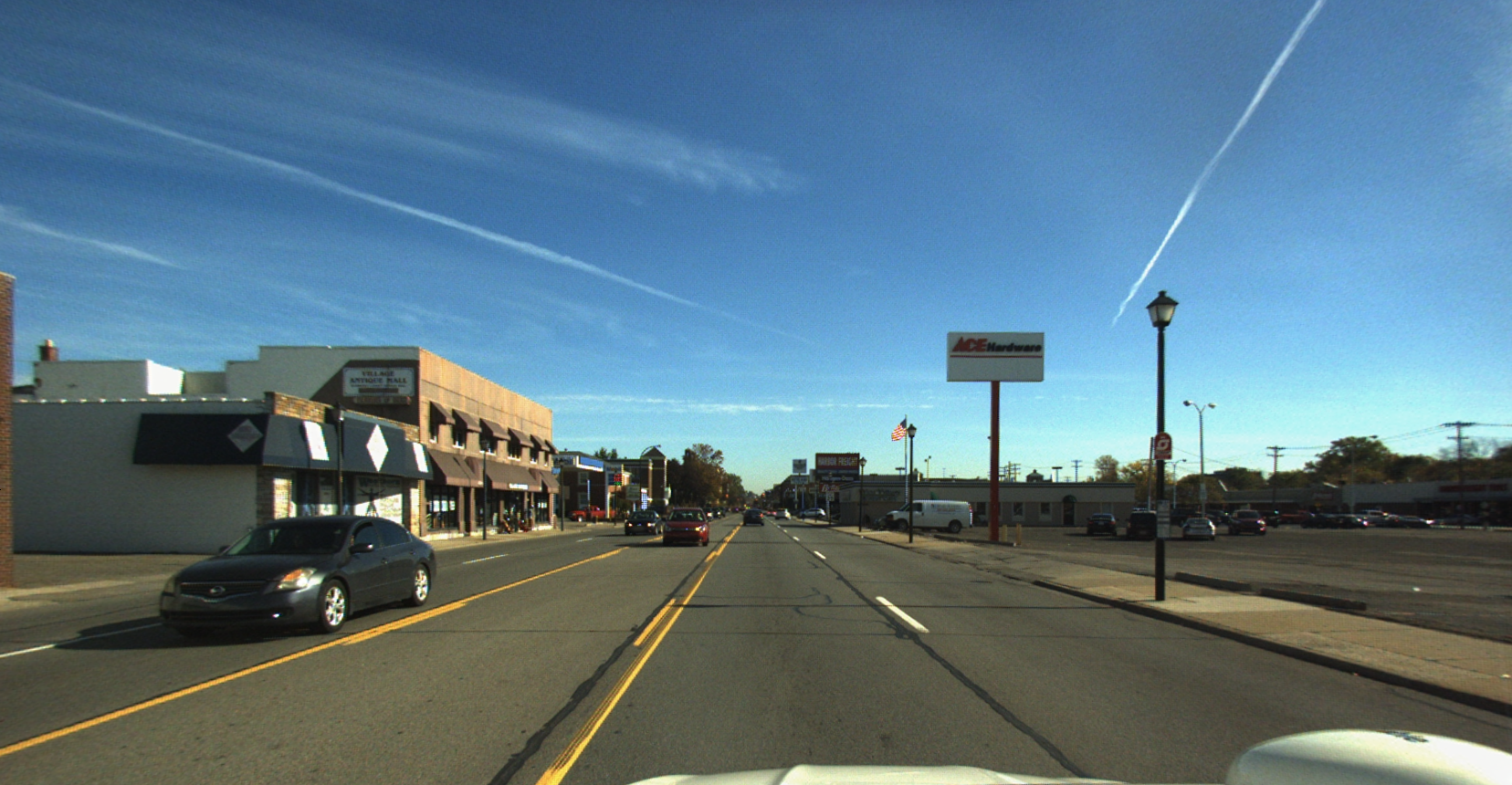}
    \includegraphics[width=0.3\linewidth]{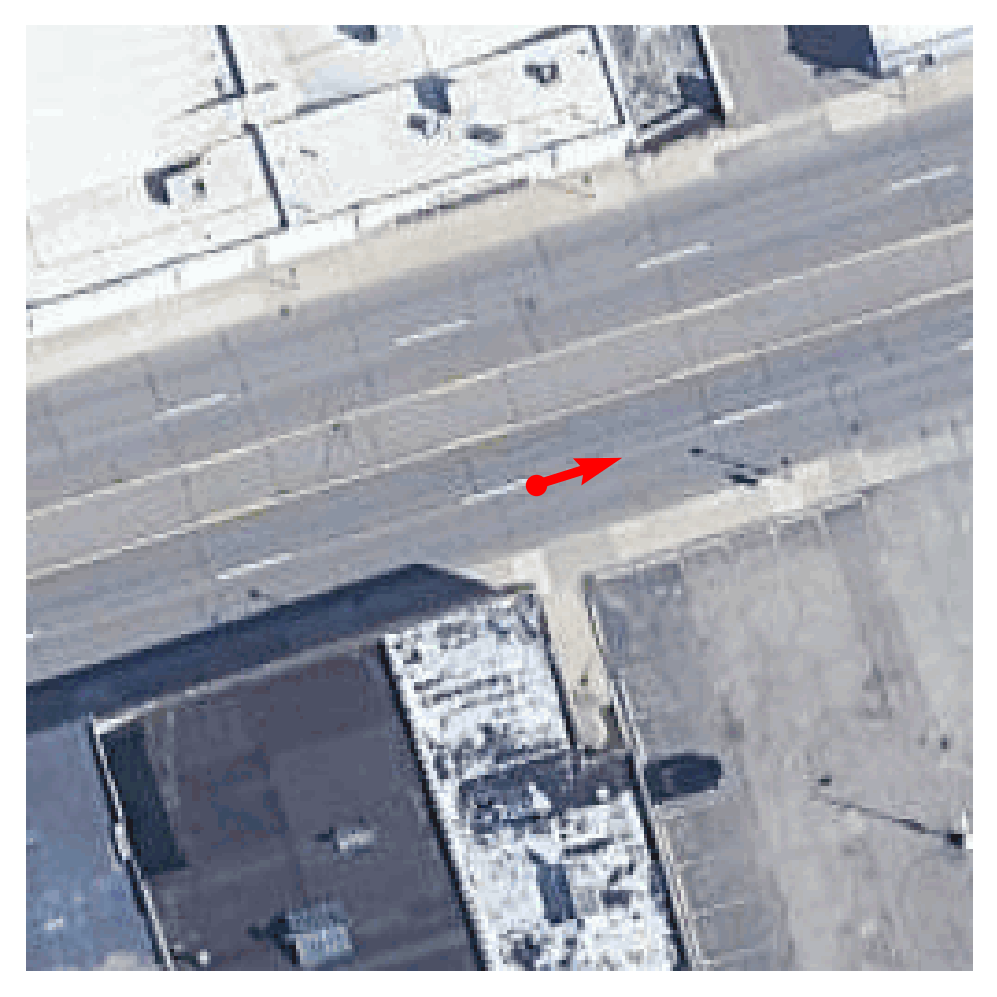}
    \caption{The camera is on the second road from the right, as indicated by the ground image, while its position computed from the GPS tag pinpoints it is near the lane line between the first and second roads from the right, as shown in the satellite image. }
    \end{subfigure}
    \begin{subfigure}{0.48\linewidth}
    \centering
    \includegraphics[width=0.66\linewidth, height = 0.3\linewidth]{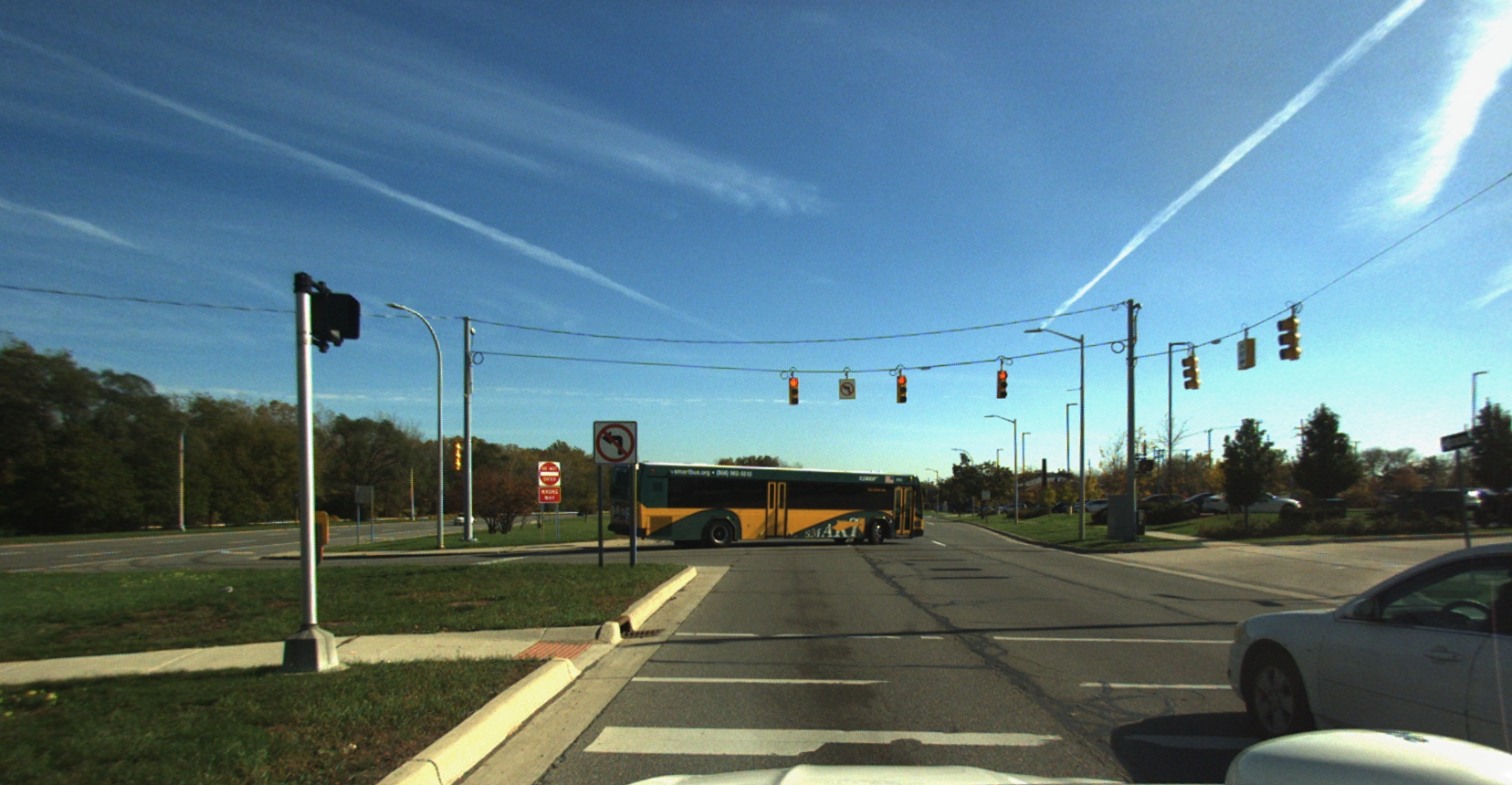}
    \includegraphics[width=0.3\linewidth]{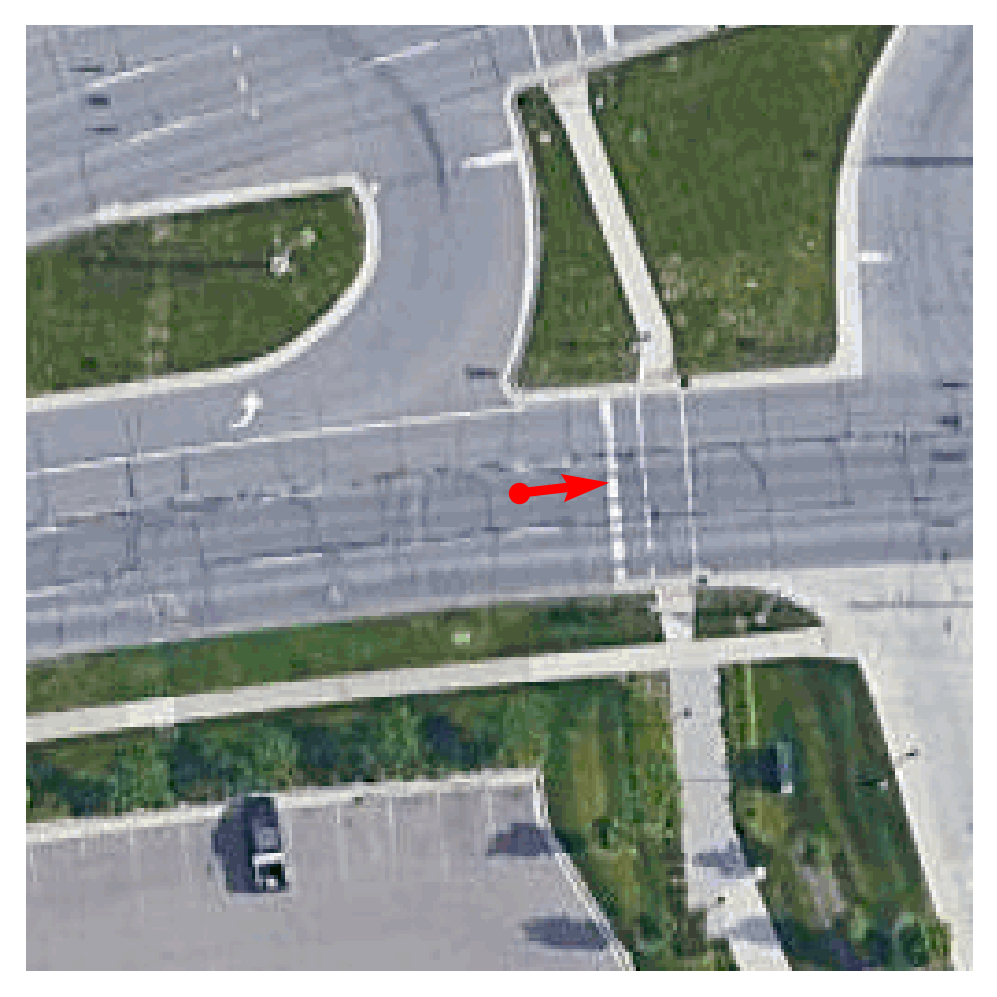}
    \caption{The camera is on the left of the main road, as indicated by the ground image, while its position computed from the GPS tag pinpoints it is in the middle of the main road, as shown in the satellite image. }
    \end{subfigure}
    \hspace{0.5em}
    \begin{subfigure}{0.48\linewidth}
    \centering
    \includegraphics[width=0.66\linewidth, height = 0.3\linewidth]{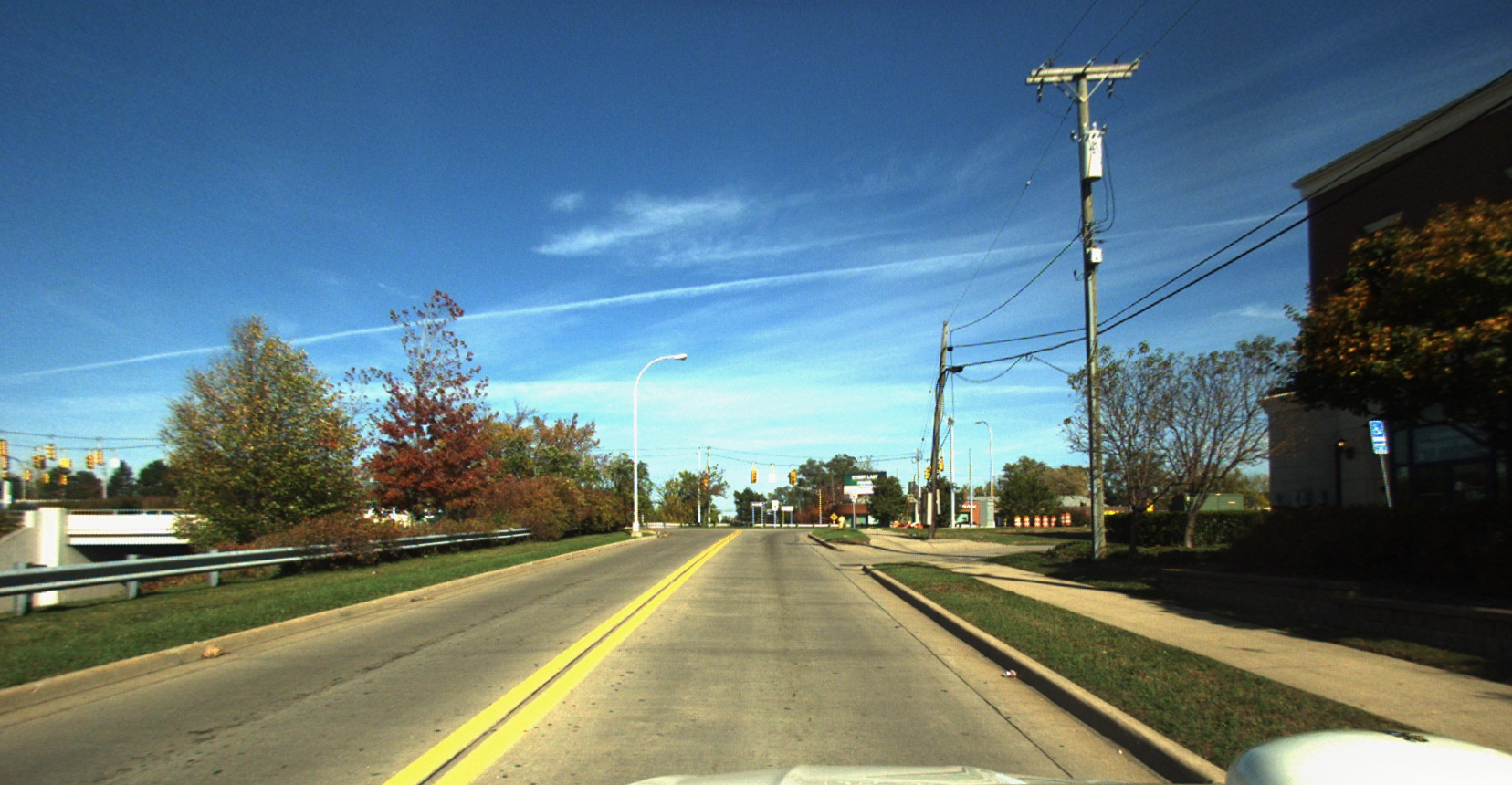}
    \includegraphics[width=0.3\linewidth]{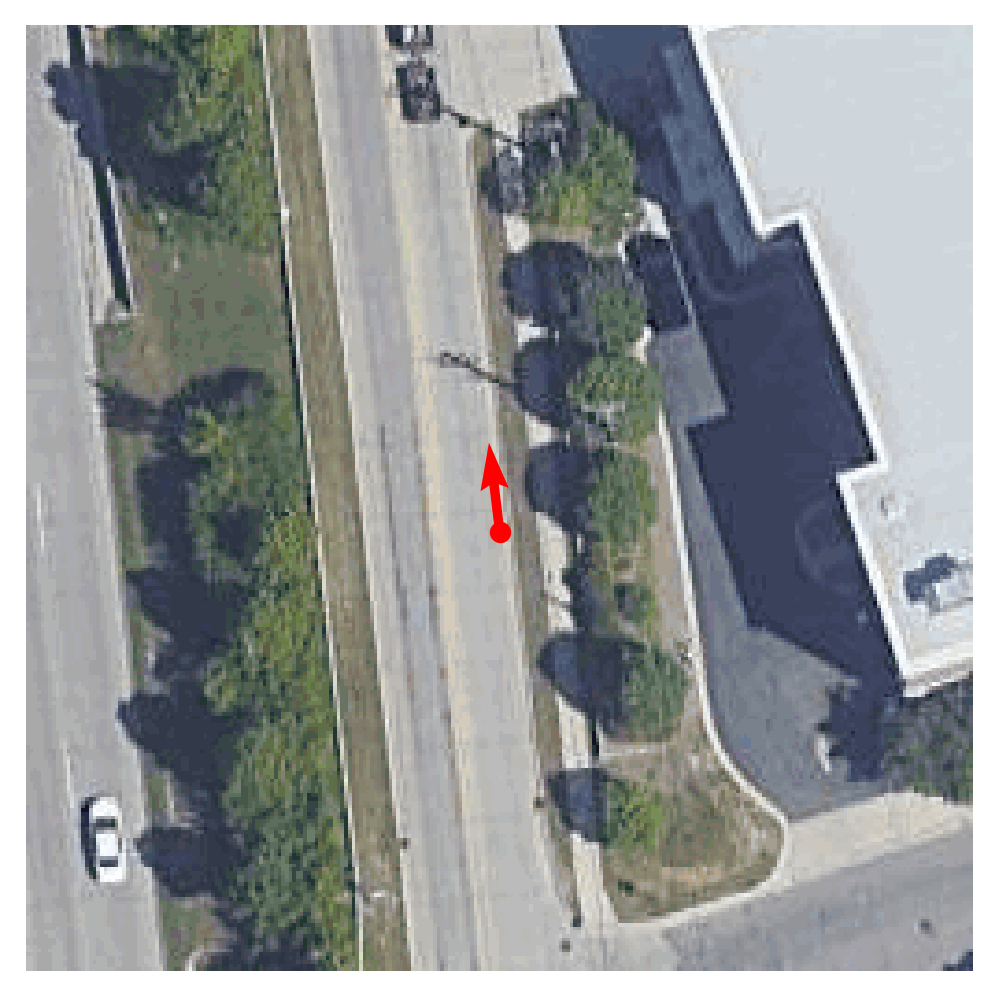}
    \caption{The camera is in the middle of the right road, as indicated by the ground image, while its position computed from the GPS tag pinpoints it is near the right boundary of the road, as shown in the satellite image. }
    \end{subfigure}
    \caption{Examples whose GPS tags are \textit{inaccurate}. In the satellite image of each sub-figure, the red point indicates the camera position computed from the GPS tag, and the red arrow marks the camera facing direction. The images are from Log3 of drive 2017-10-26.
    }
    \label{fig:failure_ford}
\end{figure}

\begin{figure}[ht]
\setlength{\abovecaptionskip}{5pt}
\setlength{\belowcaptionskip}{0pt}
    \centering
    \begin{minipage}{0.48\linewidth}
    \centering
    \includegraphics[width=0.66\linewidth, height = 0.3\linewidth]{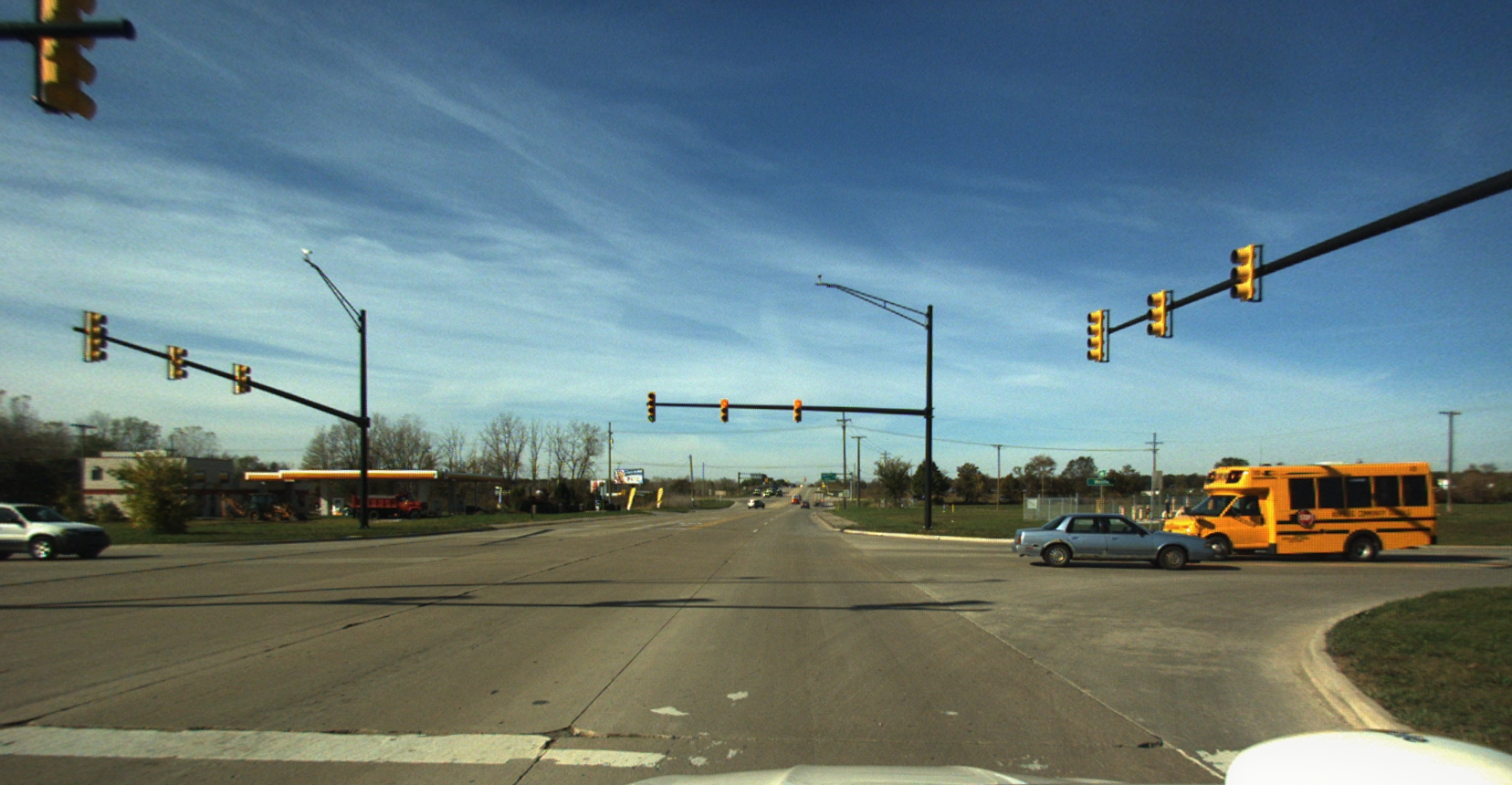}
    \includegraphics[width=0.3\linewidth]{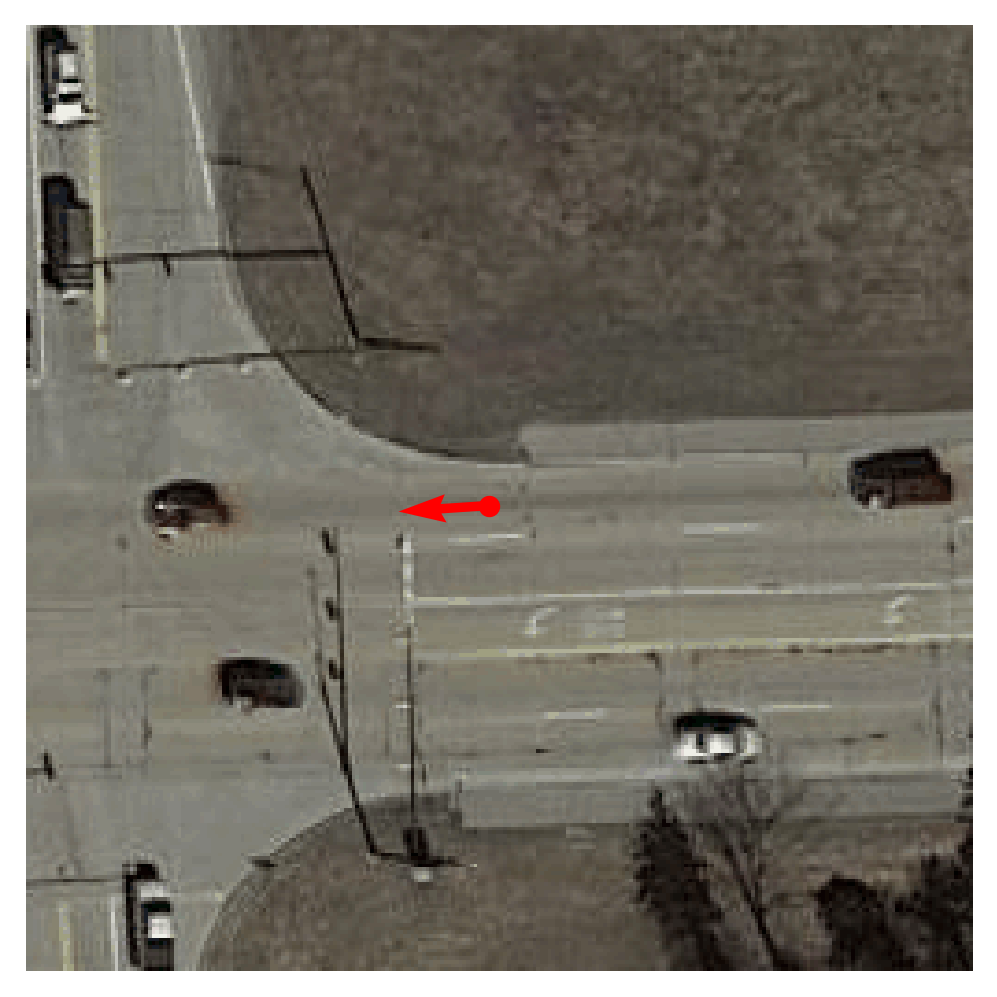}
    \end{minipage}
    \hspace{0.5em}
    \begin{minipage}{0.48\linewidth}
    \centering
    \includegraphics[width=0.66\linewidth, height = 0.3\linewidth]{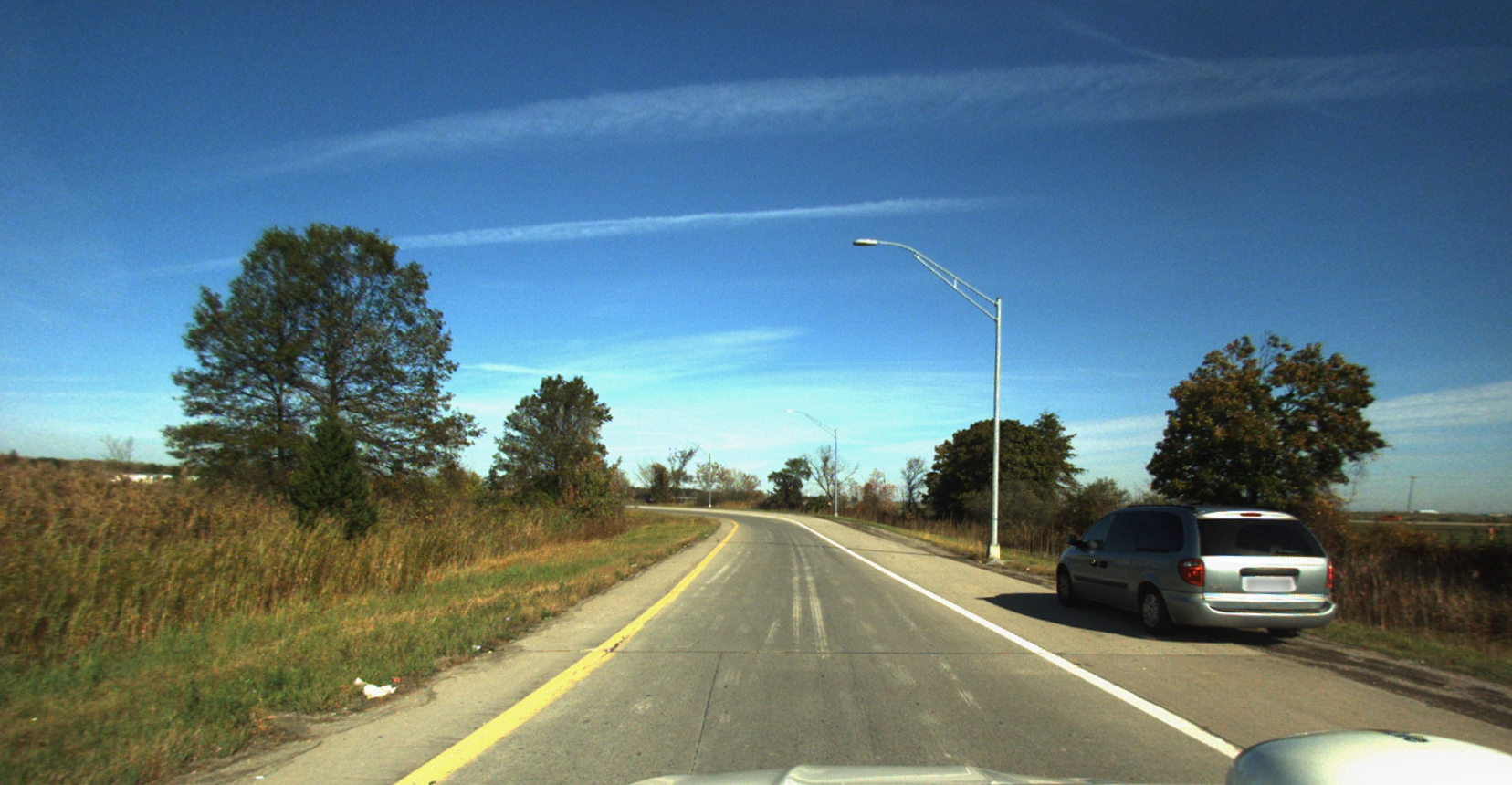}
    \includegraphics[width=0.3\linewidth]{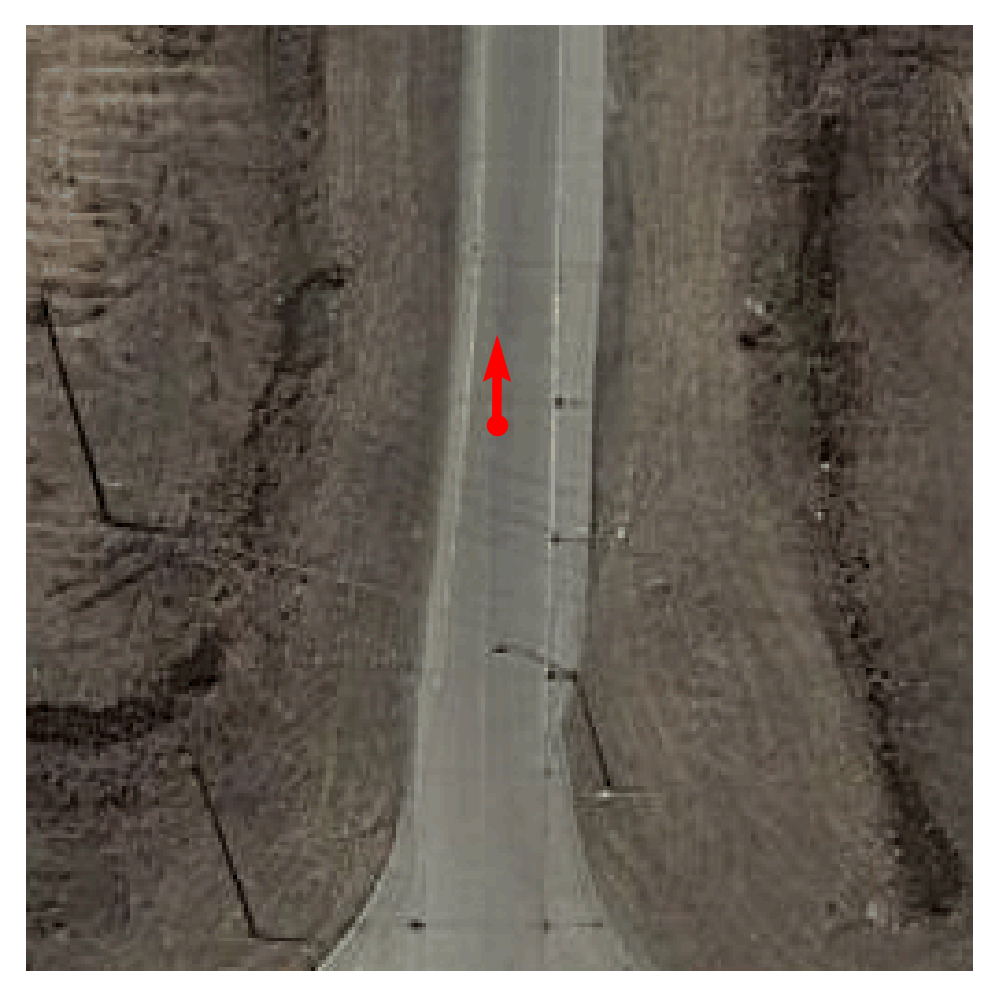}
    \end{minipage}
    \begin{minipage}{0.48\linewidth}
    \centering
    \includegraphics[width=0.66\linewidth, height = 0.3\linewidth]{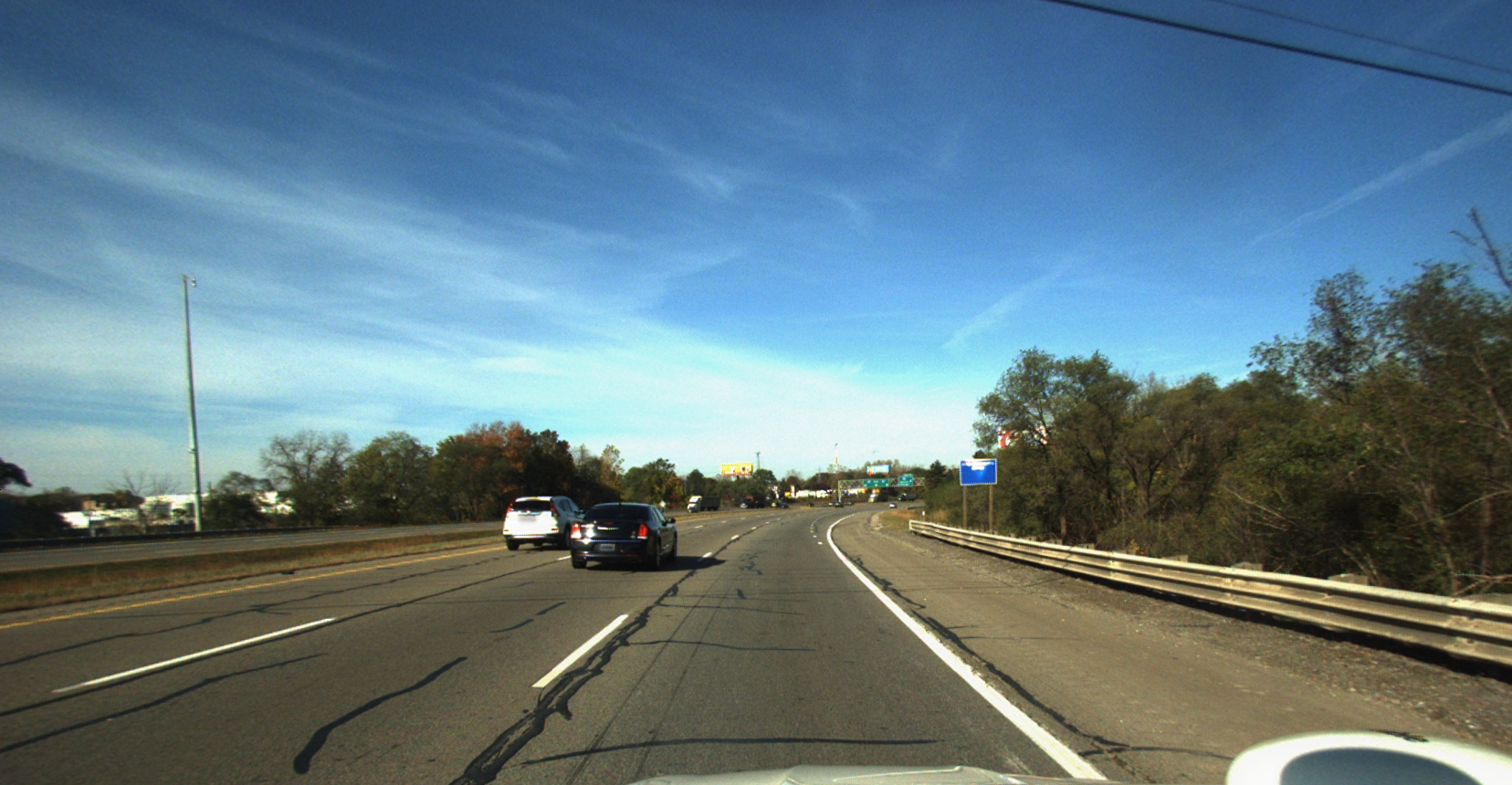}
    \includegraphics[width=0.3\linewidth]{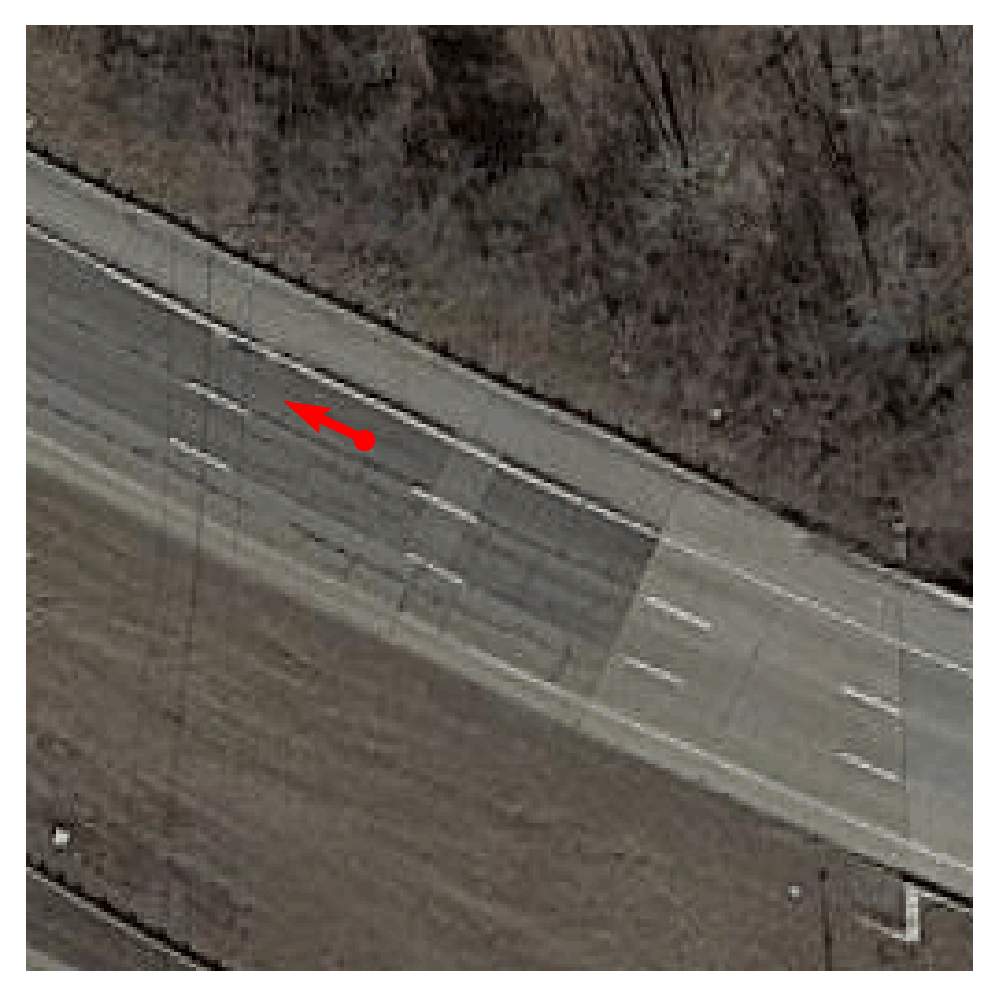}
    \end{minipage}
    \hspace{0.5em}
    \begin{minipage}{0.48\linewidth}
    \centering
    \includegraphics[width=0.66\linewidth, height = 0.3\linewidth]{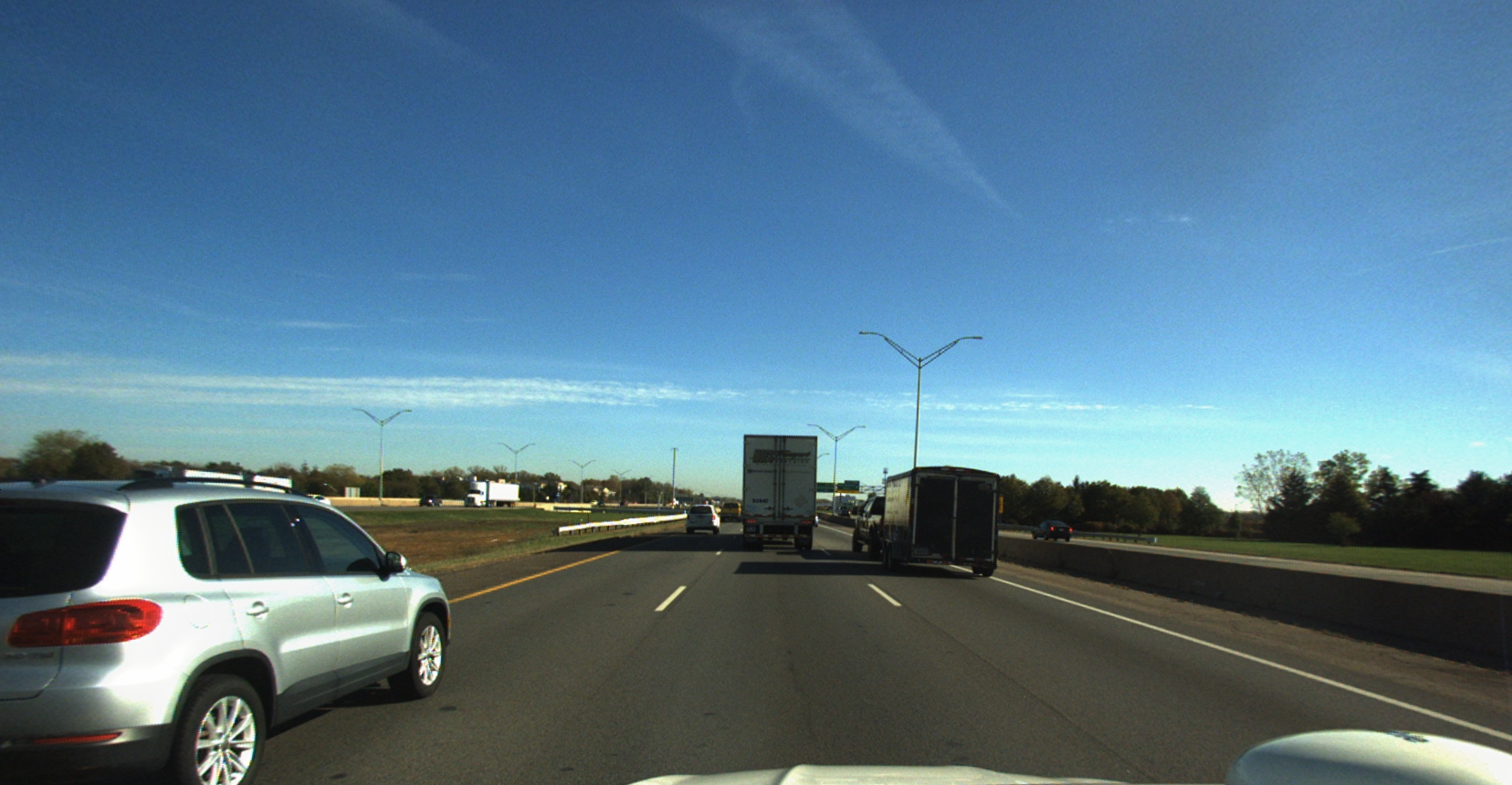}
    \includegraphics[width=0.3\linewidth]{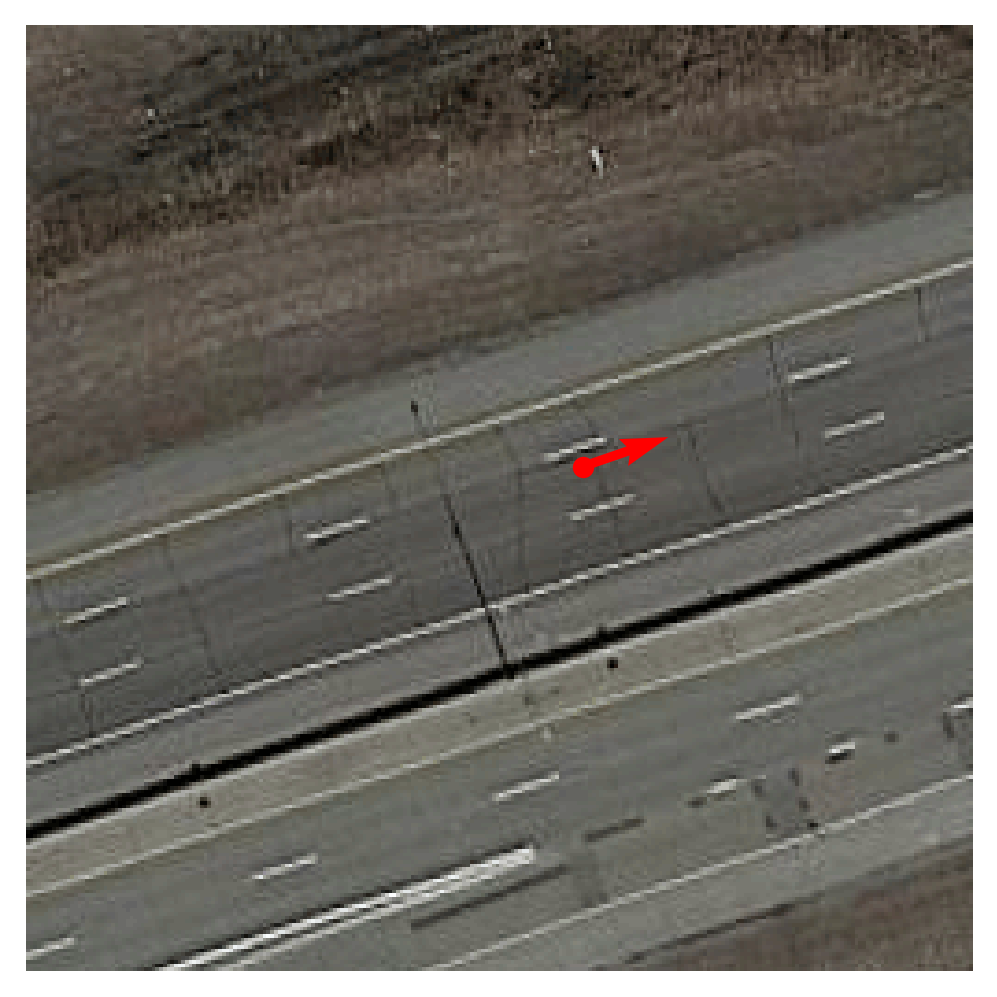}
    \end{minipage}
    \caption{Examples whose GPS tags are \textit{accurate}. In the satellite image of each sub-figure, the red point indicates the camera position computed from the GPS tag, and the red arrow marks the camera facing direction. The images are from Log2 of drive 2017-10-26.}
    \label{fig:success_ford}
\end{figure}

Despite both KITTI and the Ford Multi-AV datasets being captured by accurate survey-grade RTK-GPS systems, we have uncovered that their ground-truth GPS tags are sometimes contaminated by considerable noises.  This can be seen, for example, by marking up the GPS-reported camera position in the satellite image and visually comparing if the observed ground-level scenes as if seen from the ground plane matches well with the marked position in the satellite image.  Fig.~\ref{fig:failure_ford} and Fig.~\ref{fig:success_ford} illustrate some examples from the Ford Dataset, which clearly reveal such mismatches.

We manually filter out those inaccurate ones and construct new subsets for the KITTI and the Ford multi-AV dataset to train and evaluate our new localization method.  The training and testing image numbers of the two datasets are presented in Tab.~\ref{tab:kitti} and Tab.~\ref{tab:ford}, respectively. 

To validate such a pre-filtering is necessary, we conducted comparisons between ``training on the full dataset" and ``training on the filtered dataset" on the first two logs of the Ford multi-AV dataset. The results are presented in Tab.~\ref{tab:all_filterd}.  
They are evaluated on the same test sets for fair comparisons. 
It can be seen that the pre-filtering strategy significantly boosts the performance, especially for lateral translation optimization.  

We provide the performance of our method on the remaining logs (Log3$\sim$Log6) of the Ford multi-AV dataset in Tab.~\ref{tab:log3-6}, to complement our results in Sec.~{\color{red}6.1} of the main paper. 

\begin{minipage}{\linewidth}
\centering
\begin{minipage}{0.28\linewidth}
\centering
\footnotesize
\setlength{\abovecaptionskip}{0pt}
\setlength{\belowcaptionskip}{0pt}
\setlength{\tabcolsep}{4pt}
    \captionof{table}{\small Training and testing image numbers for the KITTI dataset.}
    \begin{tabular}{c|c|c|c}
    \toprule
           &  Training & Test1 & Test2\\ \midrule
    \#Image & 19.655    & 3,773 & 7,542 \\ \bottomrule
    \end{tabular}
    \label{tab:kitti}
\end{minipage}
\hfill
\begin{minipage}{0.7\linewidth}
\centering
\footnotesize
\setlength{\abovecaptionskip}{0pt}
\setlength{\belowcaptionskip}{0pt}
\setlength{\tabcolsep}{2pt}
    \captionof{table}{\small Training and testing splits for the Ford multi-AV dataset. (The training and testing sets of Log3 are from the same drive but different locations.)}
    \begin{tabular}{c|c|c|c|c|c|c|c}
    \toprule
                               &                & Log1        & Log2        & Log3        & Log4       & Log5        & Log6       \\\midrule
    \multirow{2}{*}{Training}  &  Drive         & 2017-10-26  & 2017-10-26  & 2017-08-04  & 2017-10-26 & 2017-08-04  & 2017-08-04 \\
                               &  \#Image  & 4,000       & 10,350      & 1,500       & 7466       & 8430        & 3857       \\\midrule
    \multirow{2}{*}{Testing}   &  Drive         & 2017-08-04  & 2017-08-04  & 2017-08-04  & 2017-08-04 & 2017-10-26  & 2017-10-26 \\
                               &  \#Image  & 2,100       & 3,727       & 1,500       & 3,511      & 3,500       & 1,000       \\ \bottomrule
    \end{tabular}
    \label{tab:ford}
\end{minipage}
\end{minipage}



\begin{table*}[ht!]
\setlength{\abovecaptionskip}{0pt}
\setlength{\belowcaptionskip}{0pt}
\setlength{\tabcolsep}{2pt}
\centering
\footnotesize
\caption{\small Performance comparison of our method on the first two logs of the Ford multi-AV dataset, when trained on the ``Full Dataset'' or the ``Filtered Dataset''.}
\begin{tabular}{c|ccc|ccc|ccc|ccc|ccc|ccc}
\toprule
               & \multicolumn{9}{c|}{Log1}                                                                                                                             & \multicolumn{9}{c}{Log2}                                                                                                                             \\
           & \multicolumn{3}{c|}{Lateral}                          & \multicolumn{3}{c|}{Longitudinal}                         & \multicolumn{3}{c|}{Azimuth}                      & \multicolumn{3}{c|}{Lateral}                          & \multicolumn{3}{c|}{Longitudinal}                         & \multicolumn{3}{c}{Azimuth}                      \\
           & $d=1$          & $d=3$          & $d=5$          & $d=1$         & $d=3$          & $d=5$          & $\theta=1$     & $\theta=3$     & $\theta=5$     & $d=1$          & $d=3$          & $d=5$          & $d=1$         & $d=3$          & $d=5$          & $\theta=1$     & $\theta=3$     & $\theta=5$     \\\midrule
Full Dataset      & 26.67          & 64.76          & \textbf{79.76} & 5.14          & 15.48          & 24.14          & 28.81          & 66.14          & \textbf{81.24} & 22.14          & 58.06          & 71.18          & \textbf{5.47} & \textbf{16.15} & \textbf{25.95} & \textbf{9.98} & 30.35          & 49.26          \\
Filtered Dataset & \textbf{46.10} & \textbf{70.38} & 72.90          & \textbf{5.29} & \textbf{16.38} & \textbf{26.90} & \textbf{44.14} & \textbf{72.67} & 80.19          & \textbf{31.20} & \textbf{66.46} & \textbf{78.27} & 4.80          & 15.27          & 25.76          & 9.74          & \textbf{30.83} & \textbf{51.62} \\ \bottomrule
\end{tabular}
\label{tab:all_filterd}
\end{table*}


\begin{table*}[ht]
\setlength{\abovecaptionskip}{0pt}
\setlength{\belowcaptionskip}{0pt}
\setlength{\tabcolsep}{2pt}
\centering
\footnotesize
\caption{\small Performance of our method on the remaining logs of the Ford multi-AV dataset.}
\begin{tabular}{c|ccc|ccc|ccc|c|ccc|ccc|ccc}
\toprule
     & \multicolumn{3}{c|}{Lateral} & \multicolumn{3}{c|}{Longitudinal} & \multicolumn{3}{c|}{Azimuth}          &      & \multicolumn{3}{c|}{Lateral} & \multicolumn{3}{c|}{Longitudinal} & \multicolumn{3}{c}{Azimuth}          \\ 
     & $d=1$   & $d=3$   & $d=5$   & $d=1$     & $d=3$     & $d=5$    & $\theta=1$ & $\theta=3$ & $\theta=5$ &      & $d=1$   & $d=3$   & $d=5$   & $d=1$     & $d=3$     & $d=5$    & $\theta=1$ & $\theta=3$ & $\theta=5$ \\\midrule
Log3 & 11.40   & 34.00   & 58.13   & 4.47      & 13.13     & 22.47    & 8.93       & 29.73      & 48.80      & Log4 & 29.96   & 66.28   & 74.88   & 4.96      & 15.52     & 25.92    & 14.33      & 43.69      & 67.45      \\
Log5 & 15.26   & 54.60   & 76.71   & 6.23      & 19.89     & 32.34    & 17.74      & 47.60      & 67.74      & Log6 & 20.20   & 45.20   & 59.00   & 3.90      & 14.30     & 24.50    & 10.80      & 31.80      & 52.50      \\ \bottomrule
\end{tabular}
\label{tab:log3-6}
\end{table*}

\section{Increasing the Grid Sample Density for Image Retrieval-based Methods}
\begin{figure}[ht!]
\setlength{\abovecaptionskip}{5pt}
\setlength{\belowcaptionskip}{0pt}
    \centering
    \includegraphics[width=0.12\linewidth]{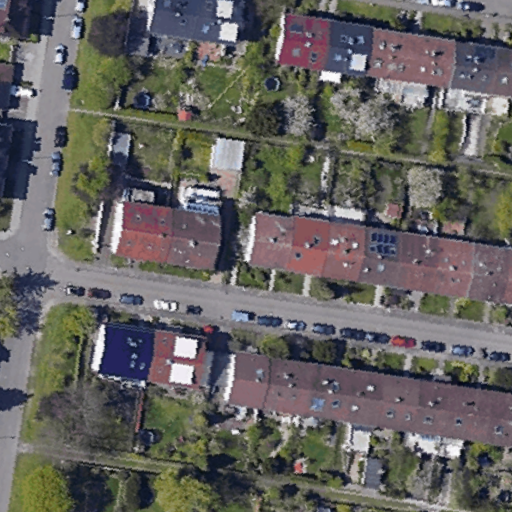}
    \includegraphics[width=0.12\linewidth]{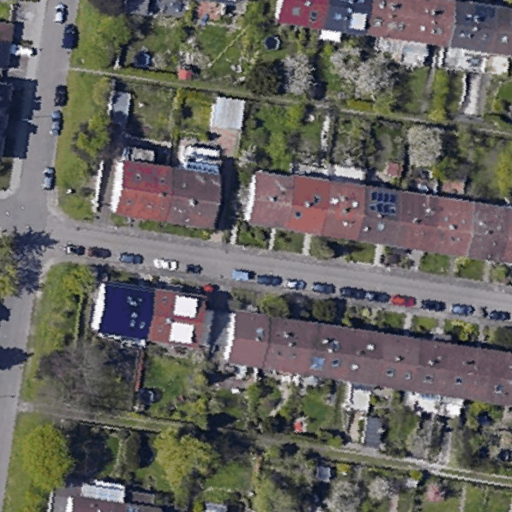}
    \includegraphics[width=0.12\linewidth]{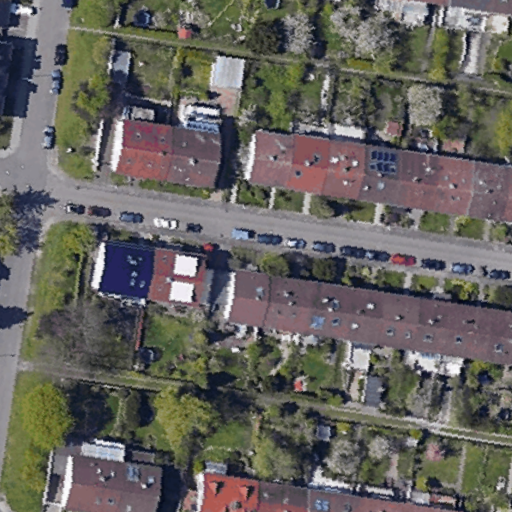}
    \includegraphics[width=0.12\linewidth]{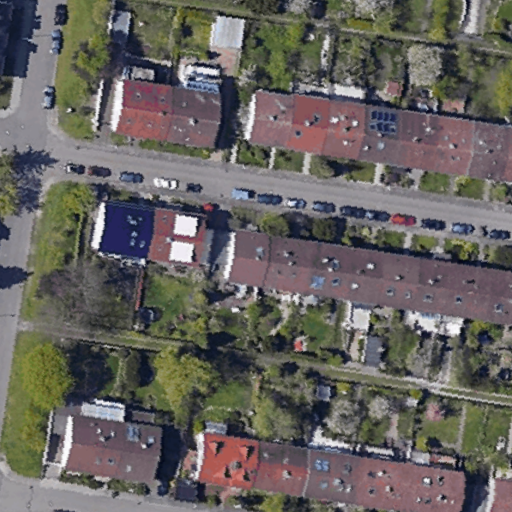}
    \includegraphics[width=0.12\linewidth]{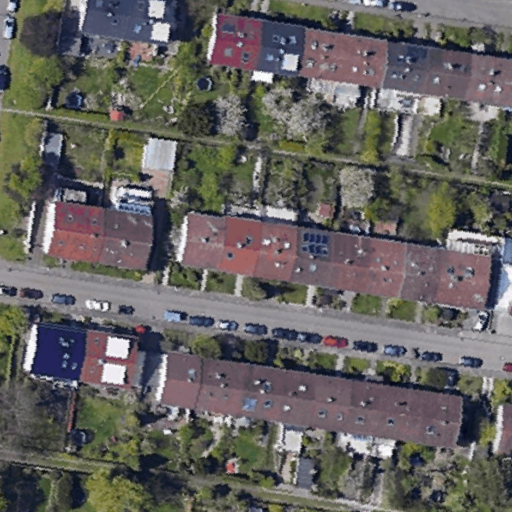}
    \includegraphics[width=0.12\linewidth]{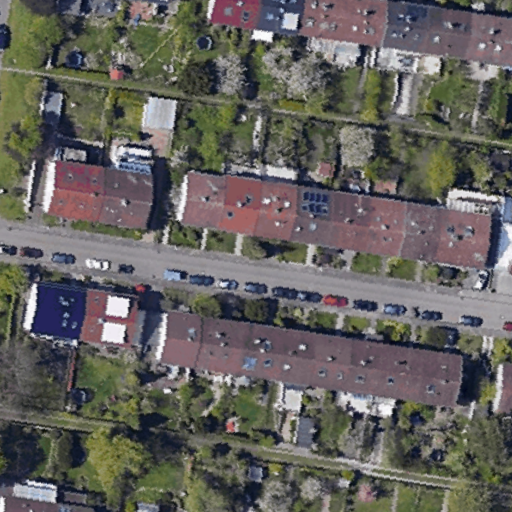}
    \includegraphics[width=0.12\linewidth]{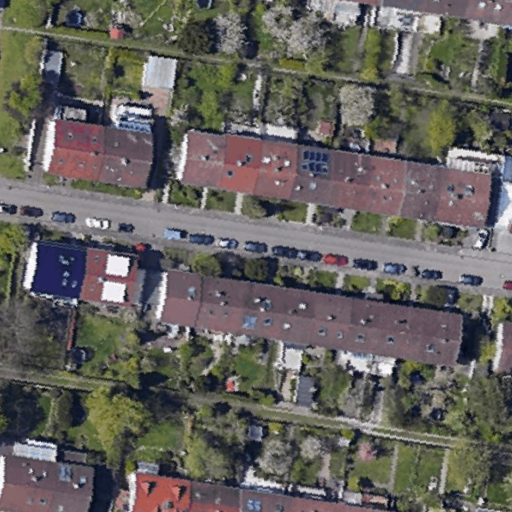}
    \includegraphics[width=0.12\linewidth]{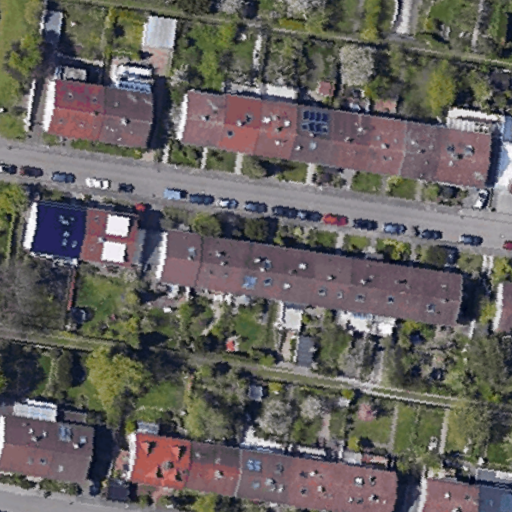}\\
    \includegraphics[width=0.12\linewidth]{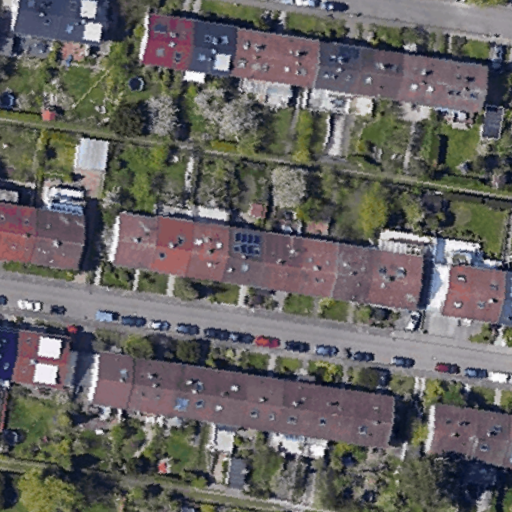}
    \includegraphics[width=0.12\linewidth]{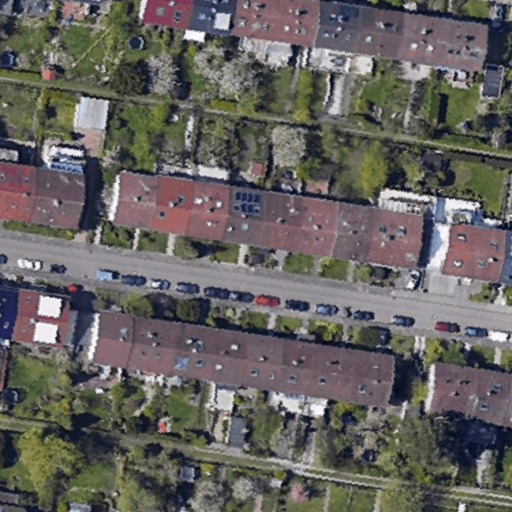}
    \includegraphics[width=0.12\linewidth]{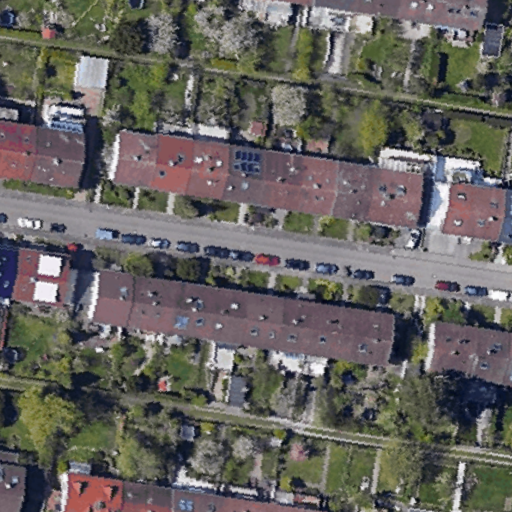}
    \includegraphics[width=0.12\linewidth]{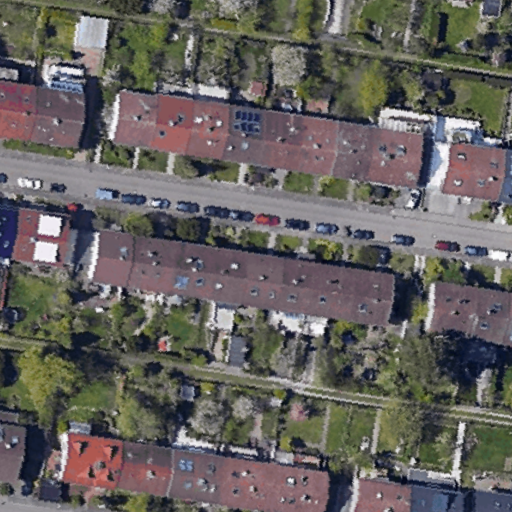}
    \includegraphics[width=0.12\linewidth]{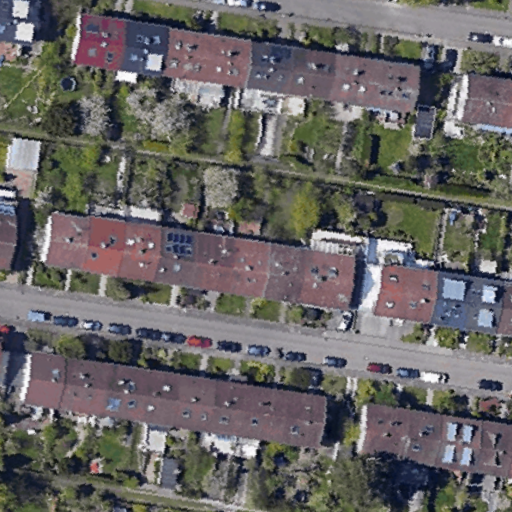}
    \includegraphics[width=0.12\linewidth]{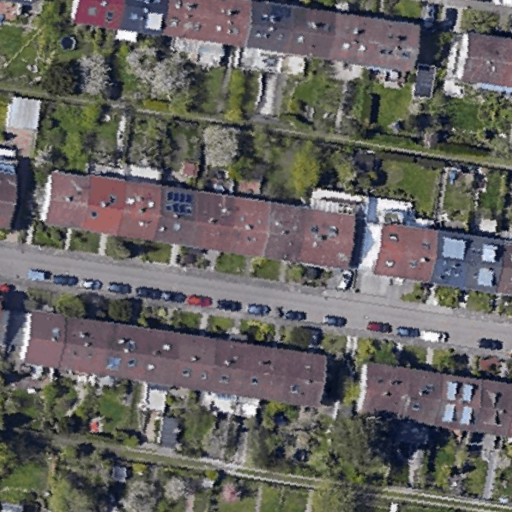}
    \includegraphics[width=0.12\linewidth]{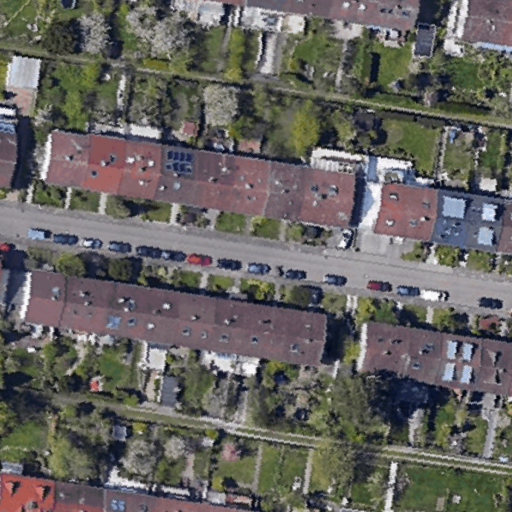}
    \includegraphics[width=0.12\linewidth]{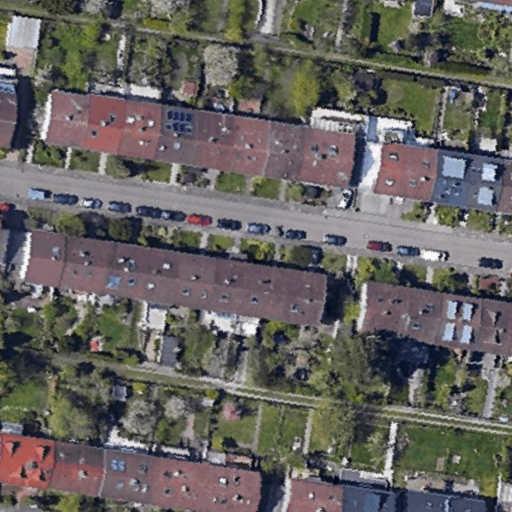}\\
    \caption{The database images for fine-grained image retrieval using a grid of $4\times 4$. They are very similar and hard to discriminate.}
    \label{fig:IR_grids}
\end{figure}

In this section, we provide additional experiments to investigate the performance of image retrieval-based methods when increasing the grid sample density in constructing the database. 
Among the state-of-the-art cross-view image retrieval algorithms, DSM [{\color{green}45}] and VIGOR [{\color{green}67}] are two of the performers.  We therefore only compared ours with these two algorithms. 
From the results in Tab.~\ref{tab:IR_grid}, we did not observe consistent positive effects when increasing the grid sample density. 
This might be because, in the fine-grained retrieval-based localization, the database images using a grid of $4\times 4$ are already very similar and hard to discriminate. 
Thus, increasing the sample density of database images does not help. 
Fig.~\ref{fig:IR_grids} presents some examples of the database images sampled using a grid of $4\times 4$.

\begin{table*}[ht]
\setlength{\abovecaptionskip}{0pt}
\setlength{\belowcaptionskip}{0pt}
\setlength{\tabcolsep}{2pt}
\centering
\footnotesize
\caption{\small Performance of image retrieval-based methods when increasing the grid sample density on the KITTI dataset.}
\begin{tabular}{c|c|ccc|ccc|ccc|ccc|ccc|ccc}
\toprule
\multirow{3}{*}{}              & \multirow{3}{*}{Grid} & \multicolumn{9}{c|}{Test1}                                                                                                                                                                                   & \multicolumn{9}{c}{Test2}                                                                                                                                                                 \\
                               &                       & \multicolumn{3}{c|}{Lateral}                                        & \multicolumn{3}{c|}{Longitudinal}                                  & \multicolumn{3}{c|}{Azimuth}                                        & \multicolumn{3}{c|}{Lateral}                                        & \multicolumn{3}{c|}{Longitudinal}                & \multicolumn{3}{c}{Azimuth}                                        \\
                               &                       & $d=1$                & $d=3$                & $d=5$                & $d=1$               & $d=3$                & $d=5$                & $\theta=1$           & $\theta=3$           & $\theta=5$           & $d=1$                & $d=3$                & $d=5$                & $d=1$         & $d=3$          & $d=5$          & $\theta=1$           & $\theta=3$           & $\theta=5$           \\ \midrule
\multirow{4}{*}{DSM [{\color{green}45}]}           & $4 \times 4$          & 12.00                & 35.29                & {{53.67}} & 4.33                & 12.48                & 21.43                & 3.52                & 13.33                & 23.67                & 8.45                 & 24.85                & 37.64                & 3.94                & 12.24                & 21.41                & 2.23                & 7.67                 & 13.42                \\
                               & $5 \times 5$          & 11.69                & 33.34                & 50.25                & 4.51                & {{13.68}} & 21.55                & {{3.66}} & 13.65                & 24.49                & 11.44                & 33.16                & 50.76                & {{4.11}} & 12.13                & 20.35                & 3.20                & 13.35                & 23.67                \\
                               & $6 \times 6$          & 12.72                & 34.35                & 50.15                & 4.53                & 12.70                & 21.89                & 3.45                & 13.65                & 24.44                & 12.25                & 34.31                & {{51.83}} & 4.04                & 12.49                & 21.13                & {{3.37}} & {{13.55}} & {{23.77}} \\
                               & $7 \times 7$          & {{12.80}} & {{35.38}} & 50.41                & {{4.93}} & 13.60                & {{22.55}} & 3.60                & {{13.91}} & {{25.10}} & {{12.42}} & {{34.91}} & 51.72                & 3.99                & {{12.56}} & {{21.49}} & 3.31                & 13.14                & 23.38                \\ \midrule
\multirow{4}{*}{VIGOR [{\color{green}67}]}         & $4 \times 4$          & {{20.33}} & {{52.48}} & 70.43                & {{6.19}} & {{16.05}} & {{25.76}} & -                   & -                    & -                    & {{20.87}} & {{54.87}} & {{75.64}} & \textbf{{5.98}} & \textbf{{16.88}} & \textbf{{27.23}} & -                   & -                    & -                    \\
                               & $5 \times 5$          & 18.98                & 48.85                & 70.34                & 4.59                & 13.89                & 22.77                & -                   & -                    & -                    & 16.83                & 48.38                & 71.15                & 4.08                & 12.32                & 20.91                & -                   & -                    & -                    \\
                               & $6 \times 6$          & 17.84                & 48.98                & 70.39                & 5.17                & 14.58                & 24.07                & -                   & -                    & -                    & 17.54                & 48.46                & 71.40                & 4.46                & 13.56                & 22.01                & -                   & -                    & -                    \\
                               & $7 \times 7$          & 18.50                & 49.06                & {70.55}       & 4.90                & 14.15                & 23.43                & -                   & -                    & -                    & 17.37                & 48.48                & 71.68                & 4.36                & 13.71                & 22.29                & -                   & -                    & -                    \\ \midrule
\textbf{Ours} & -                     & \textbf{35.54}       & \textbf{70.77}       & \textbf{80.36}       & \textbf{5.22}       & \textbf{15.88}       & \textbf{26.13}       & \textbf{19.64}      & \textbf{51.76}       & \textbf{71.72}       & \textbf{27.82}       & \textbf{59.79}       & \textbf{72.89}       & 5.75                & 16.36                & 26.48                & \textbf{18.42}      & \textbf{49.72}       & \textbf{71.00}      
\\ \bottomrule
\end{tabular}
\label{tab:IR_grid}
\end{table*}

\section{Different Initial Values}

In Tab.~\ref{tab:search_region}, we show the performance of our method with different pose initialization ranges. The performance increases as the search range decreases. 
The consumer-level GPS accuracy ranges from 15m to 20m, and the image retrieval methods [44, 54] can make their top-1 retrieved results be within $5$m to their ground truth. 
Since the primary purpose of this paper is to study whether we can refine an initial coarse estimate by cross-view matching, we set our search region as 40m$\times$40m in the paper. 

\begin{figure}[t!]
\setlength{\abovecaptionskip}{0pt}
    \setlength{\belowcaptionskip}{0pt}
    \centering
    \includegraphics[width=0.5\linewidth]{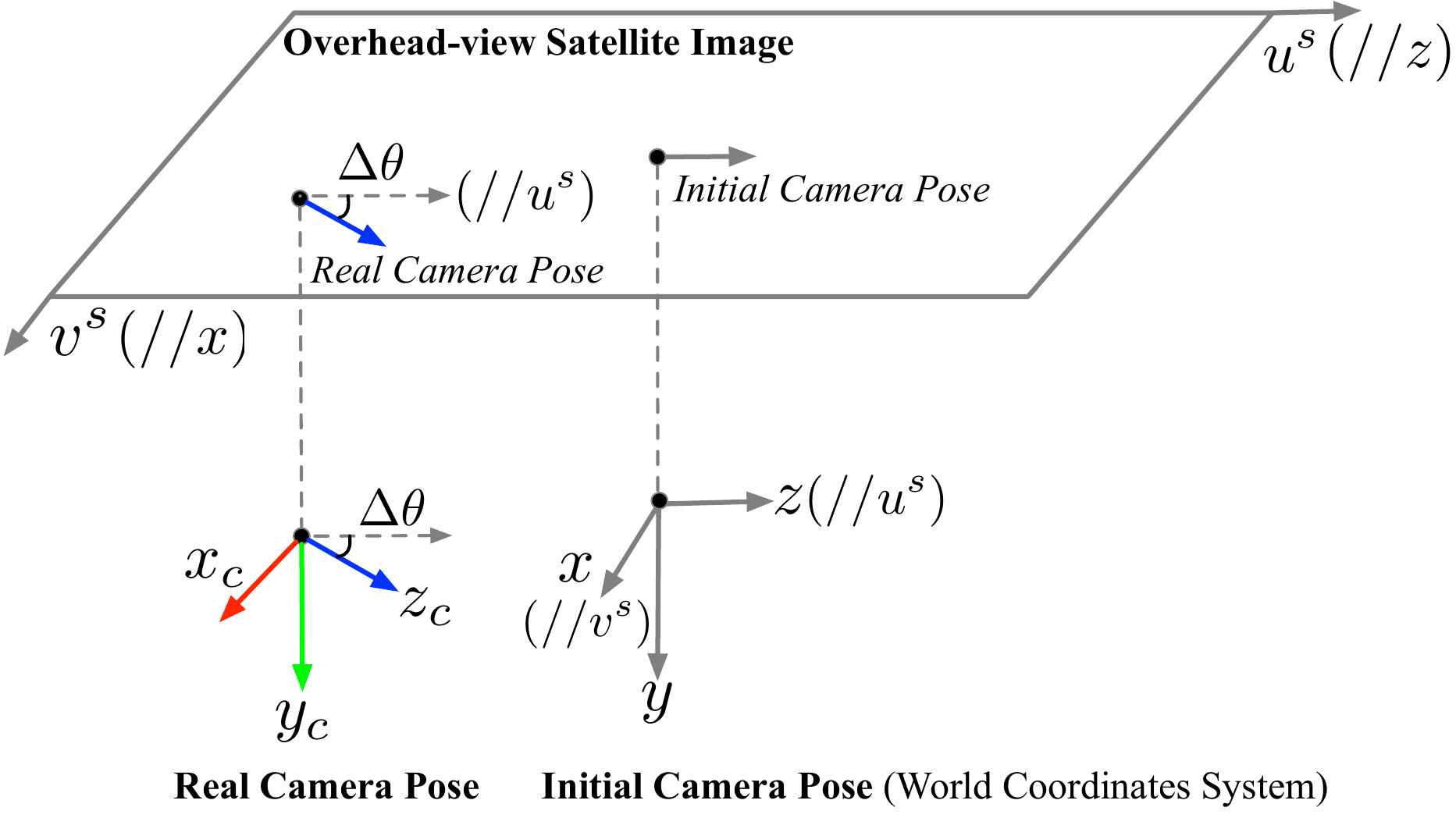}
  \caption{ Coordinates illustration. Note that this is only for illustration purpose. The coordinates used in our codes are slightly different with this one.
  }
  \label{fig:coordinates}
\end{figure}

\section{Additional Comparisons}

\smallskip \noindent
\textbf{Ours w/o Long. } 
We investigate whether the loss item on longitudinal pose estimation can be removed, denoted as ``Ours w/o Long''. 
As shown in the first row of Tab.~\ref{tab:abla}, this results in a negative effect, indicating that the longitudinal pose constraints contribute to learning discriminative features, although the ambiguity along this direction is high.

\smallskip \noindent
\textbf{Different iteration strategies.}
In our framework, the LM optimization is first applied to the multi-level features from coarse to fine (C2F), and then the C2F update is executed iteratively. 
Here, we study the performance of the LM optimization when it is first applied to the coarsest feature level until the maximum iteration and then propagates to finer levels, denoted as ``C2F Global''.  
The results are presented in the second row of Tab.~\ref{tab:abla}. 
Compared to C2F Global, our update strategy guarantees fine-tuning around more possible solutions and thus is more likely to find the global optimum. 

\section{Coordinates Illustration and Pose Parameterization}
 We set the world coordinates system to the initial camera pose estimate, as shown in Fig.~\ref{fig:coordinates}. For illustration brevity, we pre-align the satellite image to make its center correspond to the initial camera position and its $u$ direction parallel to the initial camera facing direction. 
  Here, both $z$ and $z_c$ denote the camera facing direction. 
  
  Denote $\Delta x$ is the lateral translation, $\Delta z$ is the longitudinal translation, and $\theta$ is the azimuth angle. The query ground camera pose in Eq. (2) and Eq. (4) in the main paper is parameterized as
\begin{equation}
 \mathbf{R} = \begin{bmatrix}
\cos \theta & 0 & \sin \theta \\
0 & 1 & 0 \\
-\sin \theta & 0 & \cos \theta   \\
\end{bmatrix}, \quad \mathbf{t} = \begin{bmatrix}
\Delta x  \\
 0 \\
 \Delta z \\
\end{bmatrix}.
\end{equation}

\section{Broader Impact}
This paper has introduced a new technique for high-accuracy vehicle/camera localization. 
This technique can provide accurate camera position estimation even in a GPS-denied environment.  The position of a vehicle or camera of a user is often considered sensitive or private information. The proposed technique may be abused or misused, causing privacy violations. We advocate careful data protection and model management to mitigate the risk.

\begin{table}[]
\setlength{\abovecaptionskip}{0pt}
\setlength{\belowcaptionskip}{0pt}
\setlength{\tabcolsep}{3pt}
\centering
\footnotesize
\caption{\small Performance comparison with different search regions on the KITTI dataset.}
\begin{tabular}{c|ccc|ccc|ccc|ccc|ccc|ccc}
\toprule
\multirow{3}{*}{\begin{tabular}[c]{@{}c@{}}Search \\ Region\end{tabular}} & \multicolumn{9}{c|}{Test1}                                                                             & \multicolumn{9}{c}{Test2}                                                                             \\
                                                                          & \multicolumn{3}{c|}{Lateral} & \multicolumn{3}{c|}{Longitudinal} & \multicolumn{3}{c|}{Azimuth}          & \multicolumn{3}{c|}{Lateral} & \multicolumn{3}{c|}{Longitudinal} & \multicolumn{3}{c}{Azimuth}          \\
                                                                          & $d=1$   & $d=3$   & $d=5$   & $d=1$     & $d=3$     & $d=5$    & $\theta=1$ & $\theta=3$ & $\theta=5$ & $d=1$   & $d=3$   & $d=5$   & $d=1$     & $d=3$     & $d=5$    & $\theta=1$ & $\theta=3$ & $\theta=5$ \\ \midrule
40m$\times$40m                                                            & 35.54   & 70.77   & 80.36   & 5.22      & 15.88     & 26.13    & 19.64      & 51.76      & 71.72      & 27.82   & 59.79   & 72.89   & 5.75      & 16.36     & 26.48    & 18.42      & 49.72      & 71.00      \\
20m$\times$20m                                                            & 44.66   & 73.92   & 81.18   & 12.06     & 35.62     & 54.73    & 25.31      & 57.41      & 74.48      & 34.17   & 72.30   & 81.15   & 11.56     & 35.08     & 53.77    & 11.40      & 48.18      & 65.80      \\
10m$\times$10m                                                            & 64.86   & 92.23   & 96.98   & 29.08     & 69.49     & 88.66    & 36.92      & 73.95      & 86.88      & 55.98   & 90.84   & 96.43   & 25.97     & 66.96     & 88.12    & 31.36      & 69.46      & 84.50     
\\ \bottomrule
\end{tabular}
\label{tab:search_region}
\end{table}

\begin{table*}[t!]
\setlength{\abovecaptionskip}{0pt}
\setlength{\belowcaptionskip}{0pt}
\setlength{\tabcolsep}{2.5pt}
\centering
\footnotesize
\caption{\small Additional ablation study results of our method on the KITTI dataset.}
\begin{tabular}{c|ccc|ccc|ccc|ccc|ccc|ccc}
\toprule
              & \multicolumn{9}{c|}{Test1}                                                                                                                             & \multicolumn{9}{c}{Test2}                                                                                                                             \\
              & \multicolumn{3}{c|}{Lateral}                          & \multicolumn{3}{c|}{Longitudinal}                         & \multicolumn{3}{c|}{Azimuth}                      & \multicolumn{3}{c|}{Lateral}                          & \multicolumn{3}{c|}{Longitudinal}                         & \multicolumn{3}{c}{Azimuth}                      \\
              & $d=1$          & $d=3$          & $d=5$          & $d=1$         & $d=3$          & $d=5$          & $\theta=1$     & $\theta=3$     & $\theta=5$     & $d=1$          & $d=3$          & $d=5$          & $d=1$         & $d=3$          & $d=5$          & $\theta=1$     & $\theta=3$     & $\theta=5$     \\ \toprule
Ours w/o Long  & 25.63          & 56.72          & 69.55          & \textbf{5.99} & \textbf{16.06} & \textbf{26.85} & 13.84          & 39.01          & 59.98          & 20.50          & 52.52          & 67.57          & 5.32          & 15.16          & 25.23          & 12.90          & 36.79          & 57.73          \\
C2F Global    & 23.32          & 50.60          & 61.25          & 5.27          & 15.88          & 26.05          & 11.87          & 33.66          & 54.86          & 20.43          & 45.86          & 58.51          & 5.25          & 15.82          & 26.16          & 11.65          & 33.65          & 54.02          \\
\textbf{Ours} & \textbf{35.54} & \textbf{70.77} & \textbf{80.36} & 5.22          & 15.88          & 26.13          & \textbf{19.64} & \textbf{51.76} & \textbf{71.72} & \textbf{27.82} & \textbf{59.79} & \textbf{72.89} & \textbf{5.75} & \textbf{16.36} & \textbf{26.48} & \textbf{18.42} & \textbf{49.72} & \textbf{71.00} \\ \bottomrule
\end{tabular}
\label{tab:abla}
\end{table*}

\end{document}